\documentclass[final]{cvpr}

\usepackage{longtable}  % include before arydshln

\usepackage{times}
\usepackage{epsfig}
\usepackage{graphicx}
\usepackage{amssymb}
\usepackage{amsmath}
\usepackage{multirow}
\usepackage{arydshln}
\usepackage{enumitem}
\usepackage{balance}
\usepackage{comment}
\usepackage{caption}

\usepackage{subdepth}
\usepackage{booktabs}

\usepackage[pagebackref=true,breaklinks=true,colorlinks,bookmarks=false]{hyperref}

\newcommand{\vect}[1]{#1} % 3D vectors (usually 
\newcommand{\id}{\mathbf{I}} % identity matrix

\def\cvprPaperID{3360} % *** Enter the CVPR Paper ID here
\def\confYear{CVPR 2021}
\begin{document}

%%%%%%%%% TITLE
\title{Square Root Bundle Adjustment for Large-Scale Reconstruction}
\def\myquad{\hskip2.5em\relax}
\author{Nikolaus Demmel \myquad Christiane Sommer \myquad Daniel Cremers \myquad Vladyslav Usenko\\
Technical University of Munich\\
{\tt\small \{nikolaus.demmel,c.sommer,cremers,vlad.usenko\}@tum.de}
% For a paper whose authors are all at the same institution,
% omit the following lines up until the closing ``}''.
% Additional authors and addresses can be added with ``\and'',
% just like the second author.
% To save space, use either the email address or home page, not both
}

\maketitle
\thispagestyle{empty}

%\ifcvprfinal
\let\thefootnote\relax\footnote{This work was supported by the Munich Center for Machine Learning, by the ERC Advanced Grant SIMULACRON and by the DFG project CR~250/20-1 ``Splitting Methods for 3D Reconstruction and SLAM.''}
%\fi

%%%%%%%%% ABSTRACT
\begin{abstract}
   We propose a new formulation for the bundle adjustment problem which relies on nullspace marginalization of landmark variables by QR decomposition. 
   Our approach, which we call square root bundle adjustment, is algebraically equivalent to the commonly used Schur complement trick, improves the numeric stability of computations, and allows for solving large-scale bundle adjustment problems with single-precision floating-point numbers.
   We show in real-world experiments with the BAL datasets that even in single precision the proposed solver achieves on average equally accurate solutions compared to Schur complement solvers using double precision. It runs significantly faster, but can require larger amounts of memory on dense problems.
   The proposed formulation relies on simple linear algebra operations and opens the way for efficient implementations of bundle adjustment on hardware platforms optimized for single-precision linear algebra processing.
\end{abstract}

%%%%%%%%% BODY TEXT
\section{Introduction}

\begin{figure}
  \includegraphics[width=\linewidth]{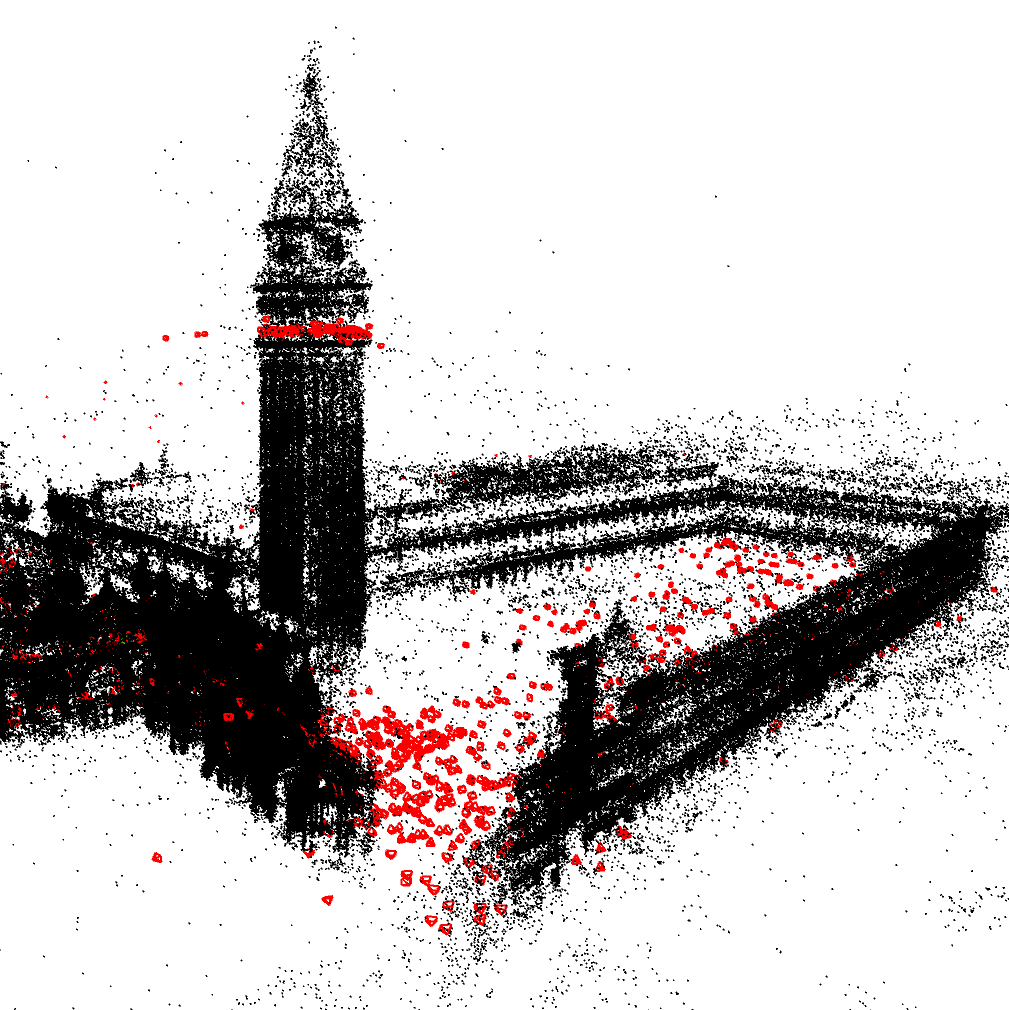}
  \caption{
  Optimized 3D reconstruction of the largest \emph{venice} BAL dataset with 1778 cameras, around one million landmarks, and five million observations.
  For this problem, the proposed square root bundle adjustment ($\sqrt{BA}$) solver is 42\% faster than the best competing method at reaching a cost tolerance of 1\%.
  }
  \label{fig:teaser}
\end{figure}

Bundle adjustment (BA) is a core component of many 3D reconstruction algorithms and structure-from-motion (SfM) pipelines.
It is one of the classical computer vision problems and has been investigated by researchers for more than six decades \cite{brown1958solution}.
While different formulations exist, the underlying question is always the same:
given a set of approximate point (landmark) positions that are observed from a number of cameras in different poses, what are the actual landmark positions and camera poses?
One can already compute accurate 3D positions with only a few images; however, with more available images we will get a more complete reconstruction.
With the emergence of large-scale internet photo collections has come the need to solve bundle adjustment on a large scale, i.e., for thousands of images and hundreds of thousands of landmarks.
This requires scalable solutions that are still efficient for large problems and do not run into memory or runtime issues.

Modern BA solvers usually rely on the Schur complement (SC) trick that computes the normal equations of the original BA problem and then breaks them down into two smaller subproblems, namely (1) the solution for camera poses and (2) finding optimal landmark positions.
This results in the drastically smaller reduced camera system (RCS), which is also better-conditioned~\cite{agarwal2010bundle} than the original normal equations.
The use of the Schur complement has become the de facto standard for solving large-scale bundle adjustment and is hardly ever questioned.

In this work, we challenge the widely accepted assumption of SC being the best choice for solving bundle adjustment, and provide a detailed derivation and analysis of an alternative problem reduction technique based on QR decomposition.
Inspired by similar approaches in the Kalman filter literature~\cite{yang2017null}, we factorize the landmark Jacobian $J_l$ such that we can project the original problem onto the nullspace of $J_l$.
Thus, we circumvent the computation of the normal equations and their system matrix $H$, and are able to directly compute a matrix square root of the RCS while still solving an algebraically equivalent problem.
This improves numerical stability of the reduction step, which is of particular importance on hardware optimized for single-precision floats.
Following terminology for nullspace marginalization on Kalman filters, we call our method \emph{square root bundle adjustment}, short $\sqrt{BA}$.
In particular, our contributions are as follows:
\setlist{nolistsep}
\begin{itemize}[noitemsep]
    \item We propose nullspace marginalization as an alternative to the traditional Schur complement trick and prove that it is algebraically equivalent.
    \item We closely link the very general theory of nullspace marginalization to an efficient implementation strategy that maximally exploits the specific structure of bundle adjustment problems.
    \item We show that the algorithm is well parallelizable and
    that the favorable numerical properties admit computation in single precision,
    resulting in an additional twofold speedup.
    \item We perform extensive evaluation of the proposed approach on the Bundle Adjustment in the Large (BAL) datasets and compare to the state-of-the-art optimization framework Ceres. 
    \item We release our implementation and complete evaluation pipeline as open source to make our experiments reproducible and facilitate further research:\\
    \url{https://go.vision.in.tum.de/rootba}.
\end{itemize}

\section{Related work}

We propose a way to solve large-scale bundle adjustment using QR decomposition, so we review both works on bundle adjustment (with a focus on large-scale problems) and works that use QR decomposition for other tasks in computer vision and robotics.
For a general introduction to numerical algorithms 
(including QR decomposition and iterative methods for solving linear systems),
we refer to~\cite{bjorck1996numerical,golub13}.

\paragraph{(Large-scale) bundle adjustment}
A detailed overview of bundle adjustment in general can be found in~\cite{triggs1999bundle}, including an explanation of the Schur complement (SC) reduction technique~\cite{brown1958solution} to marginalize landmark variables.
Byröd and Åström use the method of conjugate gradients (CG) on the normal equations~\cite{hestenes1952methods, bjorck1979accelerated} to minimize the linearized least squares problem without the Schur complement~\cite{byrod2010conjugate}.
They also QR-decompose the Jacobian, but only for block preconditioning without marginalizing landmarks.
Agarwal et~al.\ have proposed preconditioned CG on the RCS after SC 
to solve the large-scale case~\cite{agarwal2010bundle}, and Wu et~al.\ further extend these ideas to a formulation which avoids explicit computation of the SC matrix~\cite{wu2011multicore}.
A whole number of other works have proposed 
ways to further improve efficiency, accuracy and/or robustness of BA~\cite{engels2006bundle, konolige2010sparse, jeong2011pushing, zach2014robust, natesan2017distributed}, all of them using the Schur complement.
More recently, in Stochastic BA~\cite{zhou2020stochastic} the reduced system matrix is further decomposed into subproblems to improve scalability.
Several open source BA implementations are available, e.g., the SBA package~\cite{lourakis2009sba}, the g2o framework~\cite{kummerle2011g}, or Ceres Solver~\cite{ceres-solver}, which has become a standard tool for solving BA-like problems in both academia and industry.

\paragraph{Nullspace marginalization, square root filters, and QR decomposition}
The concept of nullspace marginalization has been used in contexts other than BA, e.g., for the multi-state constraint Kalman filter~\cite{mourikis2007} and earlier in~\cite{bayard2005estimation}.
\cite{yang2017null} proves the equivalence of nullspace marginalization and the Schur complement in the specific case of robot SLAM.
Several works explicitly point out the advantage of matrix square roots in state estimation~\cite{maybeck1982stochastic,bierman2006factorization,dellaert2006square,wu2015square}, but to the best of our knowledge matrix square roots have not yet been used for problem size reduction in BA.
The QRkit~\cite{svoboda2018qrkit} emphasizes the benefits of QR decomposition for sparse problems in general and also mentions BA as a possible application, but the very specific structure of BA problems and the involved matrices is not addressed.
The orthogonal projector used in the Variable Projection (VarPro) method~\cite{okatani11, hong17} is related to the nullspace marginalization in our approach.
However, VarPro focuses on separable non-linear least squares problems, which do not include standard BA. While~\cite{hong17} mentions the use of QR decomposition to improve numeric stability, we take it one step further by more efficiently multiplying in-place with $Q_2^\top$  rather than explicitly with $I - Q_1Q_1^\top$ (further discussed in Section~\ref{sec:nullspace-marginalization}). This also enables our very efficient way to compute landmark damping (not used in VarPro).
Our landmark blocks can be seen as a specific instance of Smart Factors proposed in~\cite{carlone2014eliminating}, where nullspace projection with explicit SVD decomposition is considered. Instead of the re-triangulation in~\cite{carlone2014eliminating}, we suggest updating the landmarks with back substitution. The factor grouping in~\cite{carlone2014mining} allows to mix explicit and implicit SC. This idea is orthogonal to our approach and could be considered in future work.

\section{QR decomposition}

We briefly introduce the QR decomposition, which can be computed using Givens rotations (see appendix).
For further background, we refer the reader to textbooks on least squares problems (e.g.,~\cite{bjorck1996numerical}).
Let $A$ be an $m\times n$ matrix of full rank with $m \geq n$, i.e., $\operatorname{rank}(A) = n$.
$A$ can be decomposed into an $m \times m$ orthogonal matrix $Q$ and an $m \times n$ upper triangular matrix $R$.
As the bottom ($m - n$) rows of $R$ are zeros, we can partition $R$ and $Q$:
\begin{align}
  A = QR = Q \begin{pmatrix} R_1 \\ 0 \end{pmatrix}
    = \begin{pmatrix} Q_1 & Q_2 \end{pmatrix} \begin{pmatrix} R_1 \\ 0 \end{pmatrix}
    = Q_1 R_1\,,
\end{align}
where $R_1$ is an $n \times n$ upper triangular matrix, $Q_1$ is $m \times n$, and $Q_2$ is $m \times (m-n)$.
Note that this partitioning of $Q$ directly implies that the columns of $Q_2$ form the left nullspace of $A$, i.e., $Q_2^\top A =0$.
Since $Q$ is orthogonal, we have
\begin{align}
\label{eq:q_orth}
    Q^\top Q &= \id_m = QQ^\top\,,
\end{align}
where $\id_m$ is the $m \times m$ identity matrix.
From \eqref{eq:q_orth} we derive:
\begin{gather}
    Q_1^\top Q_1 = \id_n\,, \quad
    Q_2^\top Q_2 = \id_{m-n}\,, \quad
    Q_1^\top Q_2 = 0\,, \\
\label{eq:qqt}
   Q_1Q_1^\top = \id_m - Q_2Q_2^\top\,.
\end{gather}

\section{Square root bundle adjustment}

We assume a very general form of bundle adjustment, similar to~\cite{agarwal2010bundle}:
let $\vect{x}$ be a state vector containing all the optimization variables.
We can subdivide $\vect{x}$ into a pose part $\vect{x}_p$ containing extrinsic and possibly intrinsic camera parameters for all $n_p$ images (index $i$), and a landmark part $\vect{x}_l$ consisting of the 3D coordinates of all $n_l$ landmarks (index $j$).
The total bundle adjustment energy is a sum of squared residuals
\begin{align}
\label{eq:ba_energy}
    E(\vect{x}_p,\vect{x}_l) = \sum_{i}\sum_{j\in O(i)}{r_{ij}(\vect{x}_p,\vect{x}_l)^2} = \Vert \vect{r}(\vect{x}_p,\vect{x}_l) \Vert^2\,,
\end{align}
where $j\in O(i)$ means that landmark $j$ is observed in frame $i$ and $\vect{r}(\vect{x})$ is the concatenation of all residuals $r_{ij}$ into one vector.
We call the total number of residuals $N_r$.
For a pose dimensionality $d_p$, the length of the total state vector $(\vect{x}_p,\vect{x}_l)$ is $d_pn_p+3n_l=:N_p+3n_l$.
Typically, $d_p=6$ if only extrinsic camera parameters need to be estimated, and $d_p=9$ if intrinsic calibration is also unknown.

\begin{figure*}
  \includegraphics[width=\textwidth]{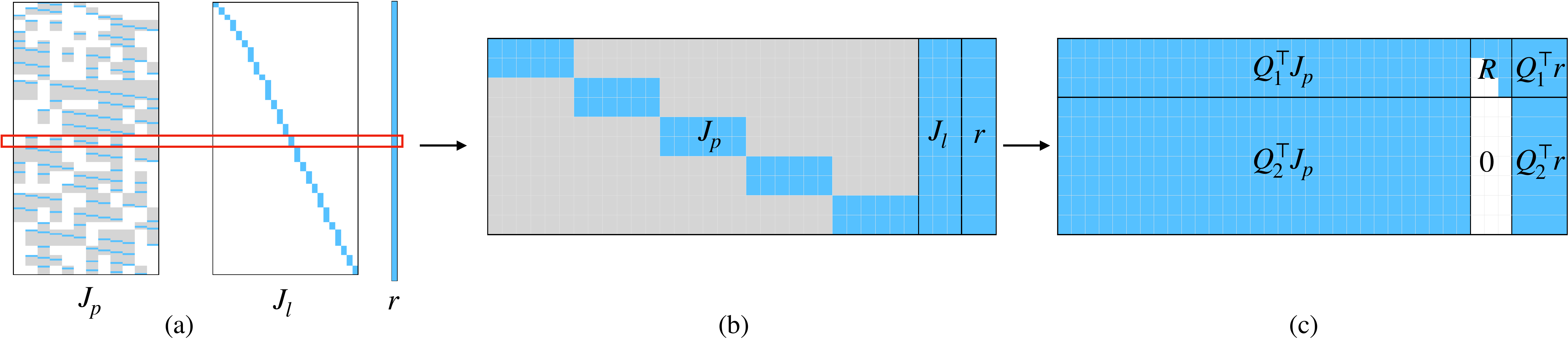}
  \caption{Dense landmark blocks. (a)~Sparsity structure of the pose Jacobian is fixed during the optimization. Non-zero elements shown in blue, potentially non-zero elements after Givens QR shown in gray, and elements that will always stay zero shown in white. (b)~Dense storage for the outlined (red) landmark block that efficiently stores all Jacobians and residuals for a single landmark. (c)~Same landmark block after in-place marginalization. As Givens rotations operate on individual rows,  marginalization can be performed for each landmark block separately, possibly in parallel.}
  \label{fig:sparsity}
\end{figure*}

\subsection{Least squares problem}

The energy in \eqref{eq:ba_energy} is usually minimized by the Levenberg-Marquardt algorithm, which is based on linearizing $\vect{r}(\vect{x})$ around the current state estimate $\vect{x}^0=(\vect{x}_p^0,\vect{x}_l^0)$ and then solving a damped linearized problem
\begin{align}
\begin{aligned}
\label{eq:linearized_residual}
 \min_{\Delta\vect{x}}&\left\Vert \begin{pmatrix}\vect{r}\\0\\0\end{pmatrix} + \begin{pmatrix}J_p & J_l \\ \sqrt{\lambda}D_p & 0 \\ 0 & \sqrt{\lambda}D_l\end{pmatrix}\begin{pmatrix}\Delta\vect{x}_p \\ \Delta\vect{x}_l\end{pmatrix}\right\Vert^2 = \\
    \min_{\Delta\vect{x}_p, \Delta\vect{x}_l}&\bigg(\Vert \vect{r} + \begin{pmatrix}J_p & J_l\end{pmatrix}\begin{pmatrix}\Delta\vect{x}_p \\ \Delta\vect{x}_l\end{pmatrix}\Vert^2  \\
    & + \lambda \Vert D_p \Delta\vect{x}_p \Vert^2 + \lambda\Vert D_l \Delta\vect{x}_l \Vert^2\bigg) \,,
\end{aligned}
\end{align}
with $\vect{r}=\vect{r}(\vect{x}^0)$, $J_p = \left.\frac{\partial \vect{r}}{\partial \vect{x}_p}\right\vert_{\vect{x}^0}$, $J_l = \left.\frac{\partial \vect{r}}{\partial \vect{x}_l}\right\vert_{\vect{x}^0}$, and $\Delta\vect{x}=\vect{x}-\vect{x}^0$.
Here, $\lambda$ is a damping coefficient and $D_p$ and $D_l$ are diagonal damping matrices for pose and landmark variables (often $D^2=\text{diag}(J^\top J)$). 

To simplify notation, in this section we consider the undamped problem (i.e., $\lambda=0$) and discuss the efficient application of damping in Section \ref{sec:lm_damping}.
The undamped problem in \eqref{eq:linearized_residual} can be solved by forming the corresponding normal equation
\begin{align}
\label{eq:normal_eq}
\begin{pmatrix}
H_{pp} & H_{pl} \\
H_{lp} & H_{ll}
\end{pmatrix}
\begin{pmatrix}
-\Delta \vect{x}_{p}  \\
-\Delta \vect{x}_{l}
\end{pmatrix}
 &= 
 \begin{pmatrix}
\vect{b}_{p}  \\
\vect{b}_{l}
\end{pmatrix},
\end{align}
where
\begin{align}
H_{pp} &= J_p^\top J_p\,, \quad
H_{ll} = J_l^\top J_l\,, \\
H_{pl} &= J_p^\top J_l = H_{lp}^\top\,, \\
%\label{eq:b_pose}
\vect{b}_{p} &= J_p^\top \vect{r}\,, \quad
%\label{eq:b_landmark}
\vect{b}_{l} = J_l^\top \vect{r}\,.
\end{align}
The system matrix $H$ of this problem is of size $(N_p+3n_l)^2$, which can become impractically large (millions of rows and columns) for problems like those in~\cite{agarwal2010bundle} (see Figure~\ref{fig:teaser} for an example).

\subsection{Schur complement (SC)}

A very common way to solve \eqref{eq:normal_eq} is by applying the Schur complement trick (see e.g.,~\cite{brown1958solution,agarwal2010bundle,wu2011multicore}):
we form the RCS
\begin{align}
\label{eq:red_normal_eq}
\tilde{H}_{pp} (- \Delta\vect{x}_p) = \tilde{\vect{b}}_p\,,
\end{align}
with
\begin{align}
\tilde{H}_{pp} &= H_{pp} - H_{pl} H_{ll}^{-1} H_{lp}\,, \\
\tilde{\vect{b}}_p &= \vect{b}_p - H_{pl} H_{ll}^{-1}\vect{b}_l\,.
\end{align}
The solution $\Delta\vect{x}_p^*$ of \eqref{eq:red_normal_eq} is the same as the pose component of the solution of \eqref{eq:normal_eq}, but now the system matrix has a much more tractable size of $N_p^2$, which is usually in the order of thousands $\times$ thousands.
Note that as $H_{ll}$ is block-diagonal with blocks of size $3\times 3$, the multiplication with $H_{ll}^{-1}$ is cheap.
Given an optimal pose update $\Delta\vect{x}_p^*$, the optimal landmark update is found by back substitution
\begin{align}
\label{eq:backsub_sc}
    - \Delta\vect{x}_{l}^* &=  H_{ll}^{-1} (\vect{b}_{l} - H_{lp} (-\Delta\vect{x}_{p}^*))\,.
\end{align}

\subsection{Nullspace marginalization (NM)}
\label{sec:nullspace-marginalization}

Using QR decomposition on $J_l=QR$, and the invariance of the L2 norm under orthogonal transformations, we can rewrite the term in \eqref{eq:linearized_residual}:
\begin{align}
\label{eq:q_linearized_residual}
\begin{aligned}
 &\Vert \vect{r} + \begin{pmatrix}J_p & J_l\end{pmatrix}\begin{pmatrix}\Delta\vect{x}_p \\ \Delta\vect{x}_l\end{pmatrix}\Vert^2 \\
 &= \Vert Q^\top\vect{r} + \begin{pmatrix}Q^\top J_p & Q^\top J_l\end{pmatrix}\begin{pmatrix}\Delta\vect{x}_p \\ \Delta\vect{x}_l\end{pmatrix}\Vert^2 \\
 &= \Vert Q_1^\top\vect{r} + Q_1^\top J_p\Delta\vect{x}_p + R_1\Delta\vect{x}_l\Vert^2 \\
 &\qquad + \Vert Q_2^\top\vect{r} + Q_2^\top J_p\Delta\vect{x}_p\Vert^2\,.
\end{aligned}
\end{align}
As $R_1$ is invertible, for a given $\Delta\vect{x}_p^*$, the first term can always be set to zero (and thus minimized) by choosing
\begin{align}
\label{eq:backsub_nm}
    \Delta\vect{x}_l^* = -R_1^{-1} (Q_1^\top \vect{r} + Q_1^\top J_p \Delta\vect{x}_{p}^*)\,.
\end{align}
So \eqref{eq:linearized_residual} reduces to minimizing the second term in \eqref{eq:q_linearized_residual}:
\begin{align}
\min_{\Delta\vect{x}_p}{\Vert Q_2^\top \vect{r} + Q_2^\top J_p \Delta \vect{x}_{p} \Vert^2}\,.
\label{eq:reduced}
\end{align}
Again, this problem is of significantly smaller size than the original one.
However, as opposed to the (explicit) Schur complement trick, we do not explicitly have to form the Hessian matrix.

\begin{figure*}
  \includegraphics[width=\textwidth]{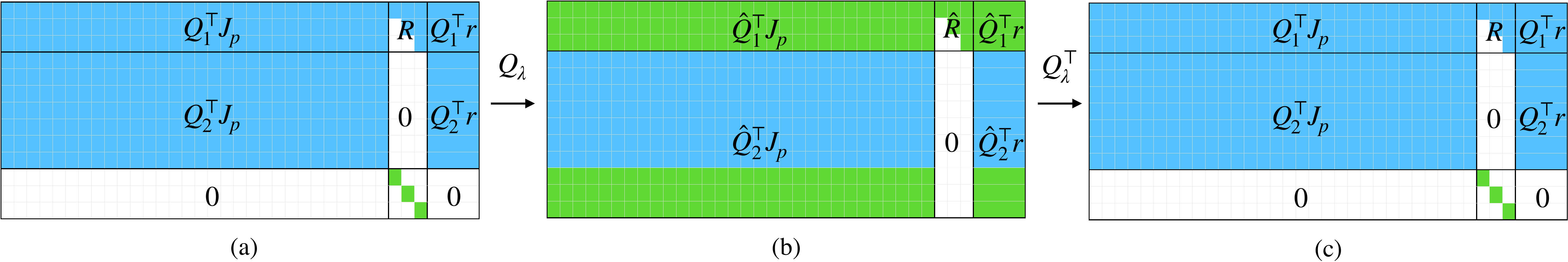}
  \caption{Illustration of the landmark damping in the Levenberg-Marquardt optimization. (a) We add three zero rows with diagonal damping for landmarks to the marginalized landmark block. (b) With 6 Givens rotations we eliminate the values on diagonal, which gives us a new landmark block with marginalized out landmark. (c) By applying the transposed rotations in reverse order and zeroing out the diagonal we can bring the landmark block to the original state. Zero entries of the landmark block are shown in white, parts that change are shown in green, and parts that stay unchanged are shown in blue.}
  \label{fig:damping}
\end{figure*}

\subsection{Equivalence of SC and NM}

With the QR decomposition $J_l = Q_1 R_1$ used in the last paragraphs, we get
\begin{align}
H_{pp} &= J_p^\top J_p\,, \\
H_{pl} &= J_p^\top Q_1 R_1\,, \\
H_{ll} &= R_1^\top Q_1^\top Q_1 R_1 = R_1^\top R_1\,, \\
\vect{b}_{p} &= J_p^\top \vect{r}\,, \\
\vect{b}_{l} &= R_1^\top Q_1^\top \vect{r}\,. \
\end{align}
Using this, we can rewrite the Schur complement matrix $\tilde{H}_{pp}$ and vector $\tilde{\vect{b}}_p$ and simplify with \eqref{eq:qqt}:
\begin{align}
&\begin{aligned}
\label{eq:red_H}
\tilde{H}_{pp} &= H_{pp} - J_p^\top Q_1 R_1 (R_1^\top R_1)^{-1} R_1^\top Q_1^\top J_p \\
&= H_{pp} - J_p^\top (\id_{N_r} - Q_2 Q_2^\top) J_p \\
&= J_p^\top Q_2 Q_2^\top J_p\,,
\end{aligned}\\
&\begin{aligned}
\label{eq:red_b}
\tilde{\vect{b}}_p &= \vect{b}_p - J_p^\top Q_1 R_1 (R_1^\top R_1)^{-1} R_1^\top Q_1^\top \vect{r} \\
&= J_p^\top Q_2 Q_2^\top \vect{r}\,.
\end{aligned}
\end{align}
Thus, the SC-reduced equation is nothing but the normal equation of problem \eqref{eq:reduced}, which proves the algebraic equivalence of the two marginalization techniques. 
Additionally, we can show that the equations for back substitution for Schur complement (\ref{eq:backsub_sc}) and nullspace marginalization (\ref{eq:backsub_nm}) are also algebraically equivalent:
 \begin{align}
 \begin{aligned}
  \Delta\vect{x}_{l}^* &=  -H_{ll}^{-1} (\vect{b}_{l} + H_{lp} \Delta\vect{x}_{p}^*) \\
  &=  -(R_1^\top R_1)^{-1} (R_1^\top Q_1^\top \vect{r} + (J_p^\top Q_1 R_1)^\top \Delta\vect{x}_{p}^*) \\
  &=  -R_1^{-1} (Q_1^\top \vect{r} + Q_1^\top J_p \Delta\vect{x}_{p}^*)\,.
 \end{aligned}
 \end{align}

Note that the above arguments also hold for the damped problem \eqref{eq:linearized_residual}, the difference being that the Hessian will have an augmented diagonal and that the matrices in the QR decomposition will have a larger size.

\section{Implementation details}

Bundle adjustment is a very structured problem, so we can take advantage of the problem-specific matrix structures to enable fast and memory-efficient computation.

\subsection{Storage}

We group the residuals by landmarks, such that $J_l$ has block-sparse structure, where each block is $2k_j\times 3$ with $k_j$ the number of observations for a particular landmark, see Figure~\ref{fig:sparsity} (a).
As each landmark is only observed by a subset of cameras, the pose Jacobian $J_p$ is also sparse.

We group the rows corresponding to each landmark and store them in a separate dense memory block, which we name a \emph{landmark block}.
We store only the blocks of the pose Jacobian that correspond to the poses where the landmark was observed, because all other blocks will always be zero. For convenience we also store the landmark's Jacobians and residuals in the same landmark block, as shown in Figure~\ref{fig:sparsity} (b).
This storage scheme can be applied both to the undamped and the damped Jacobian (see Section \ref{sec:lm_damping} for damping).

\subsection{QR decomposition}

Applying a sequence of Givens rotations in-place transforms the landmark block to the marginalized state shown in Figure~\ref{fig:sparsity} (c). The bottom part corresponds to the reduced camera system, and the top part can be used for back substitution. This transformation can be applied to each block independently, possibly in parallel.
We never have to explicitly store or compute the matrix $Q$; we simply apply the sequence of Givens rotations to the landmark block one by one, as they are computed.
Note that alternatively we can use three Householder reflections per landmark block, with which we noticed a minor improvement in runtime.

\subsection{Levenberg-Marquardt damping}
\label{sec:lm_damping}

The augmentation of the Jacobians by diagonal matrices as used in \eqref{eq:linearized_residual} consists of two parts that we treat differently to optimally exploit the nature of the BA problem in our implementation.

\paragraph{Landmark damping}
We first look at damping the landmark variables:
rather than actually attaching a large diagonal matrix $\sqrt{\lambda}D_l$ to the full landmark Jacobian $J_l$, we can again work on the landmark block from Figure~\ref{fig:sparsity} (b) and only attach a $3\times 3$ sub-block there, 
see Figure~\ref{fig:damping} (a) and (b).
To simplify the expressions in figures, we slightly abuse notation when considering a single landmark and denote the corresponding parts of $J_p$, $J_l$ and $r$ in the landmark block by the same symbols. %$J_p$, $J_l$ and $r$ as well.
The matrices involved in the QR decomposition of the undamped system are $Q_1$, $Q_2$, $R_1$ and those for the damped system are marked with a hat.
Note that $Q$ and $\hat{Q}$ are closely related; the additional three diagonal entries in the damped formulation can be zeroed using only six Givens rotations, such that
\begin{align}
\hat{Q} = \begin{pmatrix}\hat{Q}_1 & \hat{Q}_2  \end{pmatrix} = \begin{pmatrix}Q_1 & Q_2  & 0 \\ 0 & 0 & \id_{3} \end{pmatrix}Q_\lambda
\,,
\end{align}
where $Q_\lambda$ is a product of six Givens rotations.
Thus, applying and removing landmark damping is computationally cheap: we apply the Givens rotations one by one and store them individually (rather than their product $Q_\lambda$) to undo the damping later.
Figure~\ref{fig:damping} illustrates how this can be done in-place on the already marginalized landmark block.
This can speed up LM's backtracking, where a rejected increment is recomputed with the same linearization, but with increased damping.
By contrast, for an explicit SC solver, changing the damping would mean recomputing the Schur complement from scratch.

\paragraph{Pose damping}
Marginalizing landmarks using Givens rotations in the damped formulation of \eqref{eq:linearized_residual} does not affect the rows containing pose damping.
Thus, it is still the original diagonal matrix $\sqrt{\lambda}D_p$ that we append to the bottom of $\hat{Q}^\top J_p$:
\begin{align}
    \hat{J}_p = 
    \begin{pmatrix}
    \hat{Q}_1^\top J_p \\ \hat{Q}_2^\top J_p \\ \sqrt{\lambda}D_p
    \end{pmatrix}\,.
\end{align}
In practice, we do not even have to append the block, but can simply add the corresponding summand when evaluating matrix-vector multiplication for the CG iteration \eqref{eq:cg_mult}.

\subsection{Conjugate gradient on normal equations}

To solve for the optimal $\Delta x_p^*$ in small and medium systems, we could use dense or sparse QR decomposition of the stacked $Q_2^{\top} J_p$ from landmark blocks to minimize the linear least squares objective $\Vert Q_2^{\top} J_p \Delta x_p + Q_2^{\top} r\Vert^2$. However, for large systems this approach is not feasible due to the high computational cost.
Instead, we use CG on the normal equations as proposed in~\cite{agarwal2010bundle}.
Other iterative solvers like LSQR~\cite{paige1982lsqr} or LSMR~\cite{fong2011lsmr} that can be more numerically stable than CG turned out to not improve stability for the case of bundle adjustment~\cite{byrod2010conjugate}.

CG accesses the normal equation matrix $\tilde{H}_{pp}$ only by multiplication with a vector $\vect{v}$, which we can write as
\begin{align}
\label{eq:cg_mult}
\tilde{H}_{pp} ~ v = (\hat{Q}_2^\top J_p)^\top (\hat{Q}_2^\top J_p ~ v) + \lambda D^2_p ~ v
\,.
\end{align}
This multiplication can be efficiently implemented and well parallelized using our array of landmark blocks.
Thus, we do not need to explicitly form the normal equations for the reduced least squares problem.

Still, the CG part of our solver has the numerical properties of the normal equations (squared condition number compared to the marginal Jacobian $\hat{Q}_2^\top J_p$). To avoid numeric issues when using single-precision floating-point numbers, we scale the columns of the full Jacobian to have unit norm and use a block-Jacobi preconditioner, both standard procedures when solving BA problems and both also used in the other evaluated solvers.
We also note that with the Levenberg-Marquardt algorithm, we solve a strictly positive definite damped system, which additionally improves the numeric stability of the optimization.

Storing the information used in CG in square root form allows us to make sure that $\tilde{H}_{pp}$ is always strictly positive definite. As we show with our experiments (see Section \ref{sec:analysis}), for many sequences small round-off errors during SC (explicit or implicit) render $\tilde{H}_{pp}$ to be numerically indefinite with single-precision floating-point computations.

With the computed $\Delta\vect{x}_{p}^*$ we can do back substitution for each individual landmark block independently and in parallel. We already have all the necessary information ($\hat{Q}_1^\top J_p$, $\hat{R}$, $\hat{Q}_1^\top r$) stored in the landmark blocks 
after marginalization.

\subsection{Parallel implementation}
As pointed out above, the linearization, marginalization, and back substitution can be computed independently for each landmark block. There is no information shared between landmark blocks, so we can use a simple \emph{parallel for} loop to evenly distribute the workload between all available CPU cores.
The matrix-vector multiplications that constitute the most computationally expensive part of CG can also be efficiently parallelized. In this case, multiplication results of individual landmark blocks have to be summed, so we employ the common \emph{parallel reduce} pattern.
How effective these simple parallelization schemes are is underlined by our evaluation, which shows excellent runtime performance of the square root bundle adjustment implementation, compared to both our custom and Ceres' SC solvers.

\begin{figure*}
    \centering
    \includegraphics[width=0.98\textwidth]{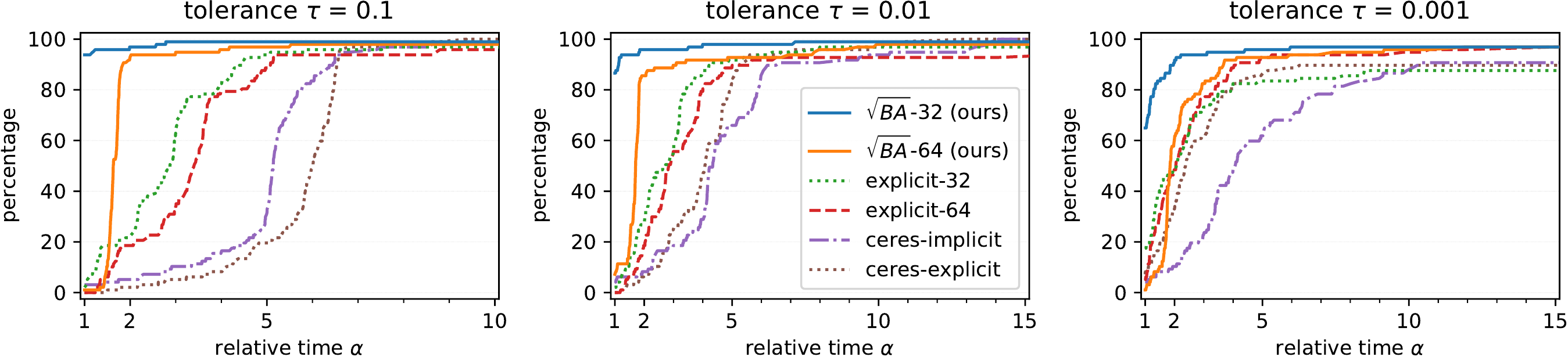}
    \caption{Performance profiles for all BAL datasets show percentage of problems solved to a given accuracy tolerance $\tau$ with increasing relative runtime $\alpha$.
    Our proposed $\sqrt{BA}$ solver outperforms other methods across all accuracy tolerances.
    In single precision, the solver is about twice as fast as with double, but does not lose in accuracy, underpinning the favorable numerical properties of the square root formulation.
    In contrast, while the SC solver in double precision is equally accurate, this is not the case for the single-precision variant. Keep in mind that Ceres is not as tailored to the exact problem setup as our custom implementation, possibly explaining its slower performance. Note that the performance profiles are cut off at the right side of the plots to show only the most relevant parts.
    }
    \label{fig:performance_all}
\end{figure*}

\section{Experimental evaluation}

\subsection{Algorithms and setup}

\begin{table}
\setlength{\tabcolsep}{0.3em}
\centering
\begin{tabular}{l|c;{1pt/1pt}c;{1pt/1pt}c;{1pt/1pt}c|c;{1pt/1pt}c|}
& \rotatebox{90}{\emph{$\sqrt{BA}$-32} (ours)} 
& \rotatebox{90}{\emph{$\sqrt{BA}$-64} (ours)} 
& \rotatebox{90}{\emph{explicit-32}} 
& \rotatebox{90}{\emph{explicit-64}}
& \rotatebox{90}{\emph{ceres-implicit}}
& \rotatebox{90}{\emph{ceres-explicit}}
\\
\hline
solver implementation  & \multicolumn{4}{c|}{custom} & \multicolumn{2}{c|}{Ceres} \\
\hline
float precision  & s & d & s & d & \multicolumn{2}{c|}{d} \\
\hline
landmark marginalization  & \multicolumn{2}{c;{1pt/1pt}}{NM} & \multicolumn{2}{c|}{SC} & \multicolumn{2}{c|}{SC} \\
\hline
RCS storage  & \multicolumn{2}{c;{1pt/1pt}}{LMB} & \multicolumn{2}{c|}{H} & -- & H\\
\hline
\end{tabular}
\vspace{0.3cm}
\caption{The evaluated solvers---proposed and baseline---are implemented either completely in our custom code base, or using Ceres, with single (s) or double (d) floating-point precision, using Nullspace (NM) or Schur complement (SC)-based marginalization of landmarks, and storing the reduced camera system sparsely in landmark blocks (LMB), sparsely as a reduced Hessian (H), or not at all (--).
}
\label{tab:solver_properties}
\end{table}

We implement our $\sqrt{BA}$ solver in C++ in single (\emph{$\sqrt{BA}$-32}) and double (\emph{$\sqrt{BA}$-64}) floating-point precision
and compare it to the methods proposed in \cite{agarwal2010bundle}
as implemented in Ceres Solver \cite{ceres-solver}.
This solver library is popular in the computer vision and robotics community, 
since it is mature, performance-tuned, and offers many linear solver variations.
That makes it a relevant and challenging baseline to benchmark our implementation against.
While Ceres is a general-purpose solver,
it is very good at exploiting the specific problem structure as
it was built with BAL problems in mind.
Our main competing algorithms are Ceres' sparse Schur complement solvers, 
which solve the RCS iteratively by either explicitly saving $\tilde{H}_{pp}$ in memory as a block-sparse matrix (\emph{ceres-explicit}), 
or otherwise computing it on the fly during the iterations (\emph{ceres-implicit}).
In both cases, the same block-diagonal of $\tilde{H}_{pp}$ that we use in $\sqrt{BA}$ is used as preconditioner.
As the bottleneck is not computing Jacobians, but marginalizing points and the CG iterations, we use analytic Jacobians for our custom solvers and Ceres' exact and efficient autodiff with dual numbers.
For an even more direct comparison,
we additionally implement the sparse iterative explicit Schur complement solver without Ceres, sharing much of the code with our $\sqrt{BA}$ implementation. While Ceres always uses double precision, 
we use our custom implementation to evaluate numerical stability by considering single (\emph{explicit-32}) and double (\emph{explicit-64}) precision.
Table~\ref{tab:solver_properties} summarizes the evaluated configurations.

For Ceres we use default options, unless otherwise specified.
This includes the scaling of Jacobian columns to avoid numerical issues~\cite{agarwal2010bundle}.
Just like in our custom implementation, 
we configure the number of threads to be equal to the number of (virtual) CPU cores. 
Our Levenberg-Marquardt loop is in line with Ceres: 
starting with initial value $10^{-4}$, we update the damping factor $\lambda$ according to the ratio of actual and expected cost reduction,
and run it for at most 50 iterations, terminating early if a relative function tolerance of $10^{-6}$ is reached. 
In the inner CG loop we use the same forcing sequence as Ceres, with a maximum of 500 iterations and no minimum.
We run experiments on an Ubuntu 18.04 desktop with 64GB RAM and an Intel Xeon W-2133 with 12 virtual cores at 3.60GHz.
In our own solver implementation we rely on Eigen \cite{eigenweb} for dense linear algebra and TBB\cite{tbbweb} for 
simple \emph{parallel for} and \emph{parallel reduce} constructs.

\begin{figure*}
    \centering
    \begin{tabular}{c@{\hskip 0.005\textwidth}c@{\hskip 0.005\textwidth}c@{\hskip 0.005\textwidth}c}
        \includegraphics[width=0.24\textwidth]{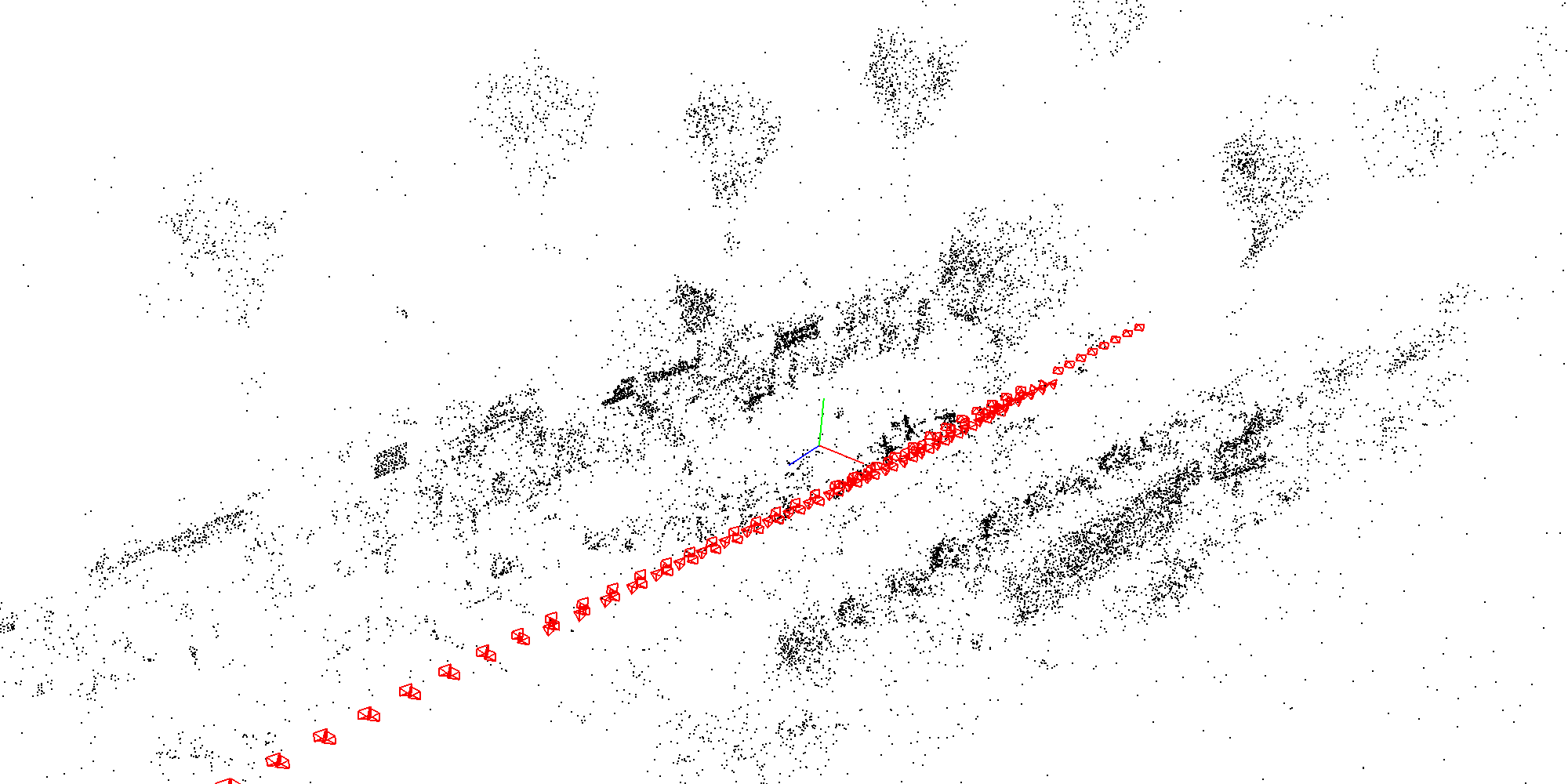} & 
        \includegraphics[width=0.24\textwidth]{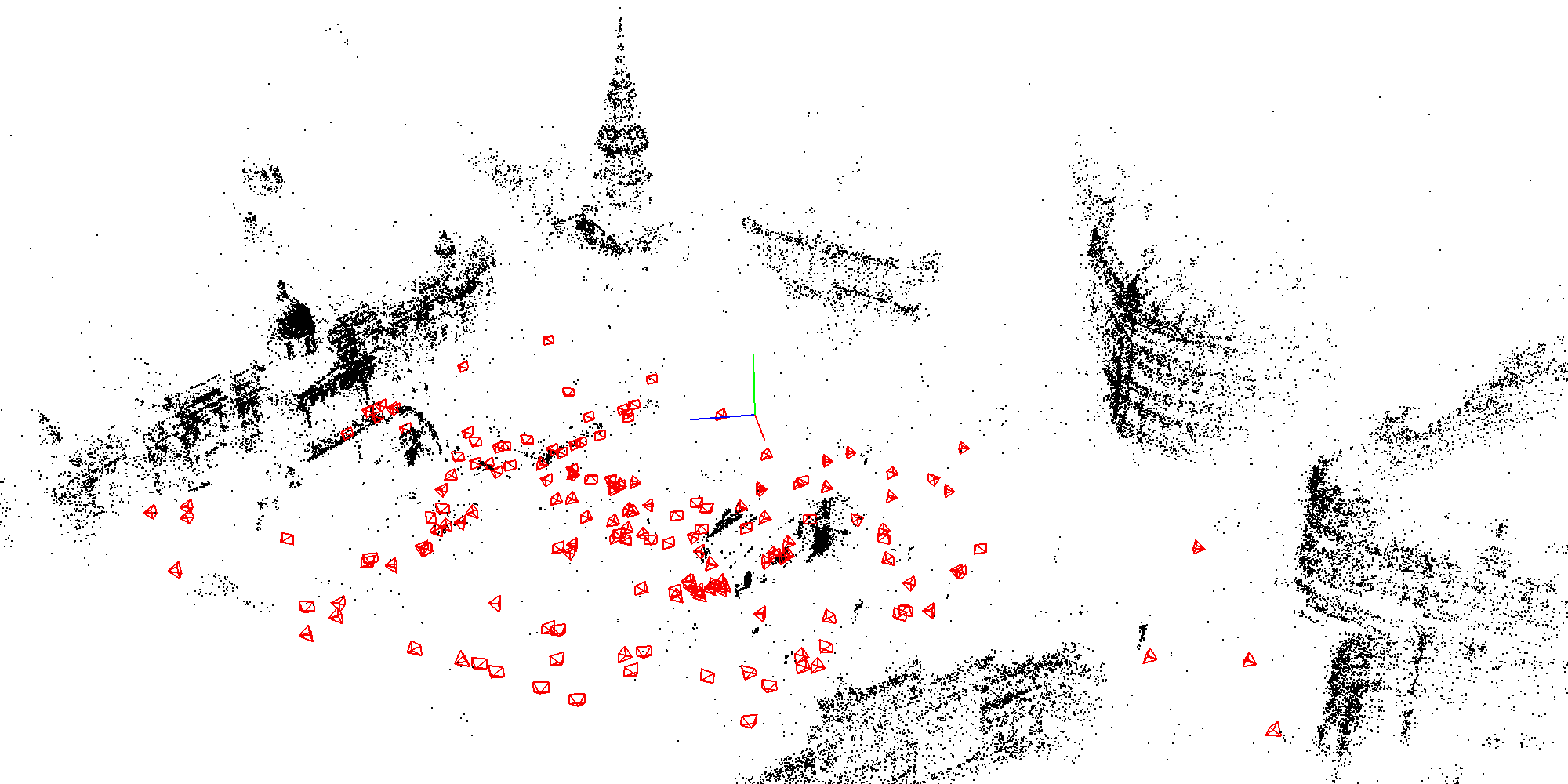} & 
        \includegraphics[width=0.24\textwidth]{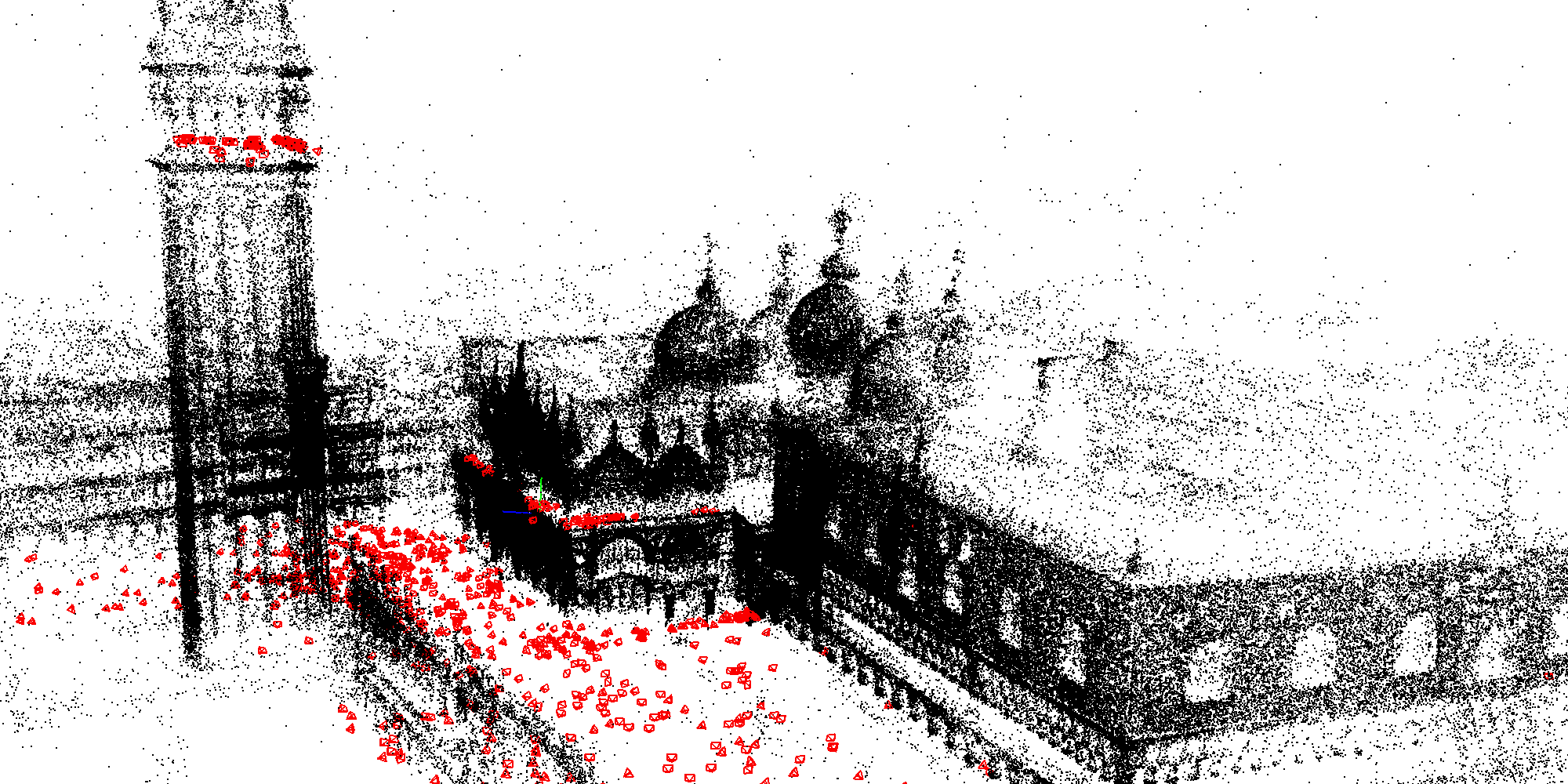} & 
        \includegraphics[width=0.24\textwidth]{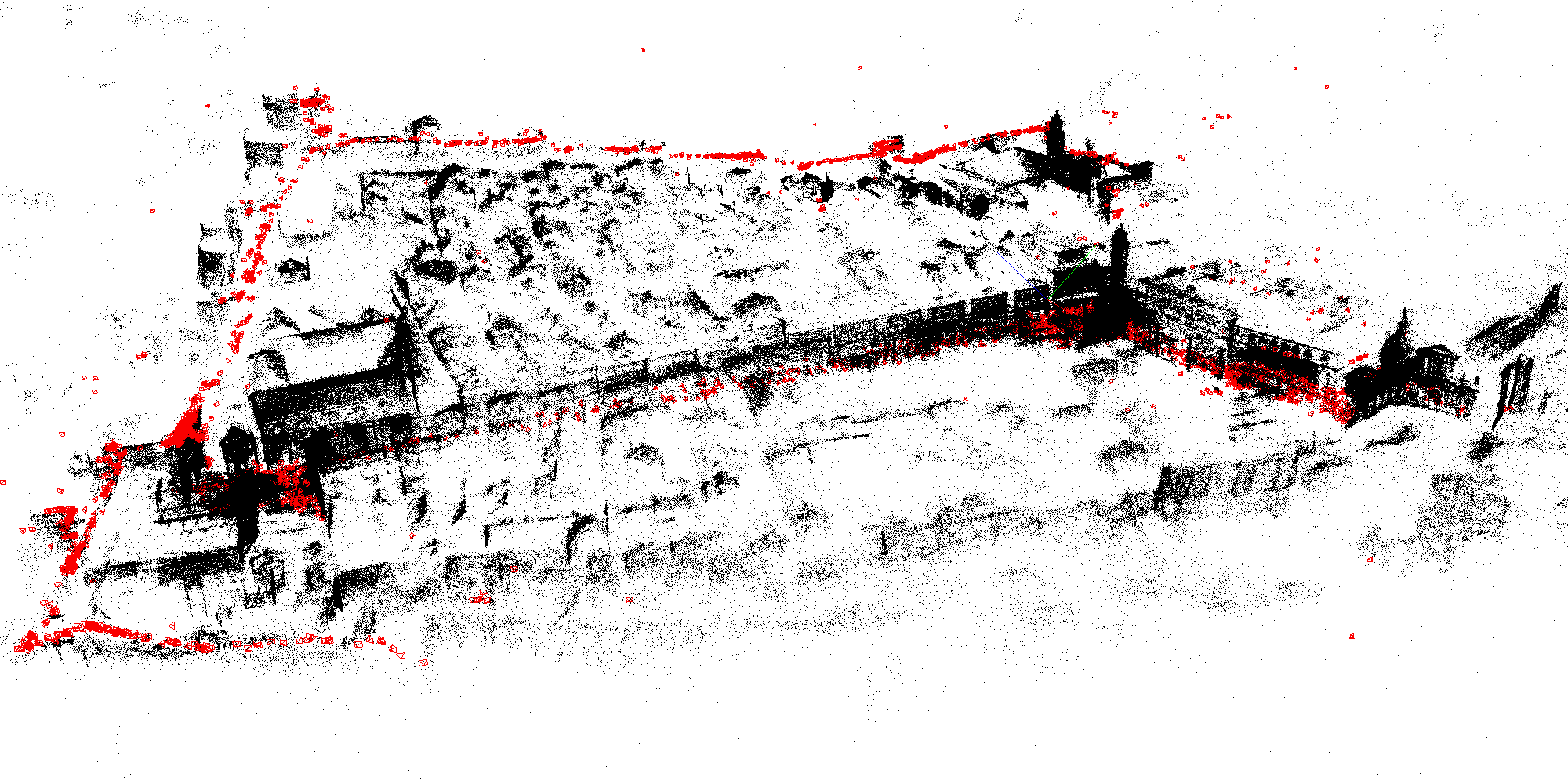} \\
        \includegraphics[width=0.24\textwidth]{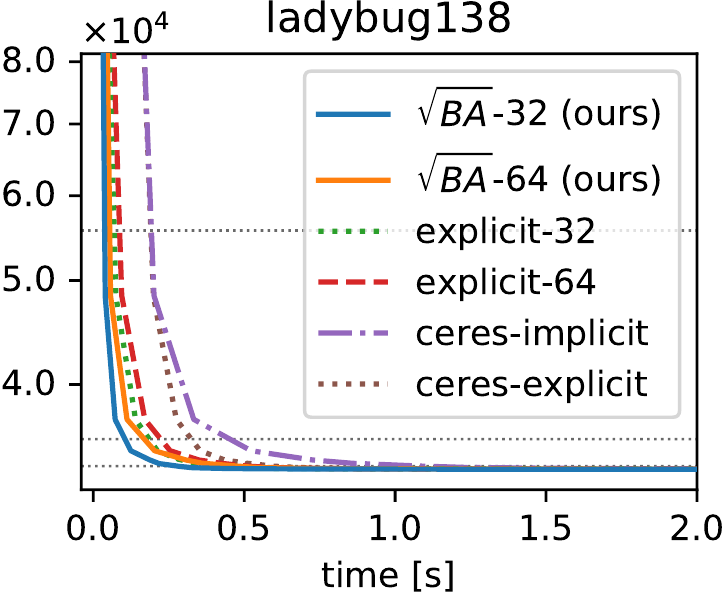} & 
        \includegraphics[width=0.24\textwidth]{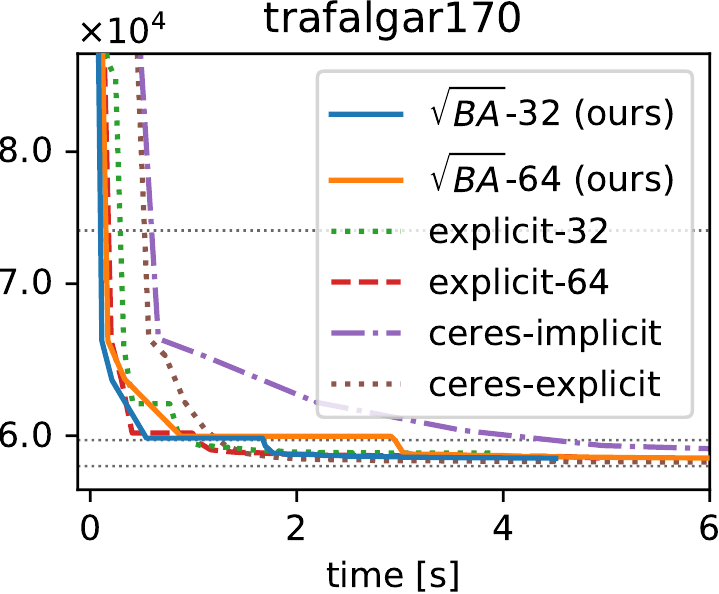} & 
        \includegraphics[width=0.24\textwidth]{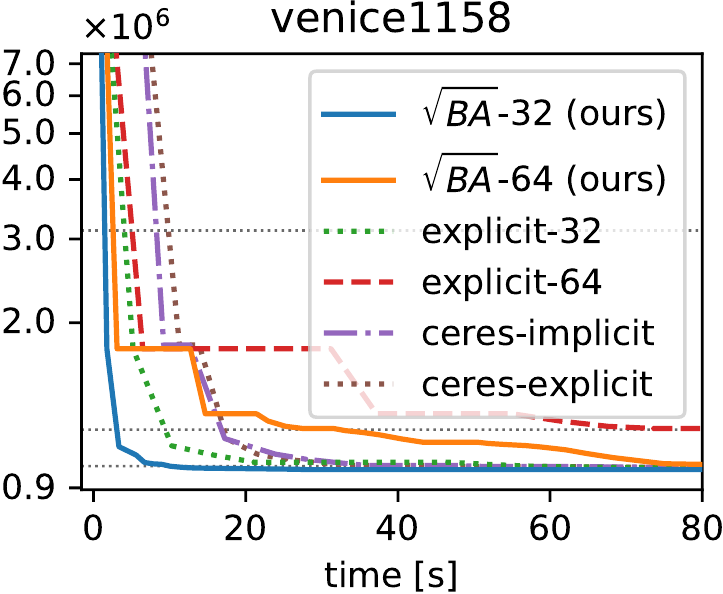} & 
        \includegraphics[width=0.24\textwidth]{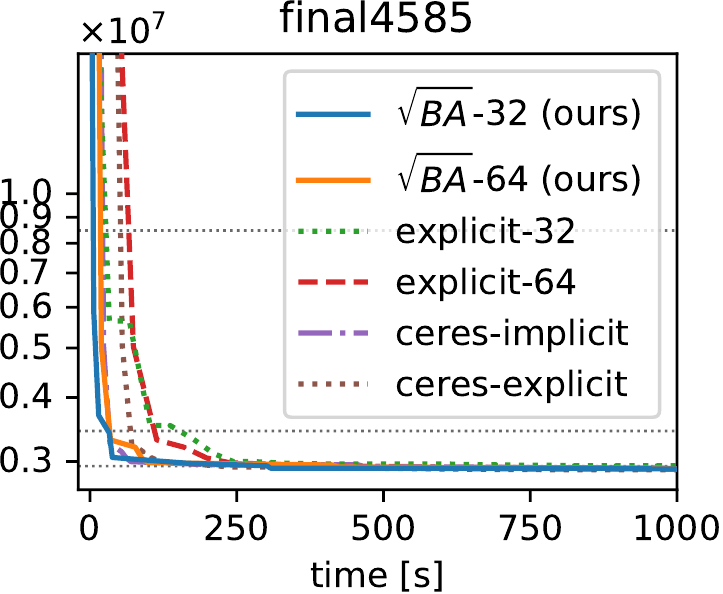}
    \end{tabular}
    \caption{Convergence plots from small to large problems and rendered optimized landmark point clouds. The $y$-axes show the total BA cost (log scale), and the horizontal lines indicate cost thresholds for the tolerances $\tau \in \{ 10^{-1}, 10^{-2}, 10^{-3}\}$.}
    \label{fig:convergence}
\end{figure*}

\subsection{Datasets}

For our extensive evaluation we use all 97 bundle adjustment problems from the BAL \cite{agarwal2010bundle} project page.
These constitute initialized bundle adjustment problems and come in different groups: the \emph{trafalgar}, \emph{dubrovnik}, and \emph{venice} problems originate from successive iterations in a skeletal SfM reconstruction of internet image collections \cite{svoboda2018qrkit}. 
They are combined with additional leaf images, which results in the thus denser \emph{final} problems.
The \emph{ladybug} sequences are reconstructions from a moving camera, but despite this we always model all camera intrinsics as independent, 
using the suggested \emph{Snavely} projection model with one focal length and two distortion parameters. Figure~\ref{fig:convergence} visualizes some exemplar problems after they have been optimized.

As is common, we apply simple gauge normalization as preprocessing:
we center the point cloud of landmarks at the coordinate system origin and rescale to median absolute deviation of 100.
Initial landmark and camera positions are then perturbed with small Gaussian noise.
To avoid close-to-invalid state, we remove all observations with a small or negative $z$ value in the camera frame, 
completely removing landmarks with less than two remaining observations.
We additionally employ the Huber norm with a parameter of 1~pixel for residuals (implemented with IRLS as in Ceres).
This preprocessing essentially follows Ceres' BAL examples. It is deterministic and identical for all solvers by using a fixed random seed and being computed on state in double precision, regardless of solver configuration.

\subsection{Performance profiles}

When evaluating a solver, we are primarily interested in accurate optimization results.
Since we do not have independent ground truth of correct camera positions, intrinsics, and landmark locations,
we use the bundle adjustment cost as a proxy for accuracy. Lower cost in general corresponds to better solutions.
But depending on the application, we may desire in particular low runtime, which can be a trade-off with accuracy.
The difficulty lies in judging the performance across several orders of magnitudes in problem sizes, cost values, and runtimes (for the BAL datasets the number of cameras $n_p$ ranges from 16 to 13682).
As proposed in prior work \cite{kushal2012visibility,dolan2002benchmarking}, we therefore use performance profiles to evaluate both accuracy and runtime jointly.
The performance profile of a given solver maps the relative runtime~$\alpha$ (relative to the fastest solver for each problem and accuracy) to the percentage of problems solved to accuracy~$\tau$.
The curve is monotonically increasing, 
starting on the left with the percentage of problems for which the solver is the fastest, 
and ending on the right with the percentage of problems on which it achieves the accuracy~$\tau$ at all.
The curve that is more to the left indicates better runtime and the curve that is more to the top indicates higher accuracy.
A precise definition of performance profiles is found in the appendix.

\subsection{Analysis}
\label{sec:analysis}

\begin{figure}
    \centering
    \includegraphics[width=0.9\columnwidth]{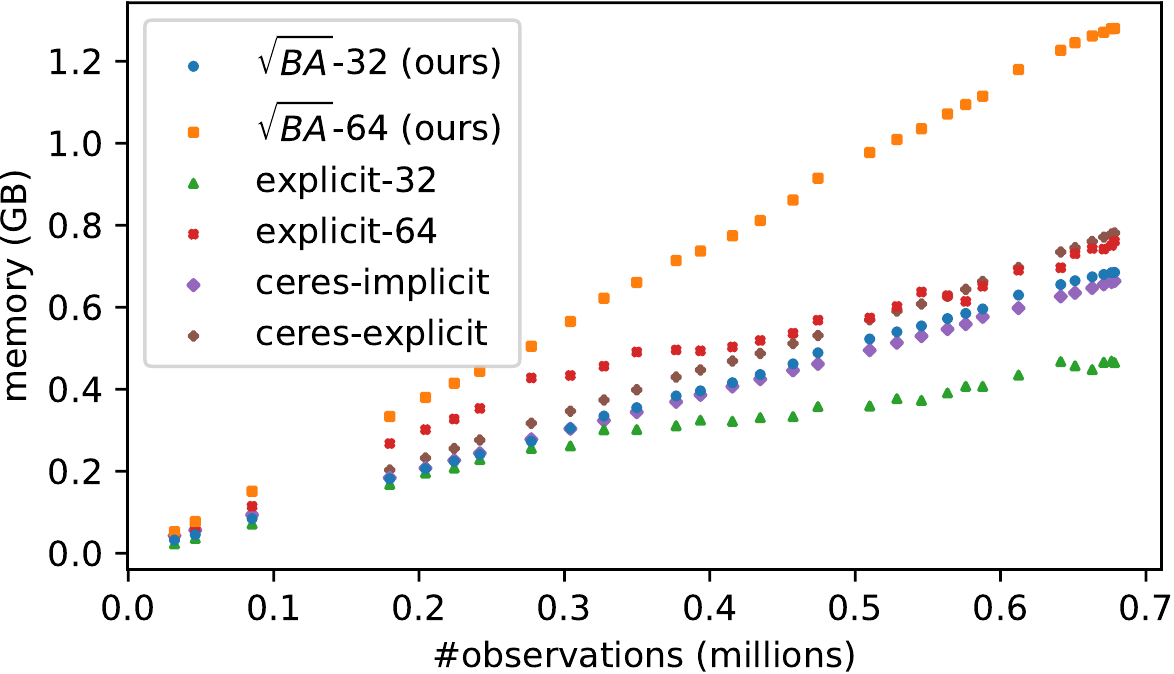}
    \caption{Memory consumption for the relatively sparse \emph{ladybug} problems grows linearly with the problem size. The number of cameras here ranges from 49 to 1723.}
    \label{fig:memory_ladybug}
\end{figure}

Figure~\ref{fig:performance_all} shows the performance profiles with all BAL datasets for a range of tolerances $\tau \in \{ 10^{-1}, 10^{-2}, 10^{-3}\}$.
We can see that our proposed square root bundle adjustment solver \emph{$\sqrt{BA}$-64} is very competitive, yielding better accuracy than some SC-based iterative solvers, and often at a lower runtime.
\emph{$\sqrt{BA}$-32} is around twice as fast and equally accurate, which highlights the good numerical stability of the square root formulation. It clearly outperforms all other solvers across all tolerances.
The fact that \emph{\mbox{explicit-32}} does not always reach the same accuracy as its double precision counterpart \emph{explicit-64} indicates that SC-based solvers do not exhibit the same numerical stability. We also do not observe the same twofold speedup, which is related to the fact that \emph{\mbox{explicit-32}} does have to backtrack in the LM loop significantly more often to increase damping when the Schur complement matrix becomes indefinite due to round-off errors. This happens at least once for 84 out of the 97 problems with \emph{explicit-32} and even with \emph{explicit-64} for 7 of the problems. With $\sqrt{BA}$, we have never encountered this.

A similar conclusion can be drawn from the convergence plots in Figure~\ref{fig:convergence}, which show a range of differently sized problems.
For the small \emph{ladybug} as well as the medium and large \emph{skeletal} problems, our solver is faster. 
Even on the large and much more dense \emph{final4585}, the square root solver is competitive. In the square root formulation memory and thus to some degree also runtime grows larger for denser problems---in the sense of number of observations per landmark---since a specific landmark block grows quadratically in size with the number of its observations. This is in contrast to density in the sense of number of cameras co-observing at least one common landmark, as for the SC.
Still, across all BAL datasets, only for the largest problem, \emph{final13682}, where the landmarks have up to 1748 observations, does \emph{$\sqrt{BA}$-32} run out of memory. % on our machine. 
For sparse problems, such as \emph{ladybug}, one can see in Figure~\ref{fig:memory_ladybug} that the memory grows linearly with the  number of landmarks, and for \emph{$\sqrt{BA}$-32} is similar to Ceres' iterative SC solvers.
In summary, while for very small problems we expect direct solvers to be faster than any of the iterative solvers, and for very large and dense problems implicit SC solvers scale better due to their memory efficiency~\cite{agarwal2010bundle}, the proposed \emph{$\sqrt{BA}$} solver outperforms alternatives for medium to large problems, i.e., the majority of the BAL dataset.

\section{Conclusion}

We present an alternative way to solve large-scale bundle adjustment that marginalizes landmarks without having to compute any blocks of the Hessian matrix.
Our square root approach $\sqrt{BA}$ displays several advantages over the standard Schur complement, in terms of speed, accuracy, and numerical stability.
We have combined a very general theoretical derivation of nullspace marginalization with a tailored implementation that maximally exploits the specific structure of BA problems.
Experiments comparing our solver to both a custom SC implementation and the state-of-the-art Ceres library show how $\sqrt{BA}$ can handle single-precision floating-point operations much better than the Schur complement methods, outperforming all evaluated competing approaches. We see great potential in $\sqrt{BA}$ to benefit other applications that play up to its strong performance on sparse problems, for example incremental SLAM.

\balance

{\small
\bibliographystyle{ieee_fullname}
\bibliography{references}
}

\pagebreak

\twocolumn[
\begin{center}
{\Large \bf Supplementary Material}
\end{center}
\vspace{2em}
]

\appendix

In this supplementary material we provide additional background, analysis of computational complexity, and a detailed account of the convergence of all solvers for each of the 97 problems from the BAL dataset used in our evaluation. 
The latter are the same experiments that are aggregated in performance profiles in Figure~4 of the main paper.
Section~\ref{sec:appendix_details} includes a concise introduction to Givens rotations and a definition of performance profiles.
Section~\ref{sec:complexities} discusses the computational complexity of our QR-based solver compared to the explicit and implicit Schur complement solvers.
Section~\ref{sec:problem_sizes} presents the size and density of all problems (tabulated in Section~\ref{sec:problem_sizes_table}).
Finally, Section~\ref{sec:convergence} discusses the convergence plots (shown in Section~\ref{sec:convergence_plots}) for all problems, grouped by \emph{ladybug} (\ref{sec:ladybug}), \emph{trafalgar} (\ref{sec:trafalgar}), \emph{dubrovnik} (\ref{sec:dubrovnik}), \emph{venice} (\ref{sec:venice}), and \emph{final} (\ref{sec:final}).

\section{Additional details}
\label{sec:appendix_details}

\subsection{Givens rotations}

The QR decomposition of a matrix $A$ can be computed using Givens rotations.
A Givens rotation is represented by a matrix $G_{ij}(\theta)$ that is equal to identity except for rows and columns $i$ and $j$, where it has the non-zero entries
\begin{align}
    \begin{pmatrix}
    g_{ii} & g_{ij} \\g_{ji} & g_{jj}
    \end{pmatrix}
    =
    \begin{pmatrix}
    \cos\theta & \sin\theta \\ -\sin\theta & \cos\theta
    \end{pmatrix}\,.
\end{align}
$G_{ij}(\theta)$ describes a rotation by angle $\theta$ in the $ij$-plane.
The multiplication of a matrix $A$ with $G_{ij}(\theta)$ changes only two rows in $A$, leaving all other rows unchanged.
$\theta$ can be chosen such that the element $(i,j)$ of $G_{ij}(\theta)A$ is zero:
\begin{align}
    \cos\theta = \frac{a_{jj}}{\sqrt{a_{jj}^2+a_{ij}^2}}\,, \quad
    \sin\theta = \frac{a_{ij}}{\sqrt{a_{jj}^2+a_{ij}^2}}\,.
\end{align}
By subsequent multiplication of $A$ with Givens matrices, all elements below the diagonal can be zeroed (see~\cite[p. 252]{golub13} for the full algorithm).
As all $G_{ij}(\theta)$ are orthogonal by construction, their product matrix $Q$ is also orthogonal.

\subsection{Performance profiles}

Let $\mathcal{P}$ be a set of BA problems and $\mathcal{S}$ be the set of evaluated solvers which we run according to the termination criteria (maximum number of iterations and function tolerance).
For a given problem $p \in \mathcal{P}$ and solver $s \in \mathcal{S}$, 
we define the minimal cost value achieved by that solver after time $t$ as $f(p, s, t)$.
The smallest achieved cost by any solver for a specific problem is denoted by $f^*(p) := \min_{s,t} f(p,s,t)$, 
which we use to define for a chosen accuracy tolerance $0 < \tau < 1$ the cost threshold
\begin{align}
  f_\tau(p) := f^*(p) +  \tau (f_0(p) - f^*(p)),
\end{align}
where $f_0(p)$ is the initial cost of problem $p$.
The runtime for solver $s$ to achieve that threshold is
\begin{align}
  t_\tau(p, s) := \min \; \{ t \; | \; f(p, s, t) \leq f_\tau(p) \} \cup \{ \infty \}.
\end{align}
With this, the performance profile of a solver~$s$ is
\begin{align}
  \rho_\tau(s, \alpha) := 100 \frac{|\{ p \; | \; t_\tau(p, s) \leq \alpha \min_s t_\tau(p, s) \}|}{|\mathcal{P}|}.
\end{align}
In other words, $\rho_\tau(s, \alpha)$ maps the relative runtime~$\alpha$ to the percentage of problems that $s$ has solved to accuracy~$\tau$.
The curve is monotonically increasing, 
starting on the left with $\rho_\tau(s, 1)$, the percentage of problems for which solver~$s$ is the fastest, 
and ending on the right with $\max_{\alpha} \rho_\tau(s, \alpha)$, 
the percentage of problems on which the solver~$s$ achieves the cost threshold $f_\tau(p)$ at all.
Comparing different solvers, the curve that is more to the left indicates better runtime 
and the curve that is more to the top indicates higher accuracy.

\section{Algorithm complexities}
\label{sec:complexities}

\begin{table*}[t]
    \centering
    \renewcommand{\arraystretch}{1.4}
    \begin{tabular}{l | l l l}
         & $\sqrt{BA}$ (ours) & explicit SC & implicit SC \\
         \hline
        \textbf{outer iterations} \\
        Jacobian computation &
        $\mathcal{O}(\mu_on_l)$ & $\mathcal{O}(\mu_on_l)$ & $\mathcal{O}(\mu_on_l)$ \\
        Hessian computation &
        $0$ & $\mathcal{O}(\mu_on_l)$ & $0$ \\
        QR &
        $\mathcal{O}((\mu_o^2+\sigma_o^2)n_l)$ & $0$ & $0$ \\
        \textbf{middle iterations} \\
        damping &
        $\mathcal{O}(\mu_on_l)$ & $\mathcal{O}(n_l+n_p)$ & $0$ \\
        SC &
        $0$ & $\mathcal{O}((\mu_o^2+\sigma_o^2)n_l)$ & $0$ \\
        preconditioner & $\mathcal{O}((\mu_o^2+\sigma_o^2)n_l+n_p)$ & $\mathcal{O}(n_p)$ & $\mathcal{O}(\mu_on_l+n_p)$ \\
        back substitution & $\mathcal{O}(\mu_on_l)$ & $\mathcal{O}(\mu_on_l)$ & $\mathcal{O}(\mu_on_l)$ \\
        \textbf{inner iterations} \\
        PCG & $\mathcal{O}((\mu_o^2+\sigma_o^2)n_l+n_p)$ & $\mathcal{O}(n_p^2)$ (worst case) & $\mathcal{O}(\mu_on_l+n_p)$ \\        
    \end{tabular}
    \caption{Complexities of the different steps in our $\sqrt{BA}$ solver compared to the two SC solvers (explicit and implicit), expressed only in terms of $n_l$, $n_p$, $\mu_o$, and $\sigma_o$.
    We split the steps into three stages:
    outer iterations, i.e., everything that needs to be done in order to setup the least squares problem (once per linearization); middle iterations, i.e., everything that needs to be done within one Levenberg-Marquardt iteration (once per outer iteration if no backtracking is required or multiple times if backtracking occurs); inner iterations, i.e., everything that happens within one PCG iteration.}
    \label{tab:complexities}
\end{table*}

In Table~\ref{tab:complexities}, we compare the theoretical complexity of our QR-based solver to the explicit and implicit Schur complement solvers in terms of number of poses $n_p$, number of landmarks $n_l$, and mean $\mu_o$ and variance $\sigma_o^2$ of the number of observations per landmark.
Note that the total number $n_o$ of observations equals $\mu_on_l$.
While most of the entries in the table are easily determined, let us briefly walk through the not so obvious ones:

\paragraph{$\sqrt{BA}$ (our solver)}
Assume landmark $j$ has $k_j$ observations.
We can express the sum over $k_j^2$ by $\mu_o$ and $\sigma_o^2$:
\begin{gather}
    \sigma_o^2=\operatorname{Var}(\{k_j\}) = \frac{1}{n_l}\sum_{j=1}^{n_l}{k_j^2} - (\frac{1}{n_l}\sum_{j=1}^{n_l}{k_j})^2\,, \\
    \Rightarrow \sum_{j=1}^{n_l}{k_j^2} = n_l(\mu_o^2 + \sigma_o^2)\,.
\end{gather}
The sum over $k_j^2$ appears in many parts of the algorithm, as the dense landmark blocks $\begin{pmatrix}\hat{J}_p & \hat{J}_l & \hat{r}\end{pmatrix}$ after QR decomposition have size $(2k_j+3)\times(d_pk_j+4)$, where the number of parameters per camera $d_p$ is 9 in our experiments.
In the QR step, we need $6k_j$ Givens rotations per landmark (out of which 6 are for the damping), and we multiply the dense landmark block by $\hat{Q}^\top$, so we end up having terms $\mathcal{O}(\sum_j{k_j})$ and $\mathcal{O}(\sum_j{k_j^2})$, leading to the complexity stated in Table~\ref{tab:complexities}.
For the block-diagonal preconditioner, each landmark block contributes summands to $k_j$ diagonal blocks, each of which needs a matrix multiplication with $\mathcal{O}(k_j)$ flops, thus we have a complexity of $\mathcal{O}(\sum_jk_j^2)$.
Preconditioner inversion can be done block-wise and is thus $\mathcal{O}(n_p)$.
In the PCG step, we can exploit the block-sparse structure of $\hat{Q}_2^\top \hat{J}_p$ and again have the $k_j^2$-dependency.
Because of the involved vectors being of size $2n_o+d_pn_p$ (due to pose damping), we additionally have a dependency on $2n_o+d_pn_p$.
Finally, for the back substitution, we need to solve $n_l$ upper-triangular $3\times 3$ systems and then effectively do $n_l$ multiplications of a $(3\times d_pk_j)$ matrix with a vector, which in total is of order $\mathcal{O}(\sum_jk_j)$.

\paragraph{Explicit SC}
The first step is the Hessian computation.
As each single residual only depends on one pose and one landmark, the Hessian computation scales with the number of observations/residuals.
Damping is a simple augmentation of the diagonal and contributes terms $\mathcal{O}(n_l)$ and $\mathcal{O}(n_p)$.
Matrix inversion of $H_{ll}$ for the Schur complement scales linearly with $n_l$, while the number of operations to multiply $H_{pl}$ by $H_{ll}^{-1}$ scales with the number of non-zero sub-blocks in $H_{pl}$, and thus with $n_o$.
The multiplication of this product with $H_{lp}$ involves matrix products of sub-blocks sized $(d_p\times 3)$ and $(3\times d_p)$ for each camera pair that shares a given landmark, i.e., $\mathcal{O}(k_j^2)$ matrix products for landmark $j$.
The preconditioner can simply be read off from $\tilde{H}_{pp}$, and its inversion is the same as for $\sqrt{BA}$.
The matrices and vectors involved in PCG all have size $d_pn_p(\times d_pn_p)$.
The sparsity of $\tilde{H}_{pp}$ is not only determined by $n_p$, $n_l$, $\mu_o$, and $\sigma_o$, but would require knowledge about which cameras share at least one landmark. In the worst case, where each pair of camera poses have at least one landmark they both observe, $\tilde{H}_{pp}$ is dense.
Thus, assuming $\tilde{H}_{pp}$ as dense we get quadratic dependence on $n_p$.
Back substitution consists of matrix inversion of $n_l$ blocks, and a simple matrix-vector multiplication.

\paragraph{Implicit SC}
Since the Hessian matrix is not explicitly computed, we need an extra step to compute the preconditioner for implicit SC.
For each pose, we have to compute a $d_p\times d_p$ block for which the number of flops scales linearly with the number of observations of that pose, thus it is $\mathcal{O}(n_o)$ in total.
Preconditioner damping contributes the $n_p$-dependency.
As no matrices except for the preconditioner are precomputed for the PCG iterations, but sparsity can again be exploited to avoid quadratic complexities, this part of the algorithm scales linearly with all three numbers (assuming the outer loop for the preconditioner computation is a \emph{parallel for} over cameras, rather than a \emph{parallel reduce} over landmarks).
Lastly, back substitution is again only block-wise matrix inversion and matrix-vector multiplications.
While avoiding a dependency on $(\mu_o^2+\sigma_o^2)n_l$ in the asymptotic runtime seems appealing, the implicit SC method computes a sequence of five sparse matrix-vector products in each PCG iteration in addition to the preconditioner multiplication, making it harder to parallelize than the other two methods, which have only one (explicit SC) or two ($\sqrt{BA}$) sparse matrix-vector products.
Thus, the benefit of implicit SC becomes apparent only for very large problems. 
As our evaluation shows, for medium and large problems, i.e.\ the majority in the BAL dataset, our $\sqrt{BA}$ solver is still superior in runtime.

\section{Problem sizes}
\label{sec:problem_sizes}

Table~\ref{tab:problem-size} in Section~\ref{sec:problem_sizes_table}
details the size of the bundle adjustment problem for each instance in the BAL dataset (grouped into \emph{ladybug}, the skeletal problems \emph{trafalgar}, \emph{dubrovnik}, and \emph{venice}, as well as the \emph{final} problems). 
Besides number of cameras $n_p$, number of landmarks $n_l$, and number of observations $n_r = \frac{N_r}{2}$, 
we also show indicators for \emph{problem density}: 
the average number of observations per camera \emph{\#obs / cam} (which equals $n_r / n_p$), 
as well as the average number of observations per landmark \emph{\#obs / lm}, including its standard deviation and maximum over all landmarks. 

In particular, a high mean and variance of \emph{\#obs / lm} indicates that our proposed $\sqrt{BA}$ solver may require a large amount of memory (see for example \emph{final961}, \emph{final1936}, \emph{final13682}), 
since the dense storage after marginalization in a landmark block is quadratic in the number of observations of that landmark.
If on the other hand the problems are sparse and the number of observations is moderate, the memory required by $\sqrt{BA}$ grows only linearly in the number of observations, similar to SC-based solvers (see Figure~6 in the main paper).

\section{Convergence}
\label{sec:convergence}

\balance

In Section~\ref{sec:convergence_plots}, each row of plots corresponds to one of the 97 bundle adjustment problems and contains from left to right a plot of optimized cost by runtime (like Figure~5 in the main paper) and by iteration, trust-region size (inverse of the damping factor~$\lambda$) by iteration, number of CG iterations by (outer) iteration, and peak memory usage by iteration. 
The cost plots are cut off at the top and horizontal lines indicate the cost thresholds corresponding to accuracy tolerances $\tau \in \{10^{-1}, 10^{-2}, 10^{-3}\}$ as used in the performance profiles. The plot by runtime is additionally cut off on the right at the time the fastest solver for the respective problem terminated.

We make a few observations that additionally support our claims in the main paper: all solvers usually converge to a similar cost, but for most problems our proposed $\sqrt{BA}$ solver is the fastest to reduce the cost. On the other hand, memory use can be higher, depending on problem size and density (see Section~\ref{sec:problem_sizes}).
Missing plots indicate that the respective solver ran out of memory, which for example for \emph{$\sqrt{BA}$-32} happens only on the largest problem \emph{final13682}, where the landmarks have up to 1748 observations.
Our single precision solver \emph{$\sqrt{BA}$-32} runs around twice as fast as its double precision counterpart, since it is numerically stable and usually requires a comparable number of CG iterations.
This is in contrast to our custom SC solver, where the twofold speedup for single precision is generally not observed. 
The good numeric properties are further supported by the evolution of the trust-region size approximately following that of the other solvers in most cases.
Finally, for the smallest problems (e.g., \emph{ladybug49}, \emph{trafalgar21}, \emph{final93}), the evolution of cost, trust-region size, and even number of CG iterations is often identical for all solvers for the initial 5 to 15 iterations, before numeric differences become noticeable.
This supports the fact that the different marginalization strategies are algebraically equivalent and that our custom solver implementation uses the same Levenberg-Marquardt strategy and CG forcing sequence as Ceres.

\onecolumn

\section{Problem sizes table}
\label{sec:problem_sizes_table}

{
\setlength{\LTcapwidth}{0.99\textwidth}
\begin{longtable}{l r r r r r r r}%
\label{tab:problem-size}
\endfirsthead
\endhead
\toprule
&\multicolumn{1}{c}{\#cam}&\multicolumn{1}{c}{\#lm}&\multicolumn{1}{c}{\#obs}&\multicolumn{1}{c}{\#obs / cam}&\multicolumn{3}{c}{\#obs / lm} \\
&\multicolumn{1}{c}{$(n_p)$}&\multicolumn{1}{c}{$(n_l)$}&\multicolumn{1}{c}{$(n_r)$}& \multicolumn{1}{c}{mean} & \multicolumn{1}{c}{mean} & \multicolumn{1}{c}{std-dev} & \multicolumn{1}{c}{max}\\
\midrule
ladybug49&49&7,766&31,812&649.2&4.1&3.3&29\\%
ladybug73&73&11,022&46,091&631.4&4.2&3.7&40\\%
ladybug138&138&19,867&85,184&617.3&4.3&4.4&48\\%
ladybug318&318&41,616&179,883&565.7&4.3&4.8&89\\%
ladybug372&372&47,410&204,434&549.6&4.3&4.8&134\\%
ladybug412&412&52,202&224,205&544.2&4.3&4.8&135\\%
ladybug460&460&56,799&241,842&525.7&4.3&4.7&135\\%
ladybug539&539&65,208&277,238&514.4&4.3&4.7&142\\%
ladybug598&598&69,193&304,108&508.5&4.4&4.9&142\\%
ladybug646&646&73,541&327,199&506.5&4.4&5.0&144\\%
ladybug707&707&78,410&349,753&494.7&4.5&5.0&145\\%
ladybug783&783&84,384&376,835&481.3&4.5&5.0&145\\%
ladybug810&810&88,754&393,557&485.9&4.4&4.9&145\\%
ladybug856&856&93,284&415,551&485.5&4.5&4.9&145\\%
ladybug885&885&97,410&434,681&491.2&4.5&4.9&145\\%
ladybug931&931&102,633&457,231&491.1&4.5&5.0&145\\%
ladybug969&969&105,759&474,396&489.6&4.5&5.2&145\\%
ladybug1064&1,064&113,589&509,982&479.3&4.5&5.1&145\\%
ladybug1118&1,118&118,316&528,693&472.9&4.5&5.1&145\\%
ladybug1152&1,152&122,200&545,584&473.6&4.5&5.1&145\\%
ladybug1197&1,197&126,257&563,496&470.8&4.5&5.1&145\\%
ladybug1235&1,235&129,562&576,045&466.4&4.4&5.1&145\\%
ladybug1266&1,266&132,521&587,701&464.2&4.4&5.0&145\\%
ladybug1340&1,340&137,003&612,344&457.0&4.5&5.2&145\\%
ladybug1469&1,469&145,116&641,383&436.6&4.4&5.1&145\\%
ladybug1514&1,514&147,235&651,217&430.1&4.4&5.1&145\\%
ladybug1587&1,587&150,760&663,019&417.8&4.4&5.1&145\\%
ladybug1642&1,642&153,735&670,999&408.6&4.4&5.0&145\\%
ladybug1695&1,695&155,621&676,317&399.0&4.3&5.0&145\\%
ladybug1723&1,723&156,410&678,421&393.7&4.3&5.0&145\\%
\midrule
&\multicolumn{1}{c}{\#cam}&\multicolumn{1}{c}{\#lm}&\multicolumn{1}{c}{\#obs}&\multicolumn{1}{c}{\#obs / cam}&\multicolumn{3}{c}{\#obs / lm} \\
&\multicolumn{1}{c}{$(n_p)$}&\multicolumn{1}{c}{$(n_l)$}&\multicolumn{1}{c}{$(n_r)$}& \multicolumn{1}{c}{mean} & \multicolumn{1}{c}{mean} & \multicolumn{1}{c}{std-dev} & \multicolumn{1}{c}{max}\\
\midrule
trafalgar21&21&11,315&36,455&1,736.0&3.2&1.8&15\\%
trafalgar39&39&18,060&63,551&1,629.5&3.5&2.4&20\\%
trafalgar50&50&20,431&73,967&1,479.3&3.6&2.7&21\\%
trafalgar126&126&40,037&148,117&1,175.5&3.7&3.0&29\\%
trafalgar138&138&44,033&165,688&1,200.6&3.8&3.3&32\\%
trafalgar161&161&48,126&181,861&1,129.6&3.8&3.4&40\\%
trafalgar170&170&49,267&185,604&1,091.8&3.8&3.5&41\\%
trafalgar174&174&50,489&188,598&1,083.9&3.7&3.4&41\\%
trafalgar193&193&53,101&196,315&1,017.2&3.7&3.4&42\\%
trafalgar201&201&54,427&199,727&993.7&3.7&3.4&42\\%
trafalgar206&206&54,562&200,504&973.3&3.7&3.4&42\\%
trafalgar215&215&55,910&203,991&948.8&3.6&3.4&42\\%
trafalgar225&225&57,665&208,411&926.3&3.6&3.3&42\\%
trafalgar257&257&65,131&225,698&878.2&3.5&3.2&42\\%
\midrule
&\multicolumn{1}{c}{\#cam}&\multicolumn{1}{c}{\#lm}&\multicolumn{1}{c}{\#obs}&\multicolumn{1}{c}{\#obs / cam}&\multicolumn{3}{c}{\#obs / lm} \\
&\multicolumn{1}{c}{$(n_p)$}&\multicolumn{1}{c}{$(n_l)$}&\multicolumn{1}{c}{$(n_r)$}& \multicolumn{1}{c}{mean} & \multicolumn{1}{c}{mean} & \multicolumn{1}{c}{std-dev} & \multicolumn{1}{c}{max}\\
\midrule
dubrovnik16&16&22,106&83,718&5,232.4&3.8&2.2&14\\%
dubrovnik88&88&64,298&383,937&4,362.9&6.0&6.0&65\\%
dubrovnik135&135&90,642&552,949&4,095.9&6.1&7.1&84\\%
dubrovnik142&142&93,602&565,223&3,980.4&6.0&7.1&84\\%
dubrovnik150&150&95,821&567,738&3,784.9&5.9&6.9&84\\%
dubrovnik161&161&103,832&591,343&3,672.9&5.7&6.7&84\\%
dubrovnik173&173&111,908&633,894&3,664.1&5.7&6.7&84\\%
dubrovnik182&182&116,770&668,030&3,670.5&5.7&6.9&85\\%
dubrovnik202&202&132,796&750,977&3,717.7&5.7&6.7&91\\%
dubrovnik237&237&154,414&857,656&3,618.8&5.6&6.6&99\\%
dubrovnik253&253&163,691&898,485&3,551.3&5.5&6.6&102\\%
dubrovnik262&262&169,354&919,020&3,507.7&5.4&6.5&106\\%
dubrovnik273&273&176,305&942,302&3,451.7&5.3&6.5&112\\%
dubrovnik287&287&182,023&970,624&3,382.0&5.3&6.5&120\\%
dubrovnik308&308&195,089&1,044,529&3,391.3&5.4&6.3&121\\%
dubrovnik356&356&226,729&1,254,598&3,524.2&5.5&6.4&122\\%
\midrule
&\multicolumn{1}{c}{\#cam}&\multicolumn{1}{c}{\#lm}&\multicolumn{1}{c}{\#obs}&\multicolumn{1}{c}{\#obs / cam}&\multicolumn{3}{c}{\#obs / lm} \\
&\multicolumn{1}{c}{$(n_p)$}&\multicolumn{1}{c}{$(n_l)$}&\multicolumn{1}{c}{$(n_r)$}& \multicolumn{1}{c}{mean} & \multicolumn{1}{c}{mean} & \multicolumn{1}{c}{std-dev} & \multicolumn{1}{c}{max}\\
\midrule
venice52&52&64,053&347,173&6,676.4&5.4&5.9&46\\%
venice89&89&110,973&562,976&6,325.6&5.1&5.9&62\\%
venice245&245&197,919&1,087,436&4,438.5&5.5&7.2&85\\%
venice427&427&309,567&1,695,237&3,970.1&5.5&7.2&119\\%
venice744&744&542,742&3,054,949&4,106.1&5.6&8.6&205\\%
venice951&951&707,453&3,744,975&3,937.9&5.3&7.7&213\\%
venice1102&1,102&779,640&4,048,424&3,673.7&5.2&7.5&221\\%
venice1158&1,158&802,093&4,126,104&3,563.1&5.1&7.4&223\\%
venice1184&1,184&815,761&4,174,654&3,525.9&5.1&7.3&223\\%
venice1238&1,238&842,712&4,286,111&3,462.1&5.1&7.3&224\\%
venice1288&1,288&865,630&4,378,614&3,399.5&5.1&7.2&225\\%
venice1350&1,350&893,894&4,512,735&3,342.8&5.0&7.1&225\\%
venice1408&1,408&911,407&4,630,139&3,288.5&5.1&7.1&225\\%
venice1425&1,425&916,072&4,652,920&3,265.2&5.1&7.1&225\\%
venice1473&1,473&929,522&4,701,478&3,191.8&5.1&7.1&226\\%
venice1490&1,490&934,449&4,717,420&3,166.1&5.0&7.1&228\\%
venice1521&1,521&938,727&4,734,634&3,112.8&5.0&7.1&230\\%
venice1544&1,544&941,585&4,745,797&3,073.7&5.0&7.1&231\\%
venice1638&1,638&975,980&4,952,422&3,023.5&5.1&7.1&231\\%
venice1666&1,666&983,088&4,982,752&2,990.8&5.1&7.2&231\\%
venice1672&1,672&986,140&4,995,719&2,987.9&5.1&7.2&231\\%
venice1681&1,681&982,593&4,962,448&2,952.1&5.1&7.2&231\\%
venice1682&1,682&982,446&4,960,627&2,949.2&5.0&7.2&231\\%
venice1684&1,684&982,447&4,961,337&2,946.2&5.0&7.2&231\\%
venice1695&1,695&983,867&4,966,552&2,930.1&5.0&7.2&231\\%
venice1696&1,696&983,994&4,966,505&2,928.4&5.0&7.2&231\\%
venice1706&1,706&984,707&4,970,241&2,913.4&5.0&7.2&232\\%
venice1776&1,776&993,087&4,997,468&2,813.9&5.0&7.1&232\\%
venice1778&1,778&993,101&4,997,555&2,810.8&5.0&7.1&232\\%
\midrule
&\multicolumn{1}{c}{\#cam}&\multicolumn{1}{c}{\#lm}&\multicolumn{1}{c}{\#obs}&\multicolumn{1}{c}{\#obs / cam}&\multicolumn{3}{c}{\#obs / lm} \\
&\multicolumn{1}{c}{$(n_p)$}&\multicolumn{1}{c}{$(n_l)$}&\multicolumn{1}{c}{$(n_r)$}& \multicolumn{1}{c}{mean} & \multicolumn{1}{c}{mean} & \multicolumn{1}{c}{std-dev} & \multicolumn{1}{c}{max}\\
\midrule
final93&93&61,203&287,451&3,090.9&4.7&5.8&80\\%
final394&394&100,368&534,408&1,356.4&5.3&10.6&280\\%
final871&871&527,480&2,785,016&3,197.5&5.3&9.8&245\\%
final961&961&187,103&1,692,975&1,761.7&9.0&29.3&839\\%
final1936&1,936&649,672&5,213,731&2,693.0&8.0&26.9&1293\\%
final3068&3,068&310,846&1,653,045&538.8&5.3&12.6&414\\%
final4585&4,585&1,324,548&9,124,880&1,990.2&6.9&12.6&535\\%
final13682&13,682&4,455,575&28,973,703&2,117.7&6.5&18.9&1748\\%

\bottomrule
\caption{Size of the bundle adjustment problem for each instance in the BAL dataset.}
\end{longtable}
}

\section{Convergence plots}
\label{sec:convergence_plots}

\subsection{Ladybug}
\label{sec:ladybug}

\begin{center}
\includegraphics[width=\textwidth]{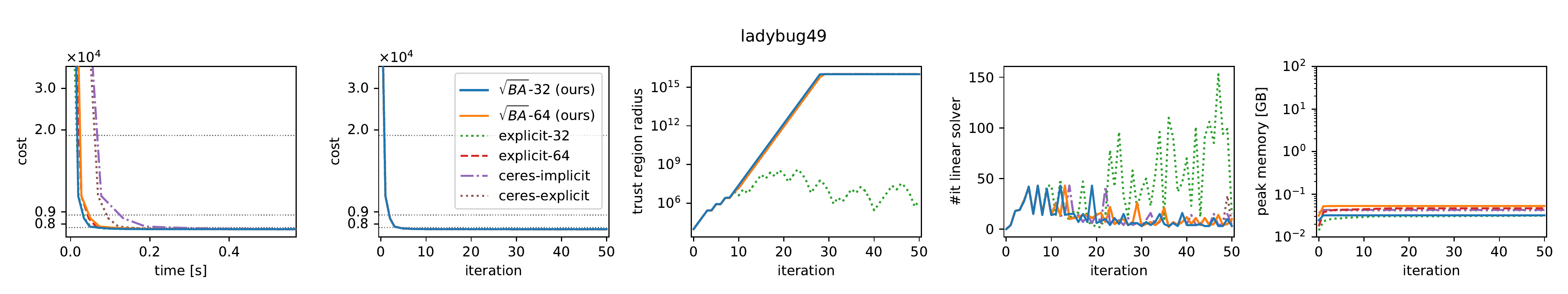}\\
\includegraphics[width=\textwidth]{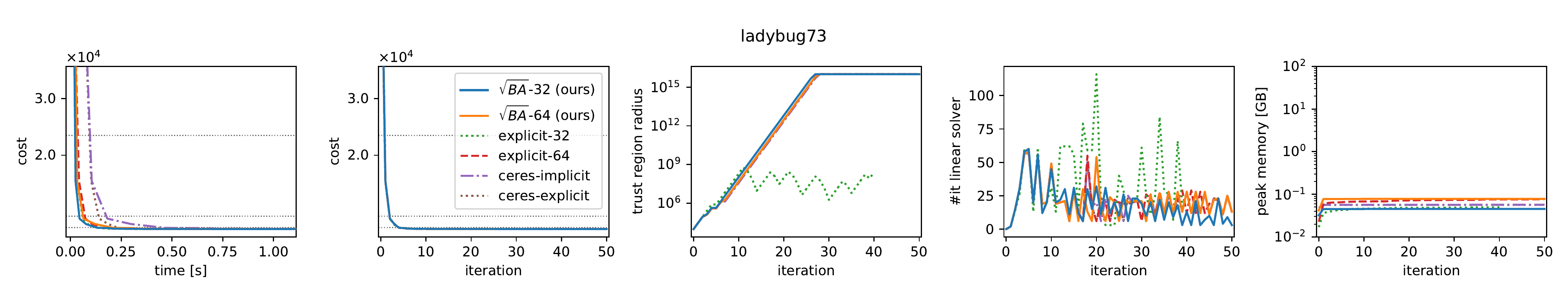}\\
\includegraphics[width=\textwidth]{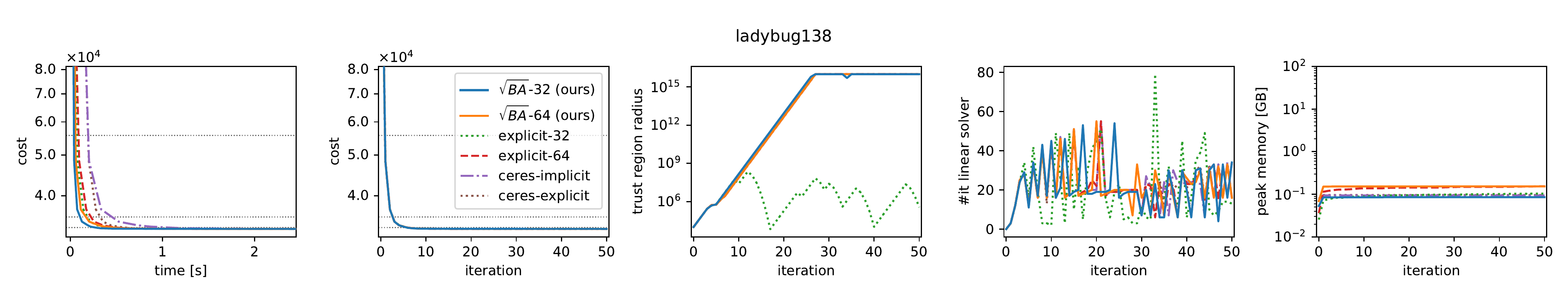}
\includegraphics[width=\textwidth]{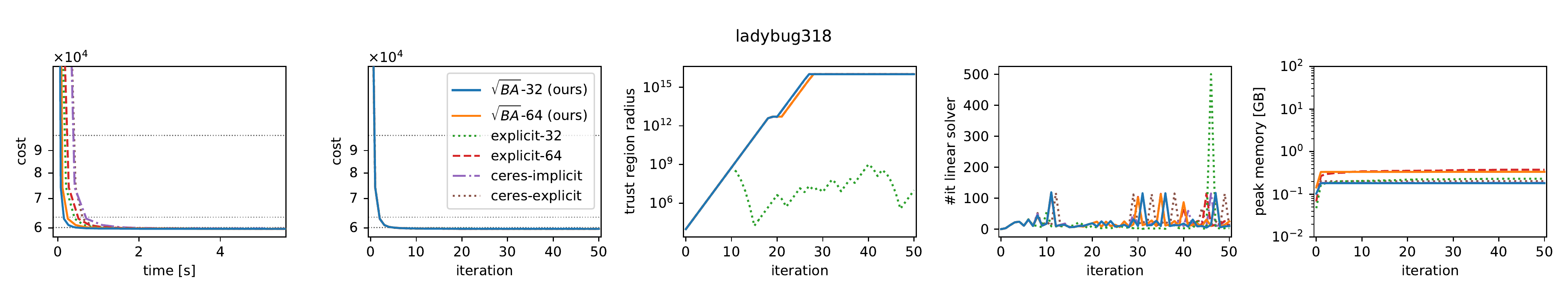}
\includegraphics[width=\textwidth]{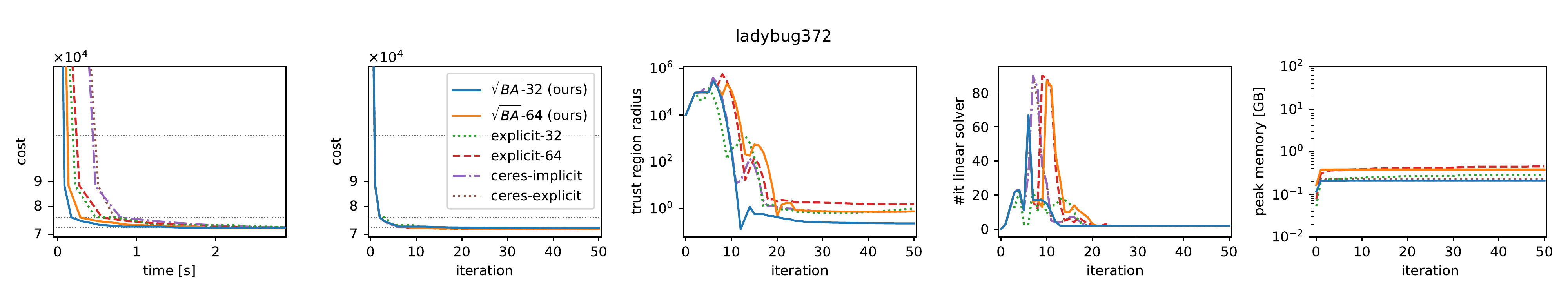}
\includegraphics[width=\textwidth]{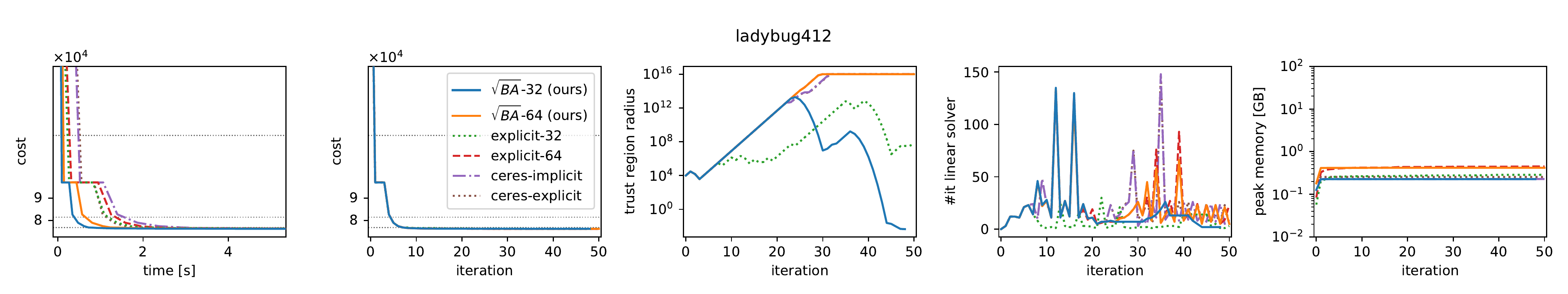}
\includegraphics[width=\textwidth]{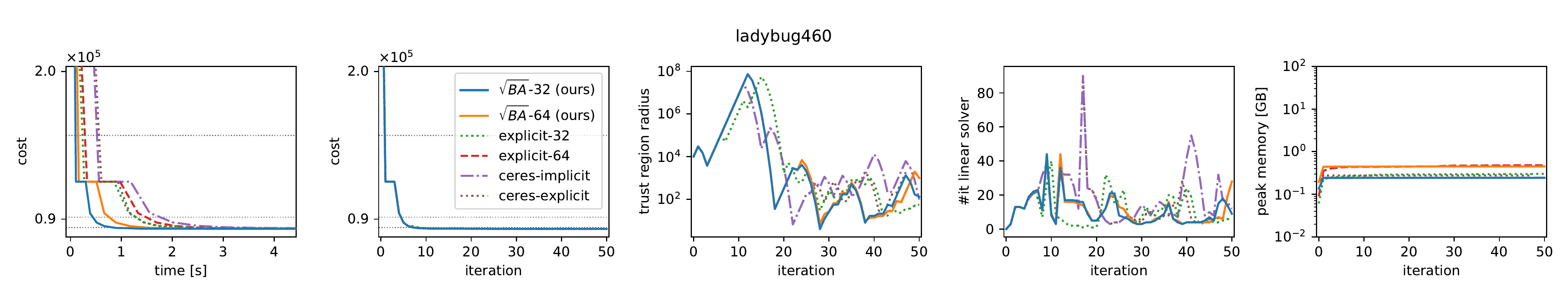}
\includegraphics[width=\textwidth]{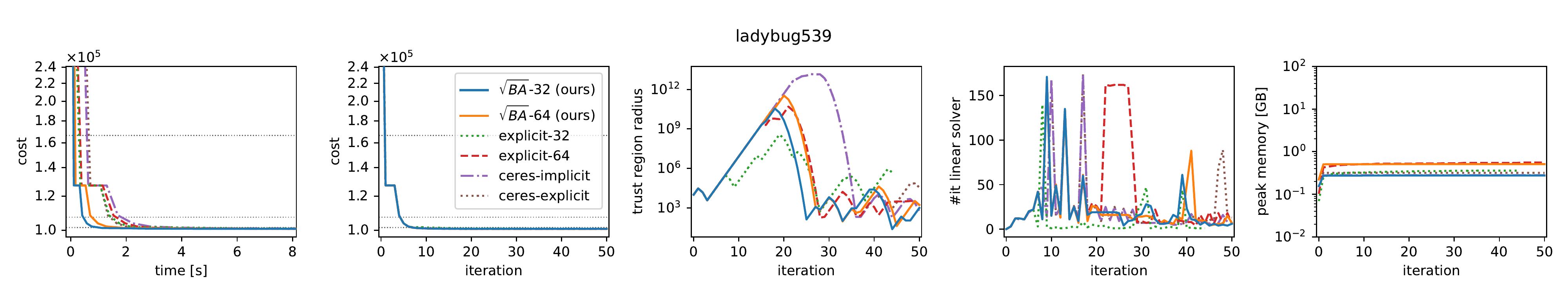}
\includegraphics[width=\textwidth]{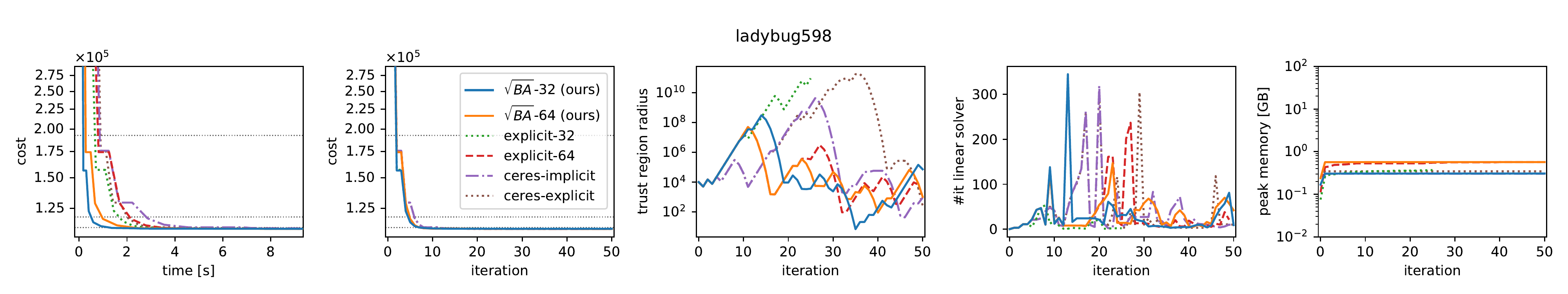}
\includegraphics[width=\textwidth]{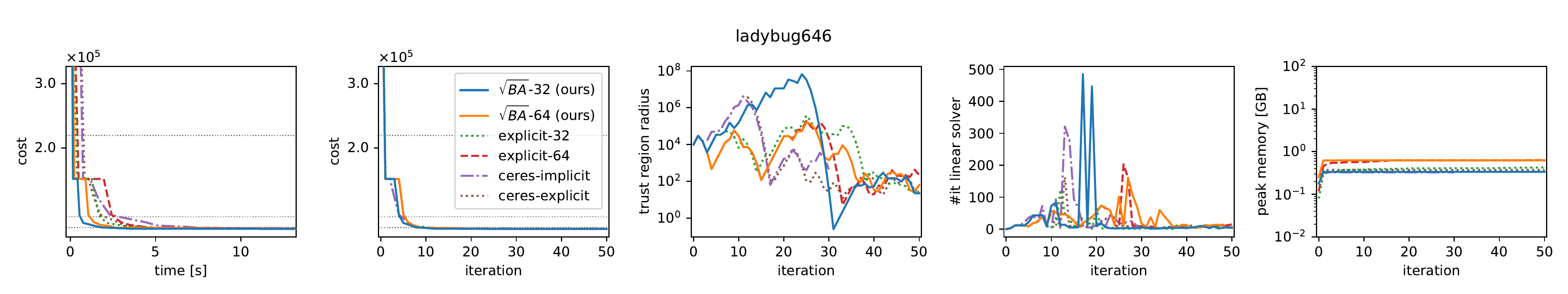}
\includegraphics[width=\textwidth]{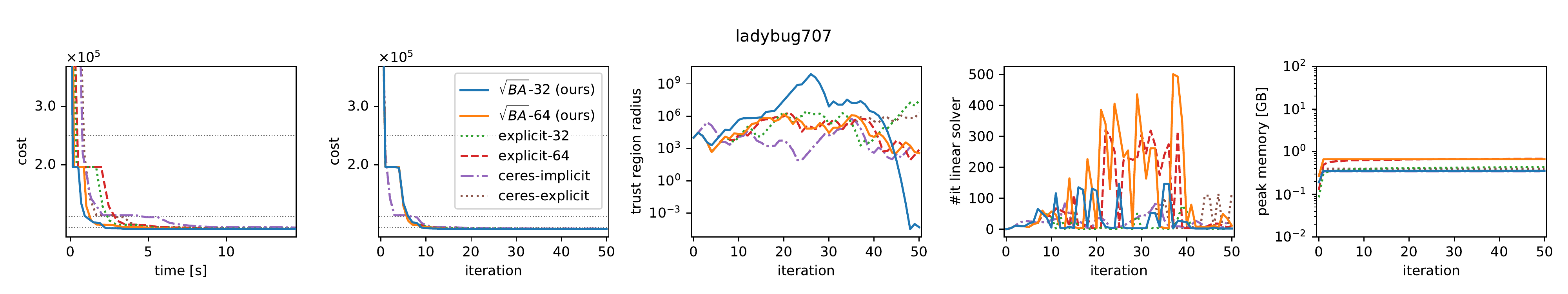}
\includegraphics[width=\textwidth]{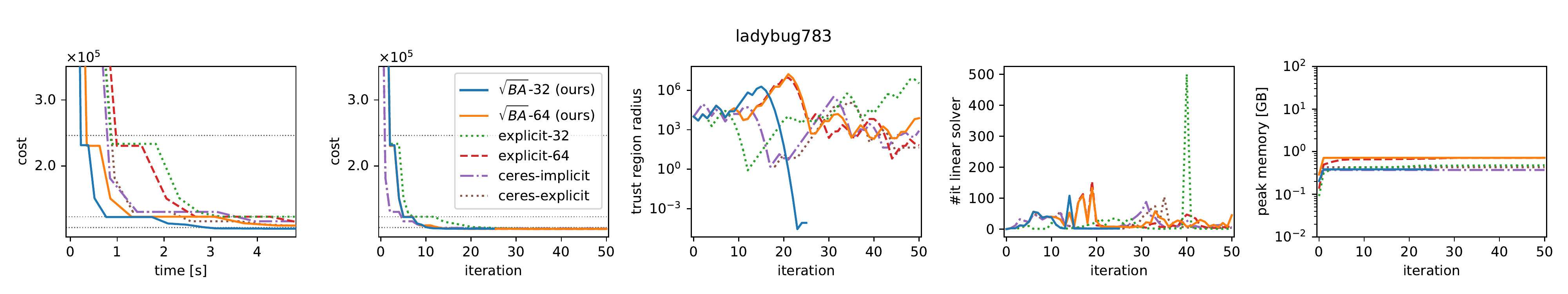}
\includegraphics[width=\textwidth]{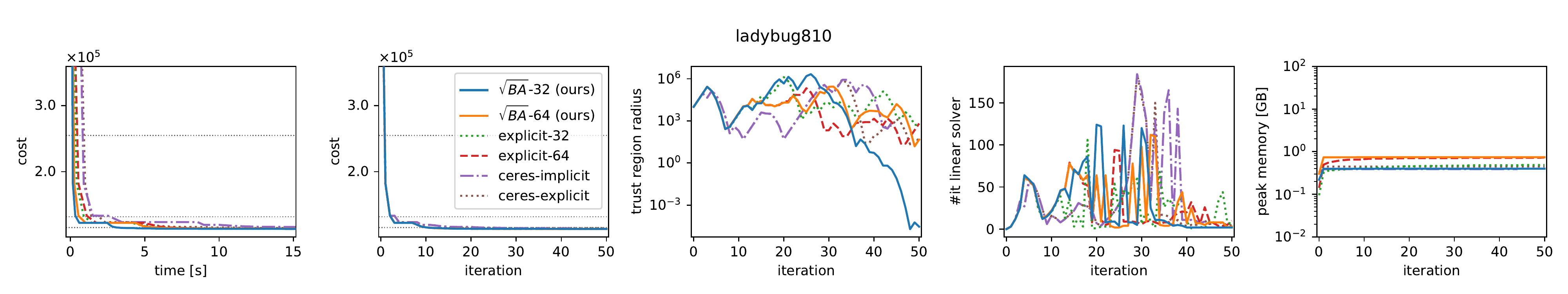}
\includegraphics[width=\textwidth]{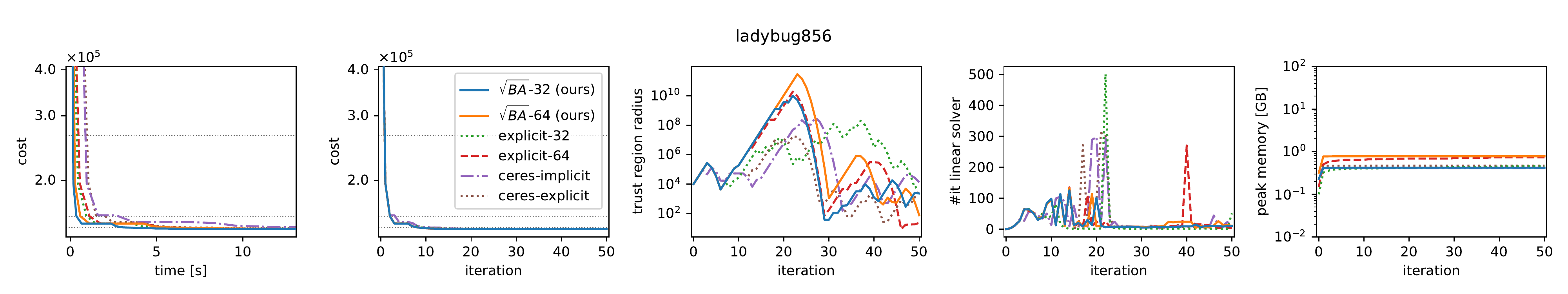}
\includegraphics[width=\textwidth]{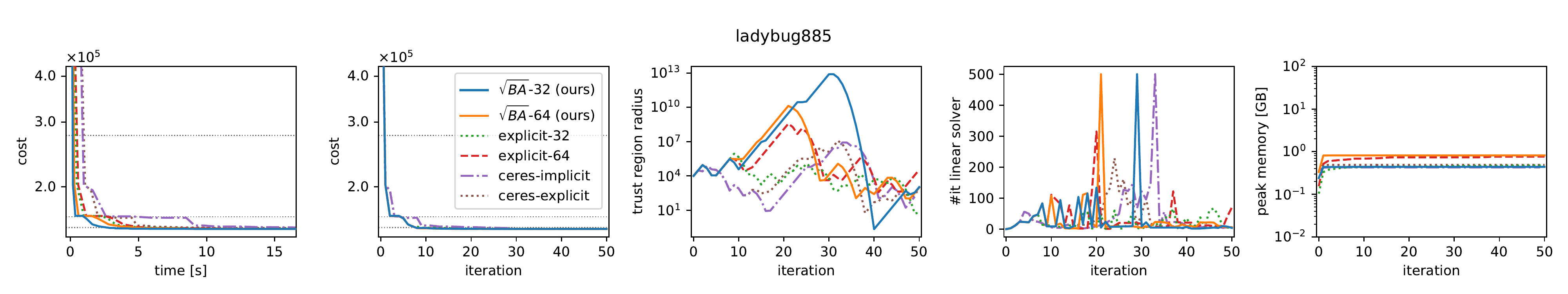}
\includegraphics[width=\textwidth]{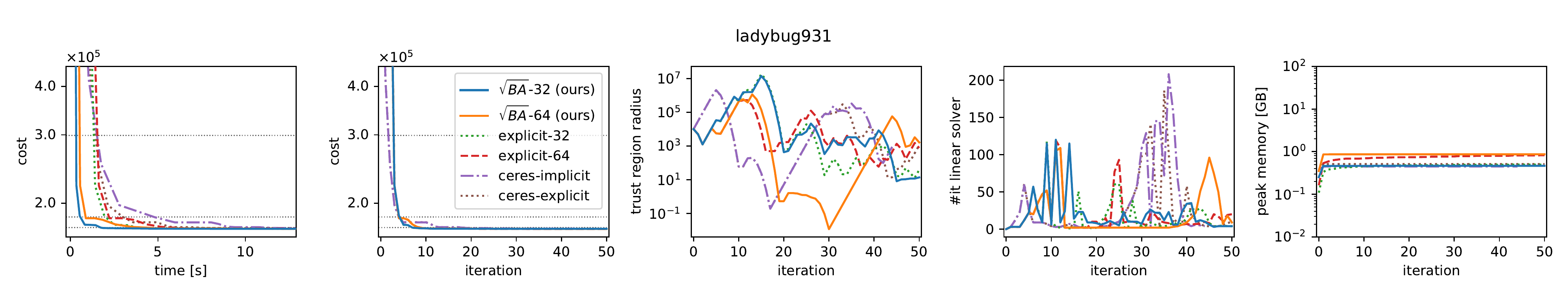}
\includegraphics[width=\textwidth]{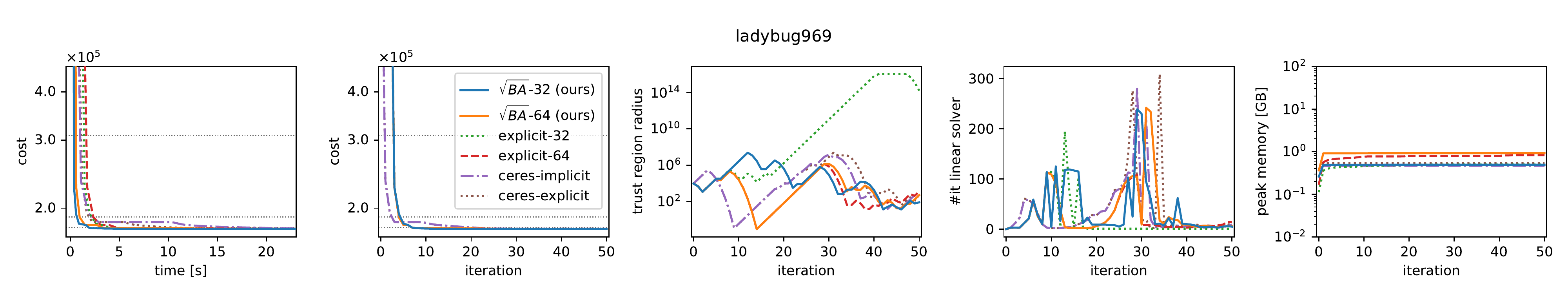}
\includegraphics[width=\textwidth]{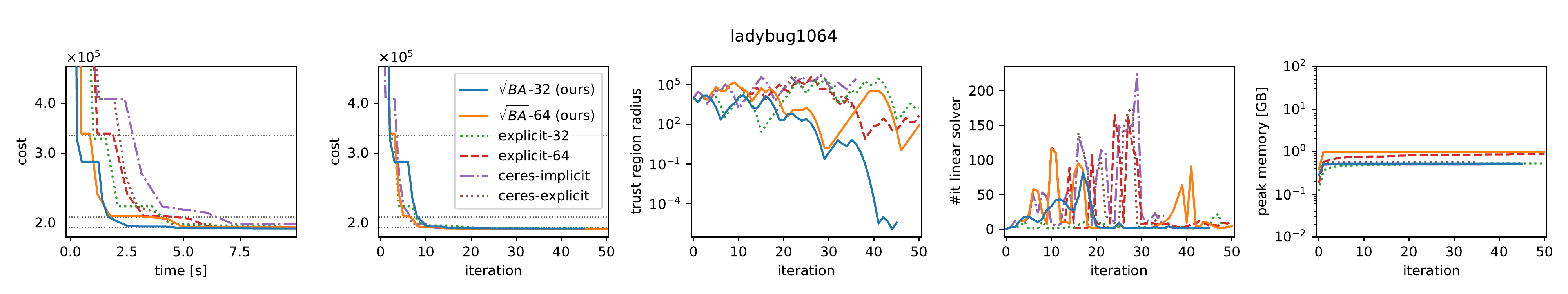}
\includegraphics[width=\textwidth]{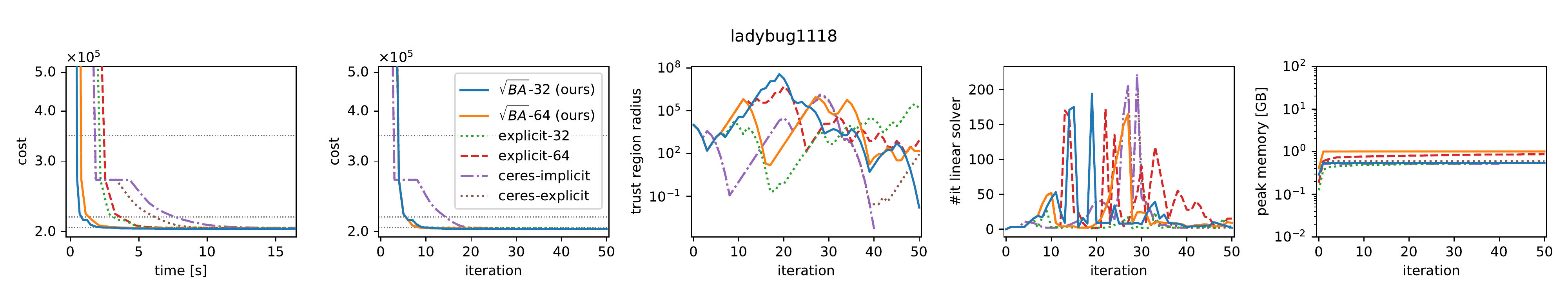}
\includegraphics[width=\textwidth]{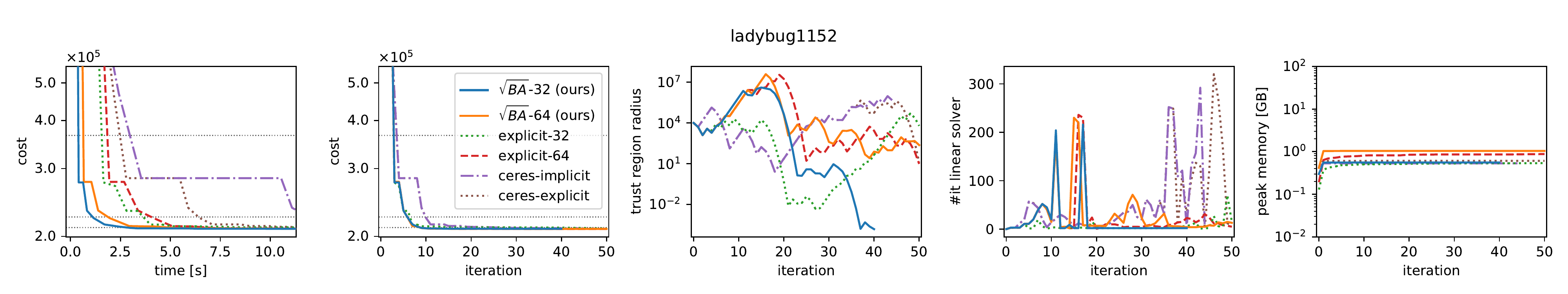}
\includegraphics[width=\textwidth]{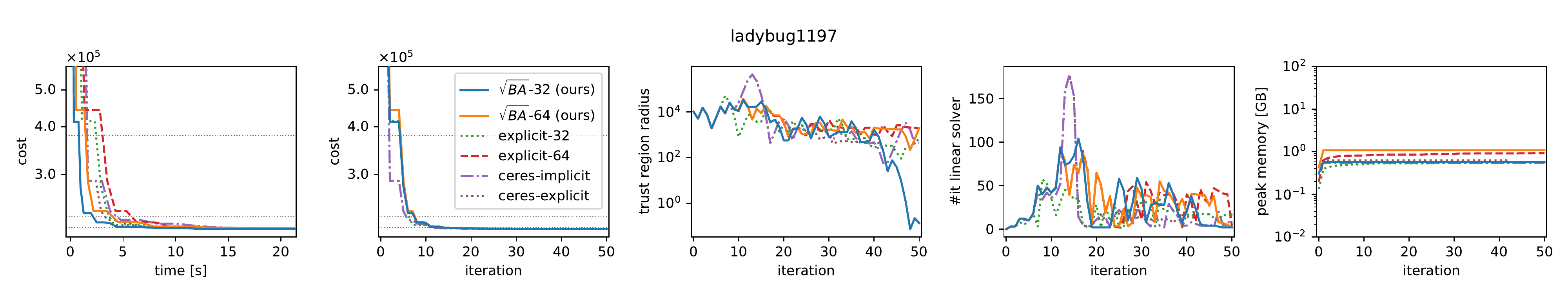}
\includegraphics[width=\textwidth]{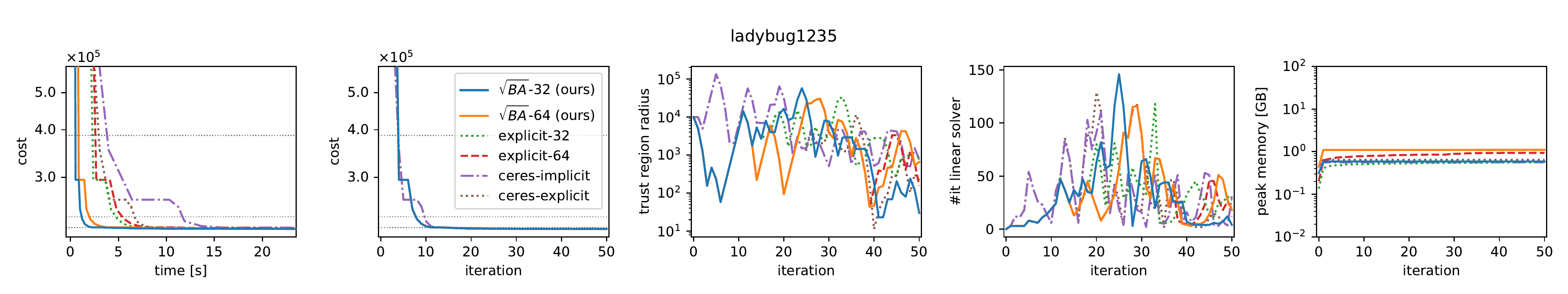}
\includegraphics[width=\textwidth]{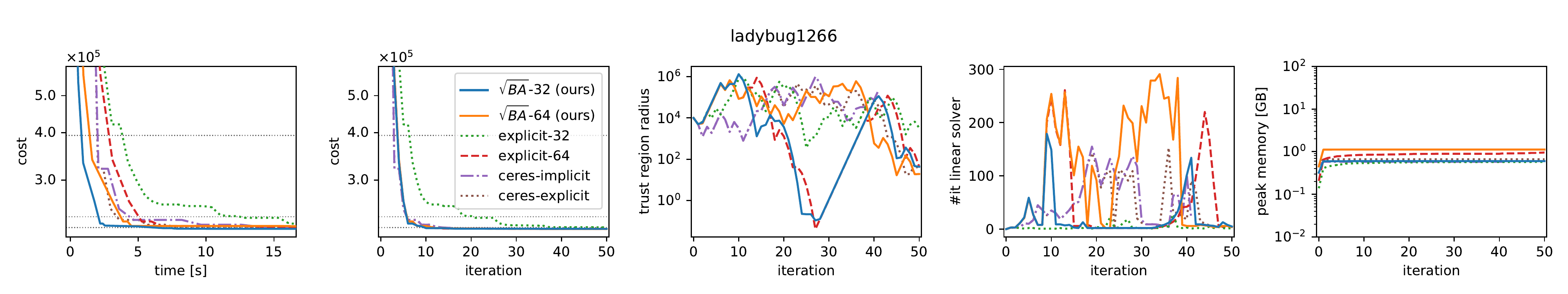}
\includegraphics[width=\textwidth]{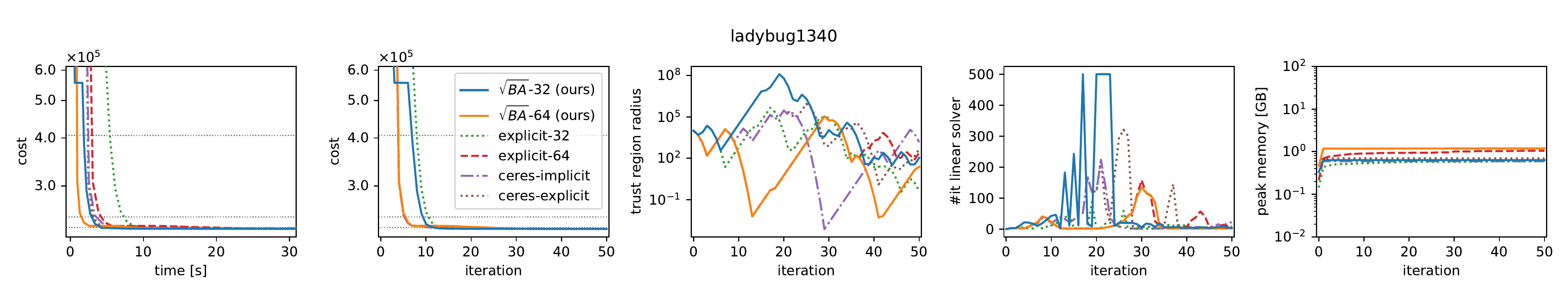}
\includegraphics[width=\textwidth]{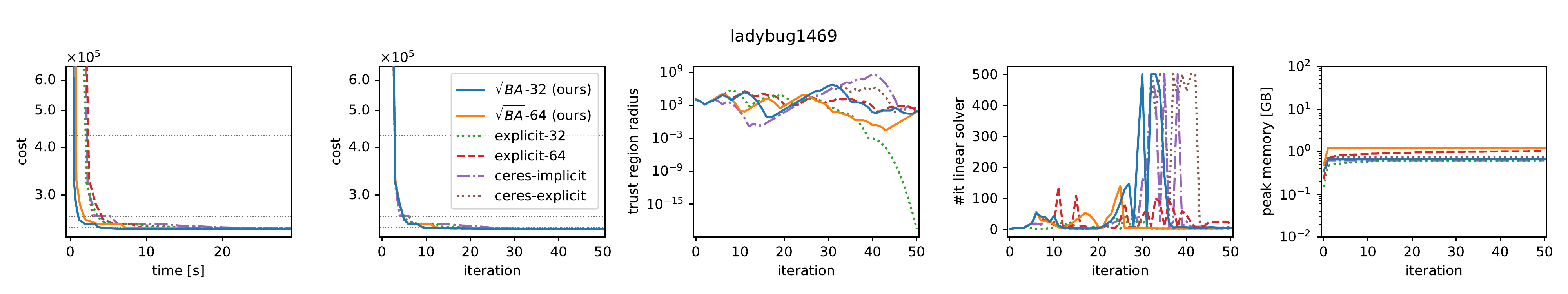}
\includegraphics[width=\textwidth]{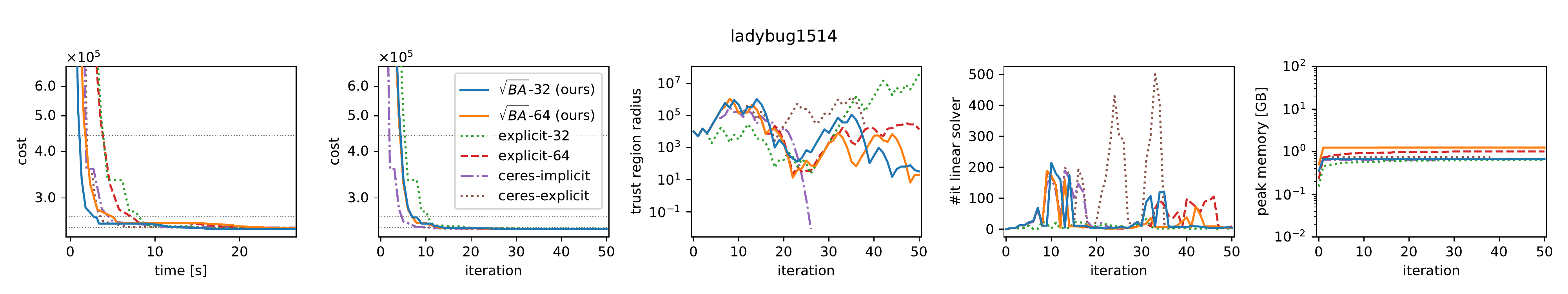}
\includegraphics[width=\textwidth]{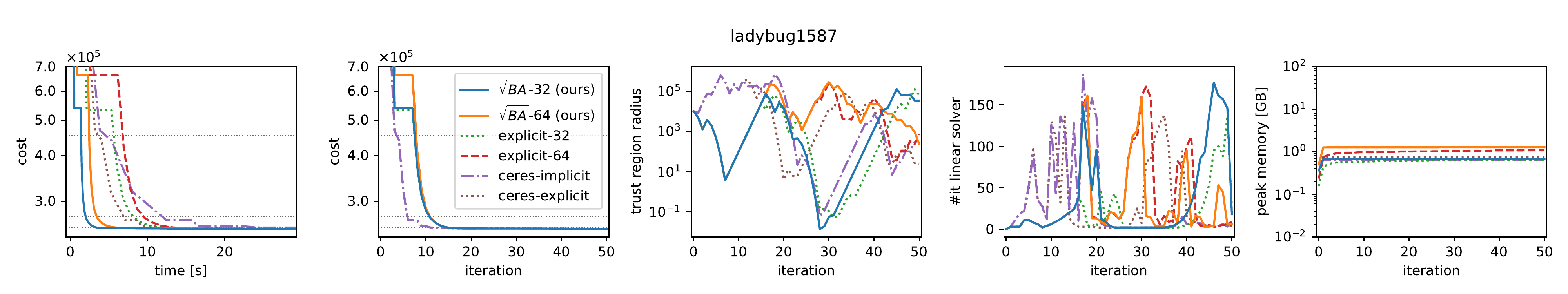}
\includegraphics[width=\textwidth]{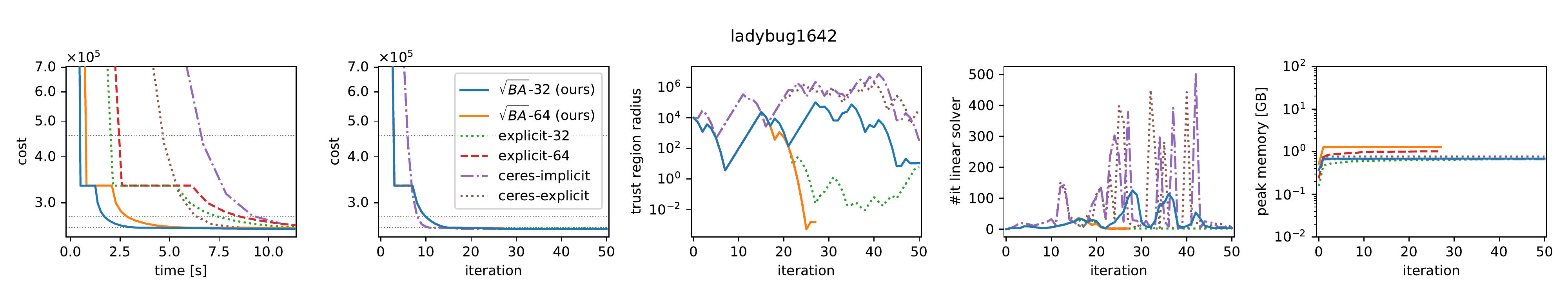}
\includegraphics[width=\textwidth]{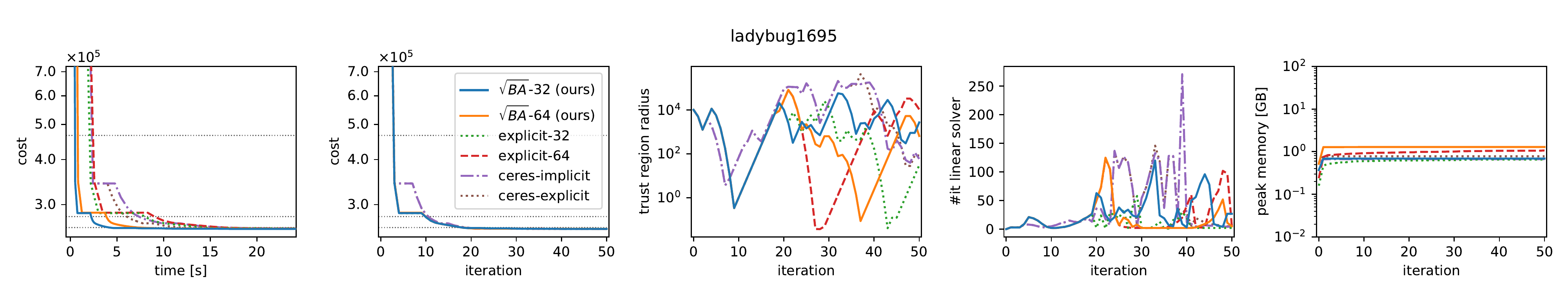}
\includegraphics[width=\textwidth]{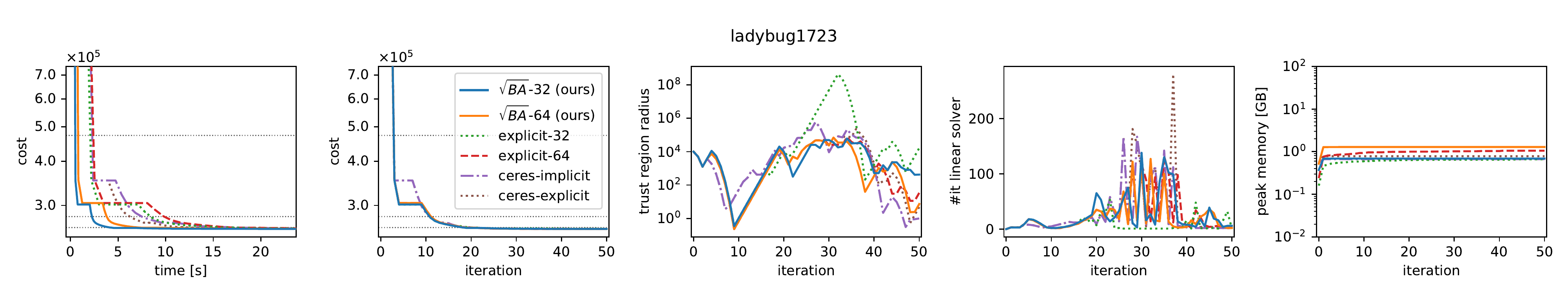}
\end{center}

\subsection{Trafalgar}
\label{sec:trafalgar}

\begin{center}
\includegraphics[width=\textwidth]{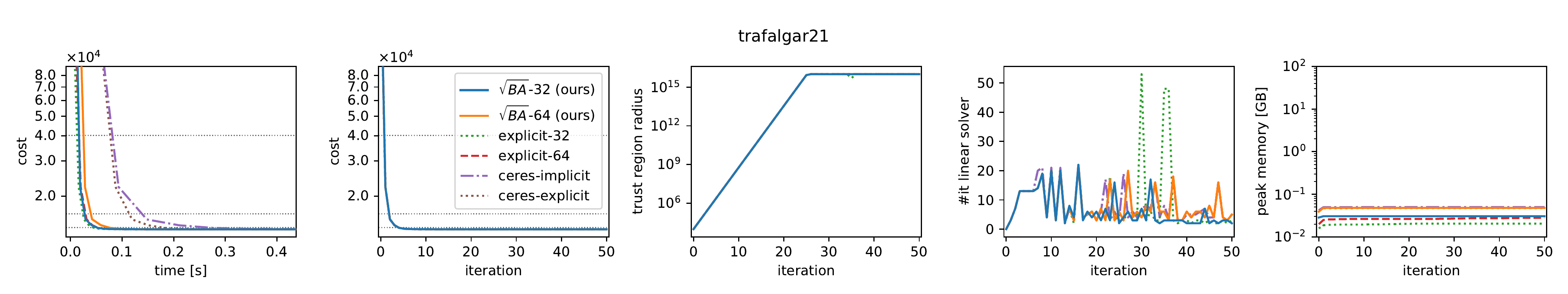}\\
\includegraphics[width=\textwidth]{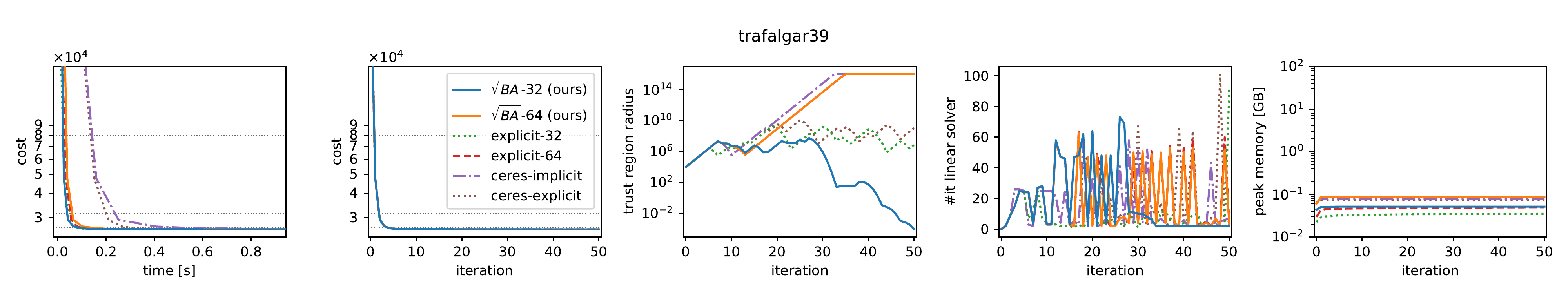}\\
\includegraphics[width=\textwidth]{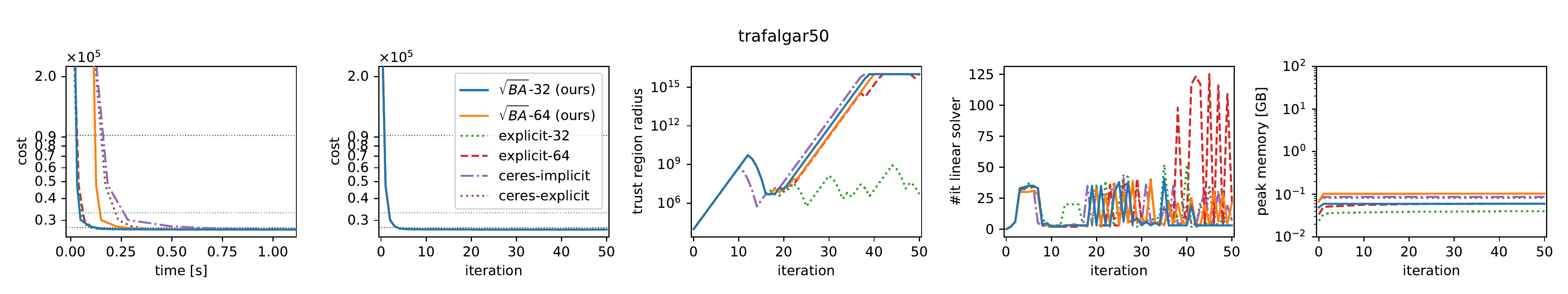}
\includegraphics[width=\textwidth]{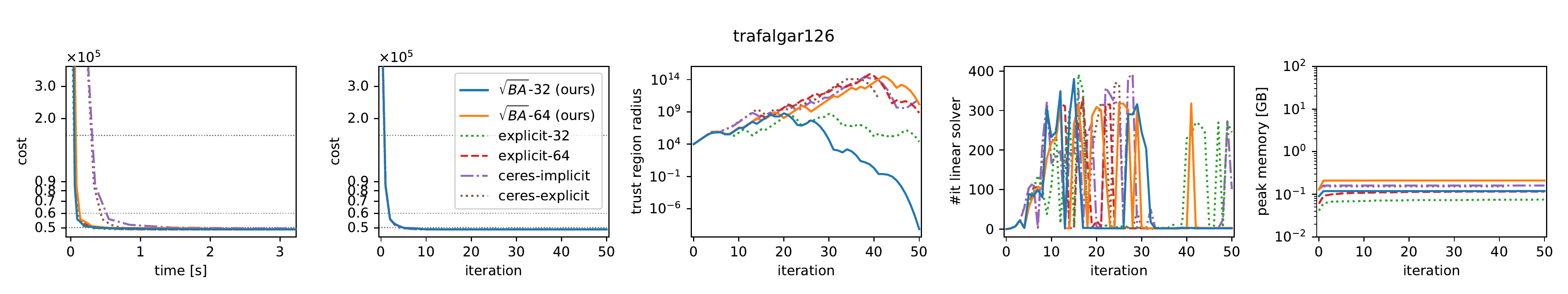}
\includegraphics[width=\textwidth]{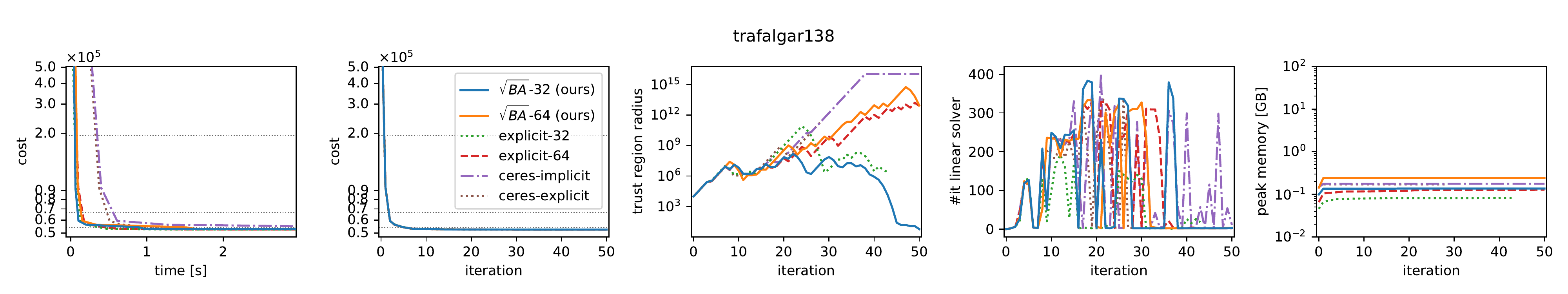}
\includegraphics[width=\textwidth]{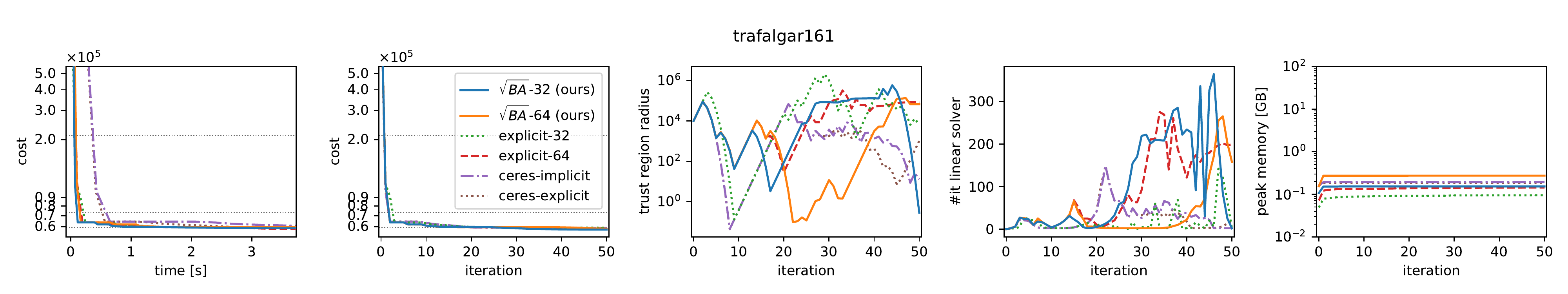}
\includegraphics[width=\textwidth]{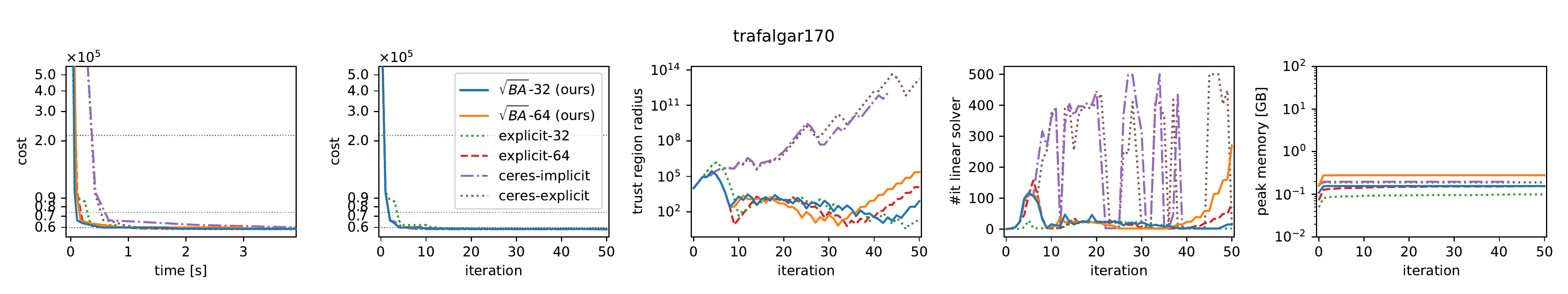}
\includegraphics[width=\textwidth]{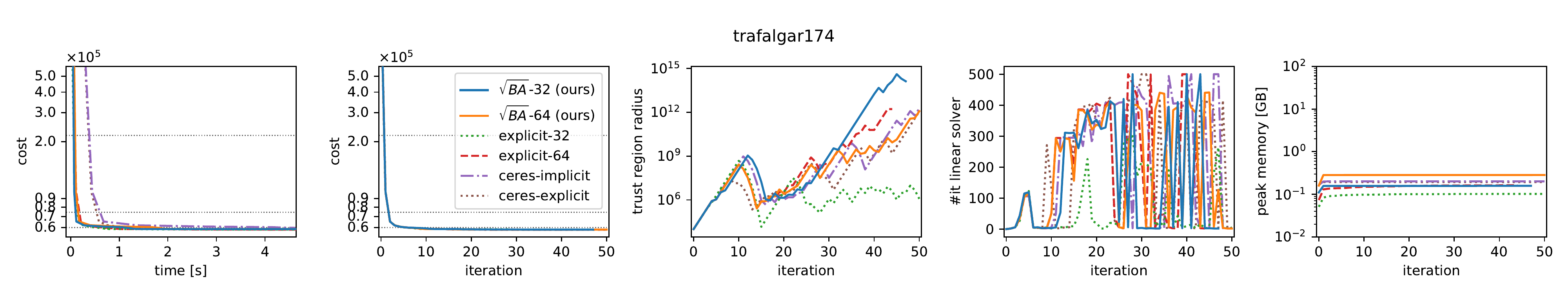}
\includegraphics[width=\textwidth]{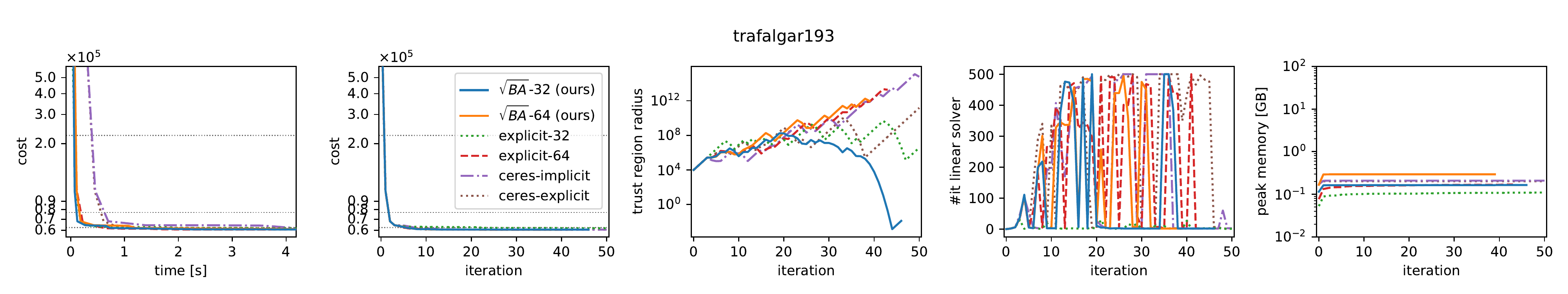}
\includegraphics[width=\textwidth]{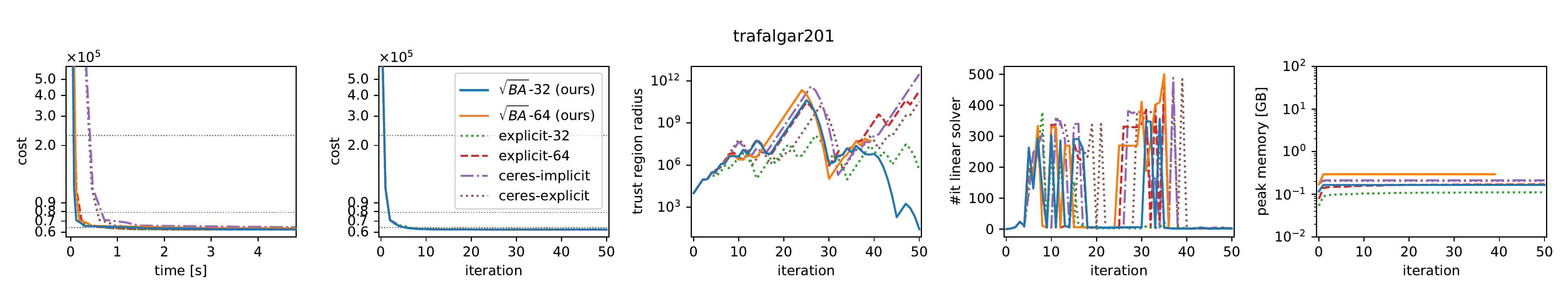}
\includegraphics[width=\textwidth]{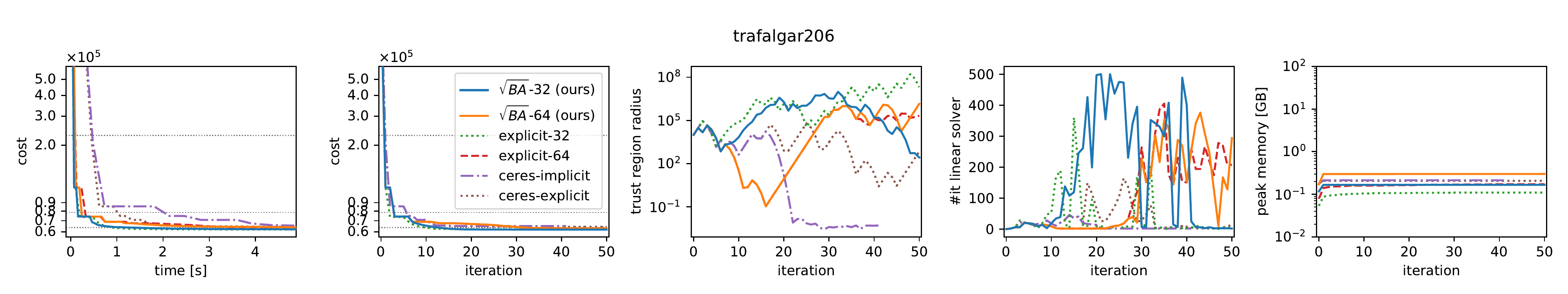}
\includegraphics[width=\textwidth]{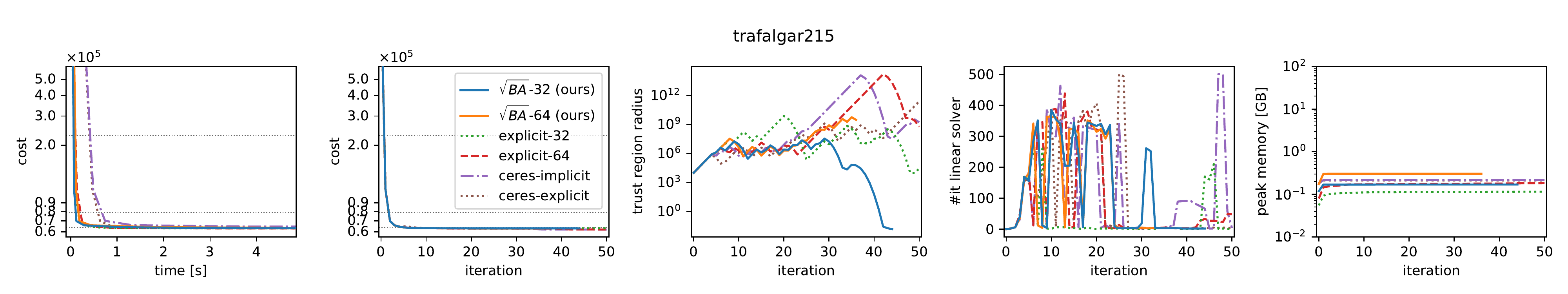}
\includegraphics[width=\textwidth]{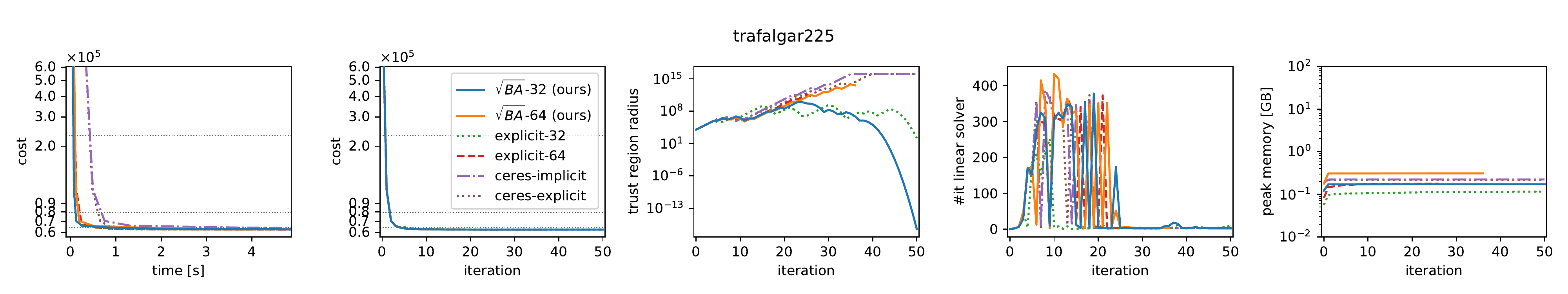}
\includegraphics[width=\textwidth]{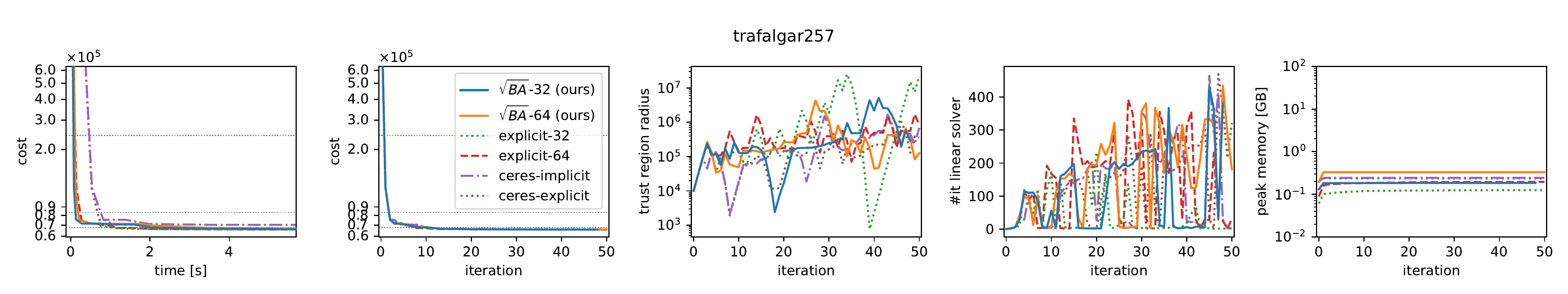}
\end{center}

\subsection{Dubrovnik}
\label{sec:dubrovnik}

\begin{center}
\includegraphics[width=\textwidth]{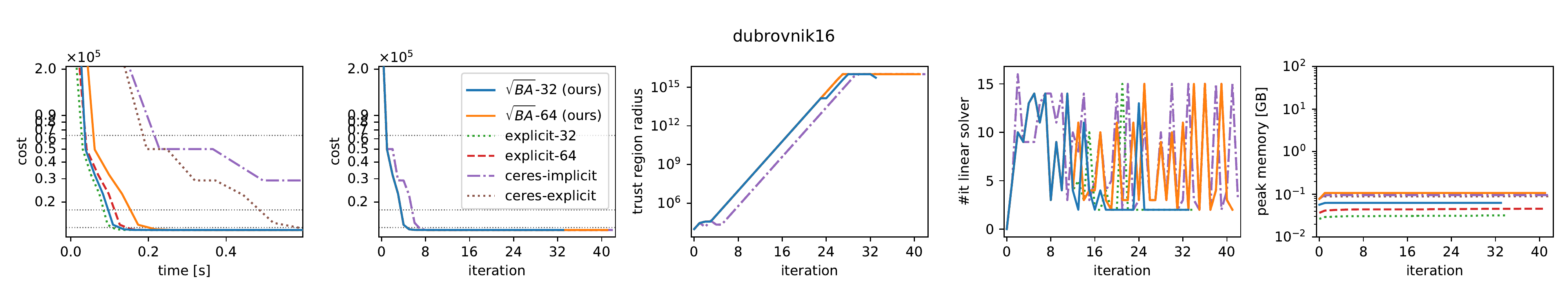}\\
\includegraphics[width=\textwidth]{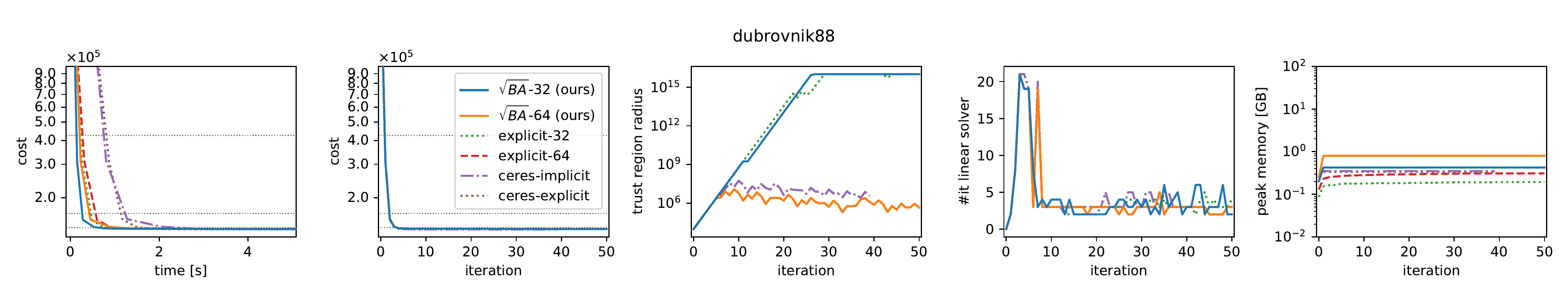}\\
\includegraphics[width=\textwidth]{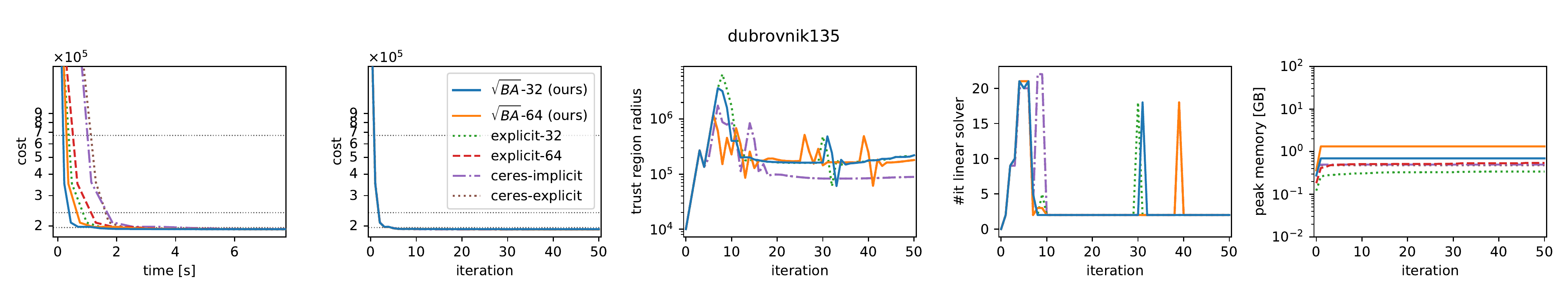}
\includegraphics[width=\textwidth]{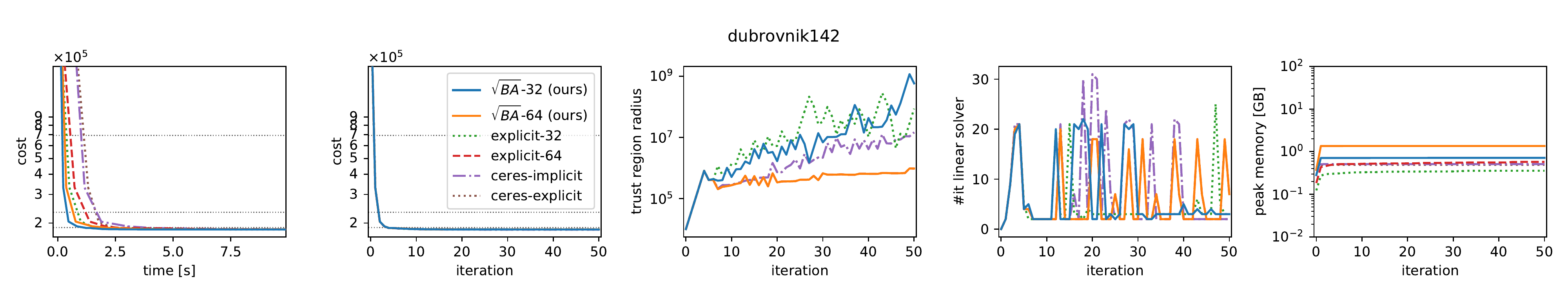}
\includegraphics[width=\textwidth]{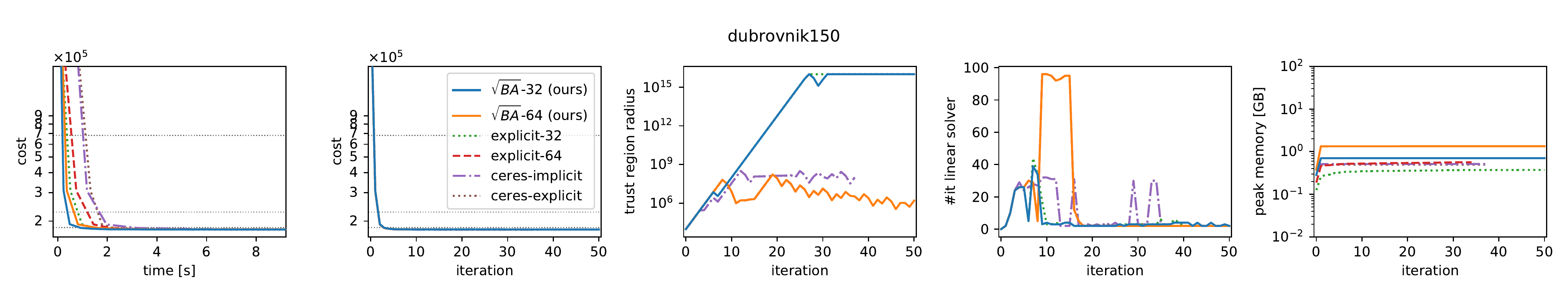}
\includegraphics[width=\textwidth]{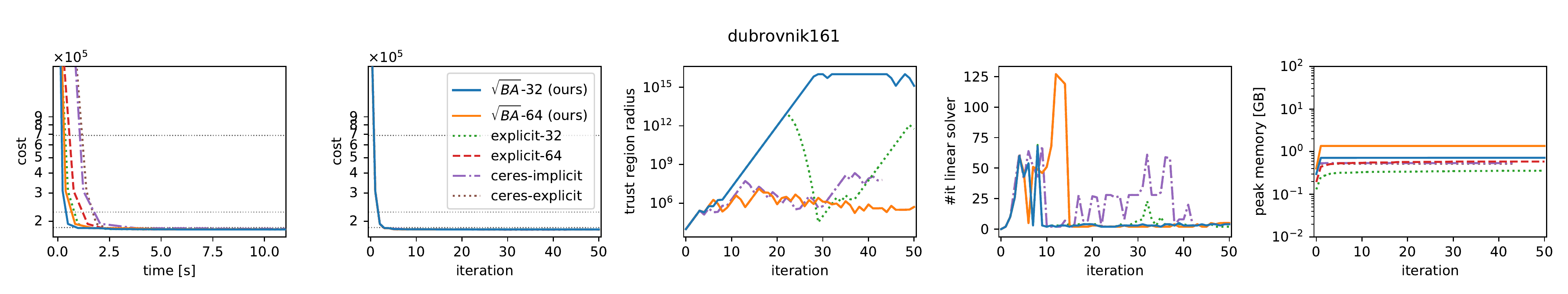}
\includegraphics[width=\textwidth]{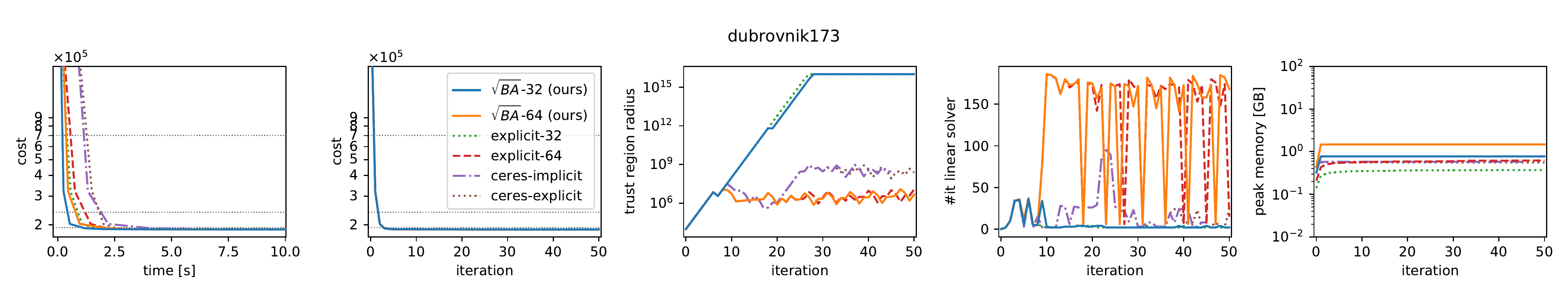}
\includegraphics[width=\textwidth]{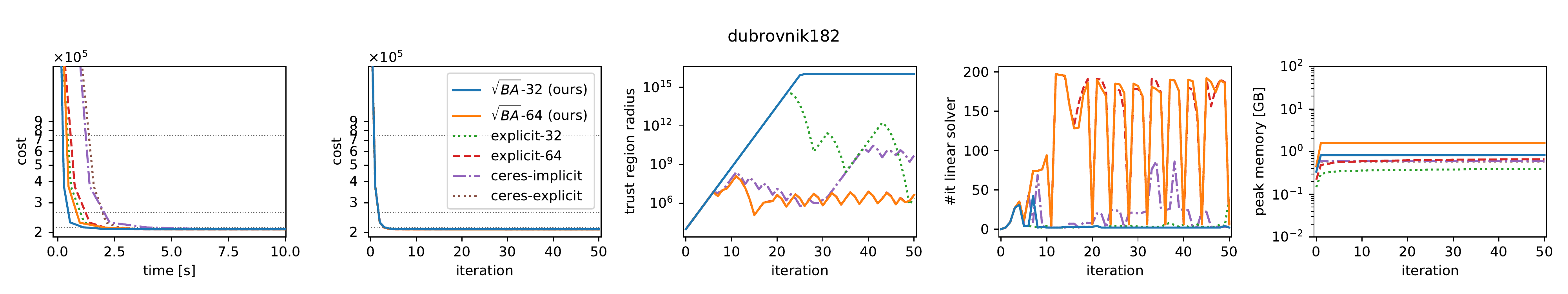}
\includegraphics[width=\textwidth]{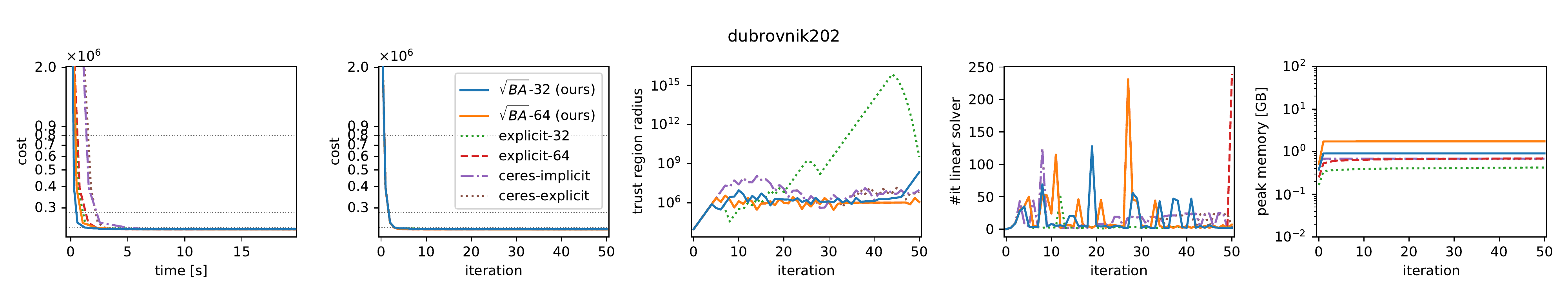}
\includegraphics[width=\textwidth]{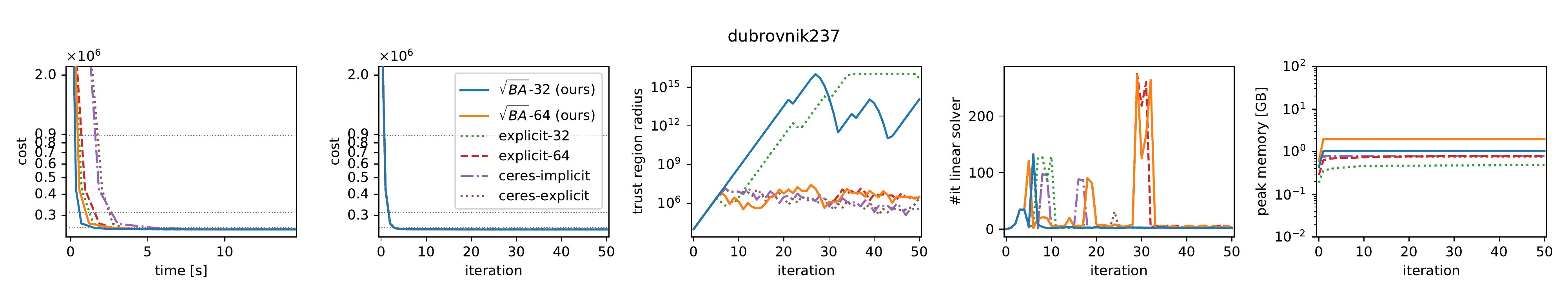}
\includegraphics[width=\textwidth]{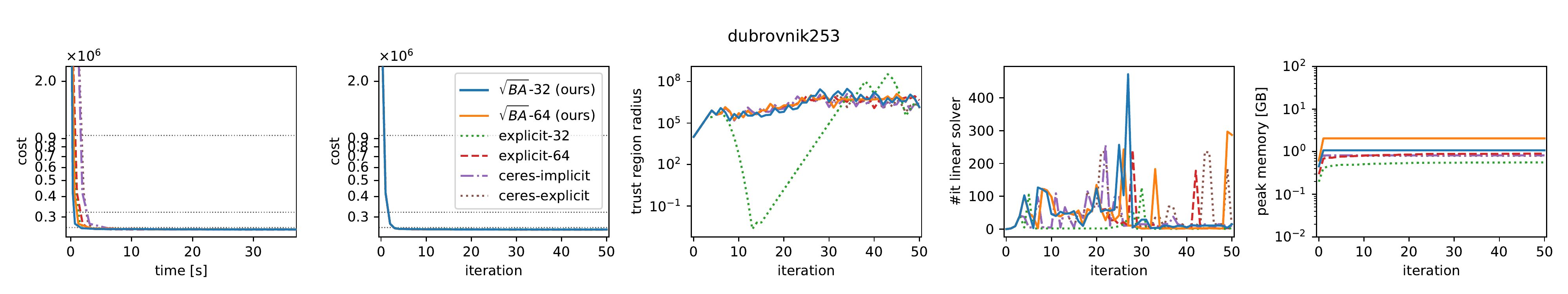}
\includegraphics[width=\textwidth]{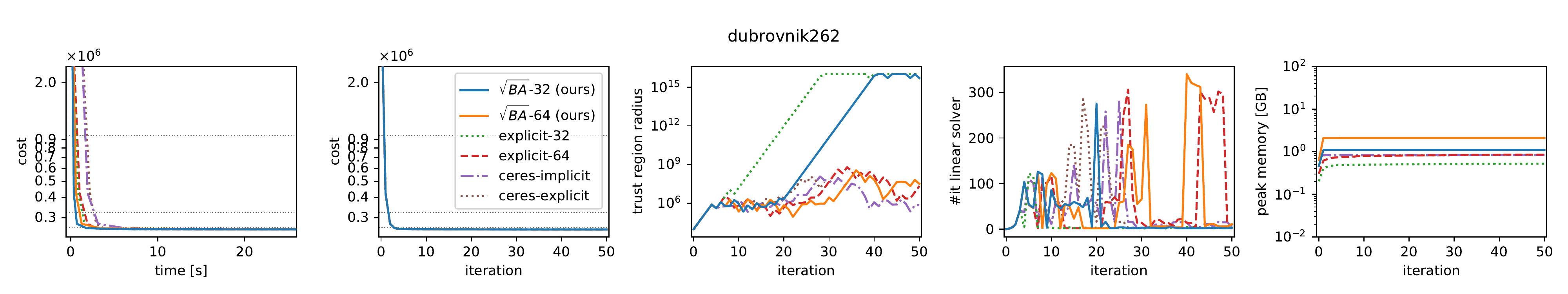}
\includegraphics[width=\textwidth]{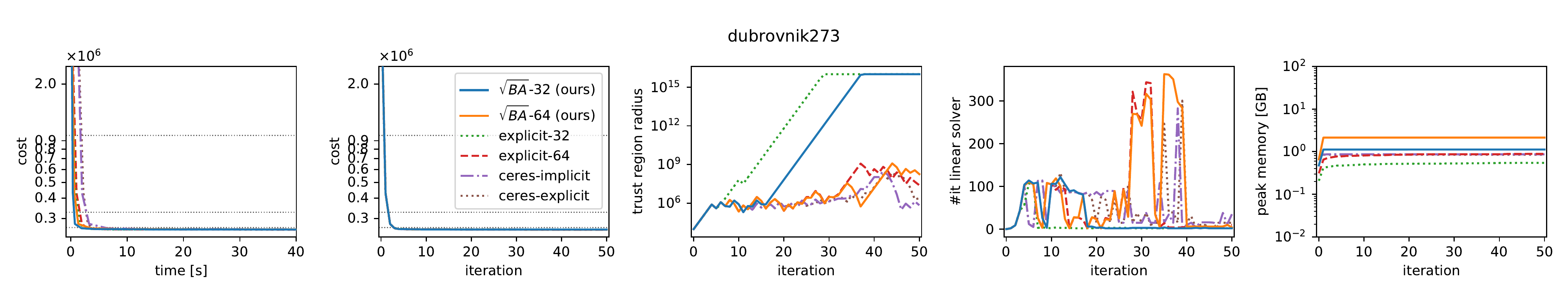}
\includegraphics[width=\textwidth]{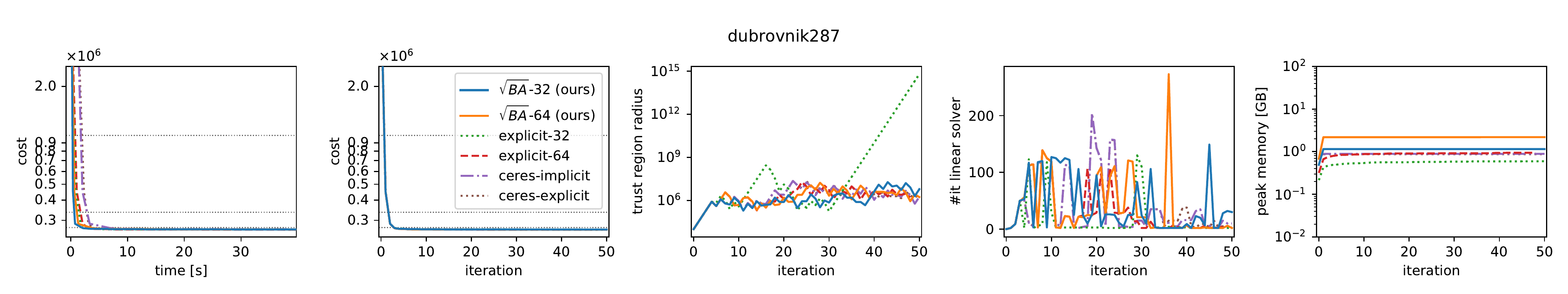}
\includegraphics[width=\textwidth]{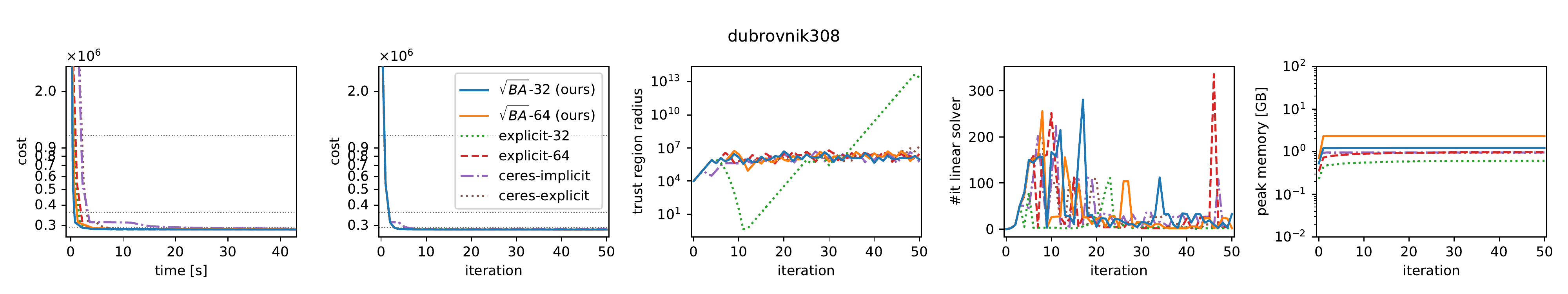}
\includegraphics[width=\textwidth]{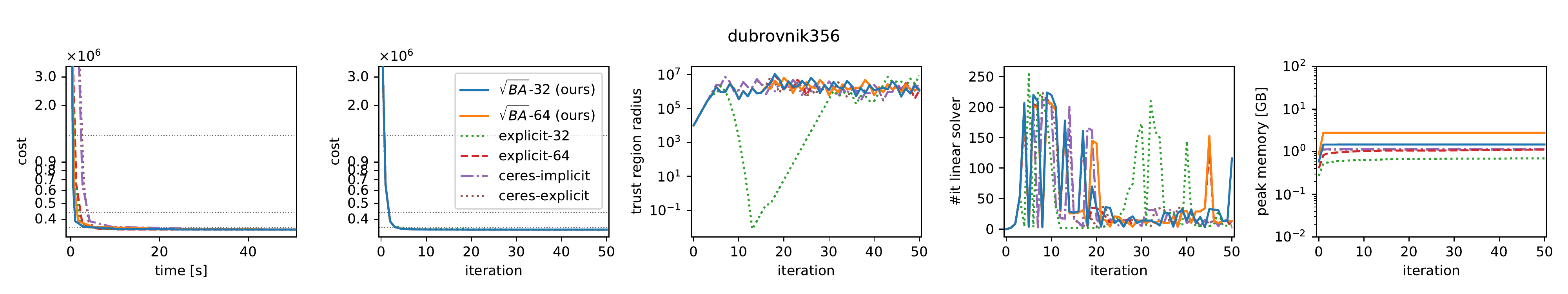}
\end{center}

\subsection{Venice}
\label{sec:venice}

\begin{center}
\includegraphics[width=\textwidth]{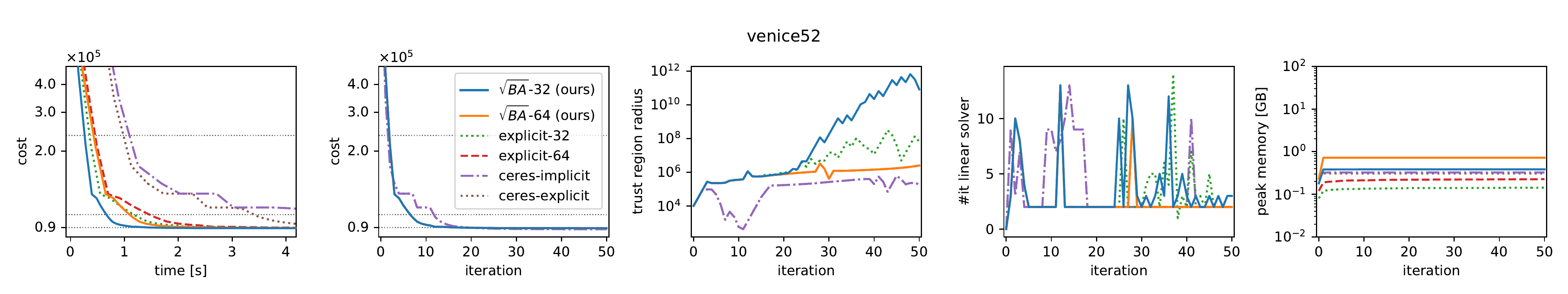}\\
\includegraphics[width=\textwidth]{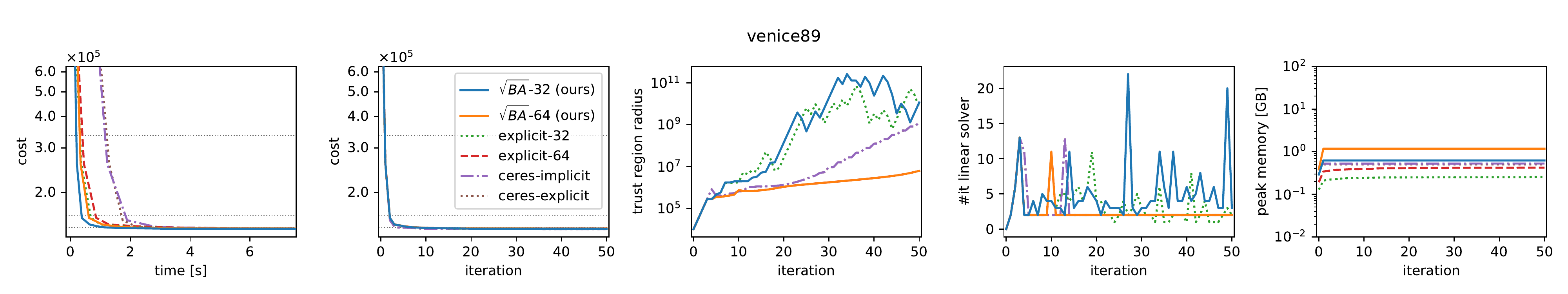}\\
\includegraphics[width=\textwidth]{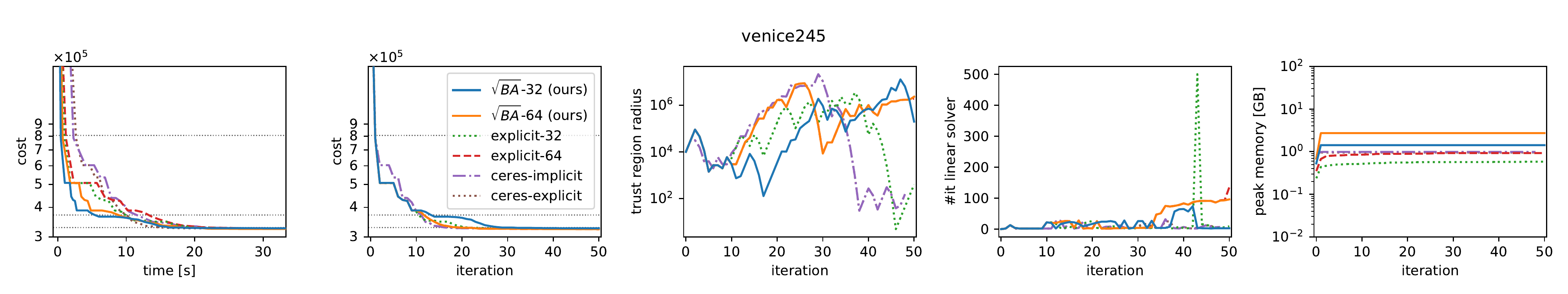}
\includegraphics[width=\textwidth]{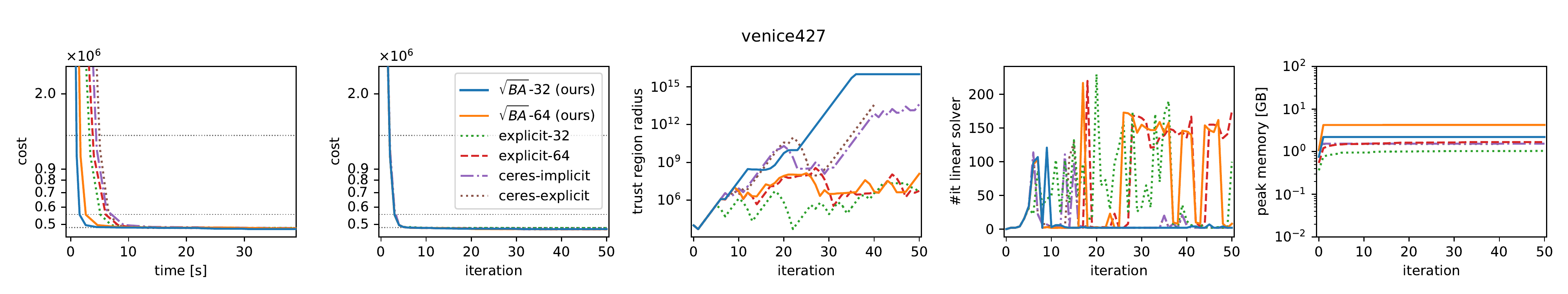}
\includegraphics[width=\textwidth]{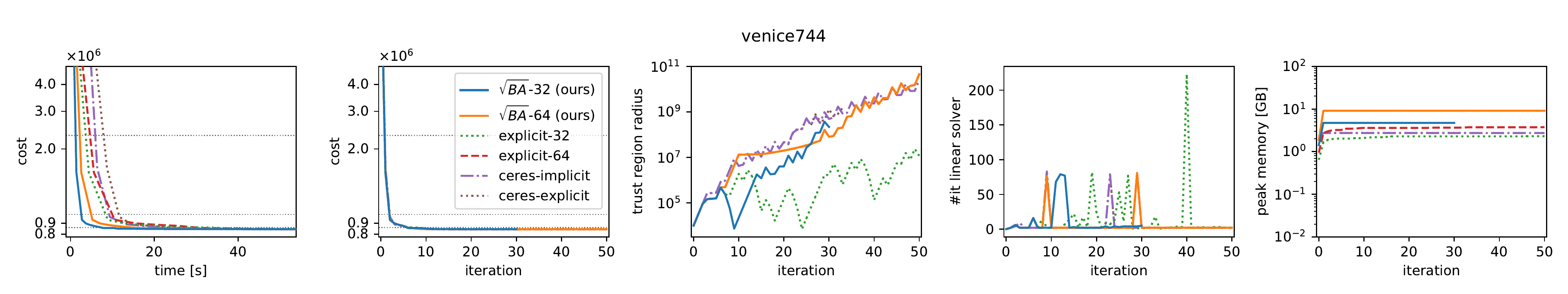}
\includegraphics[width=\textwidth]{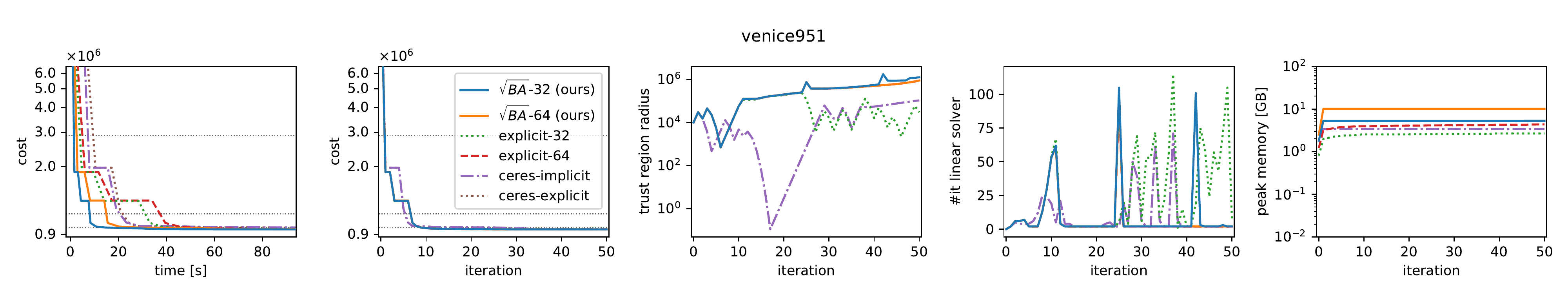}
\includegraphics[width=\textwidth]{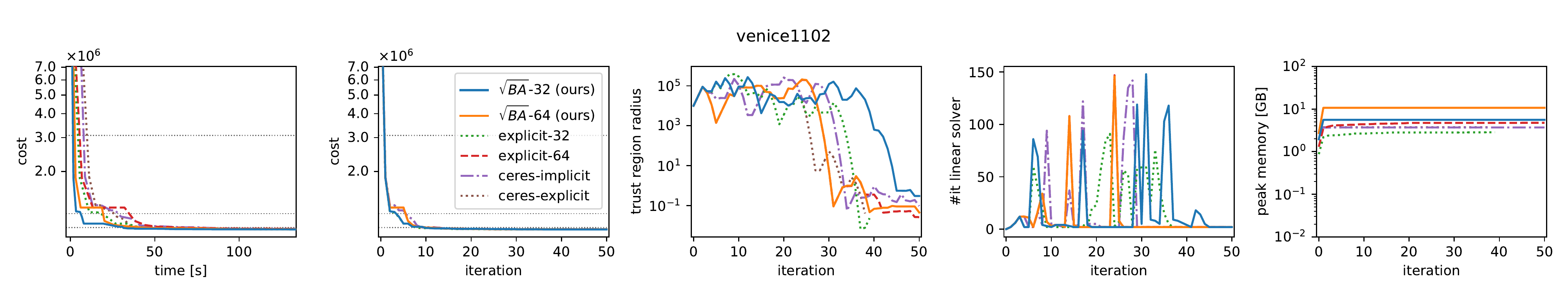}
\includegraphics[width=\textwidth]{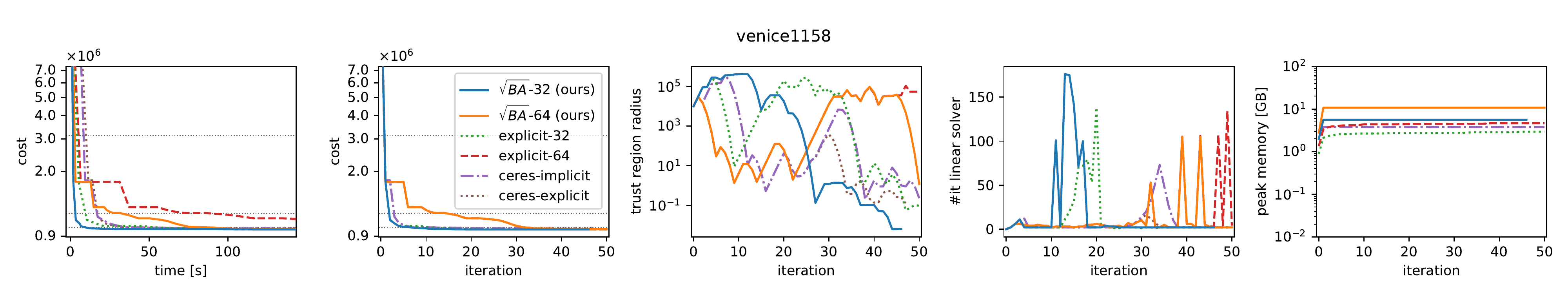}
\includegraphics[width=\textwidth]{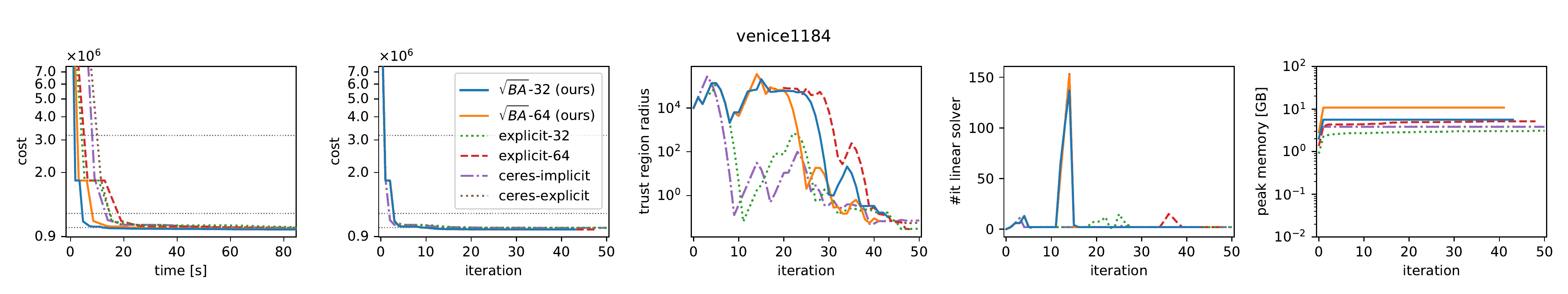}
\includegraphics[width=\textwidth]{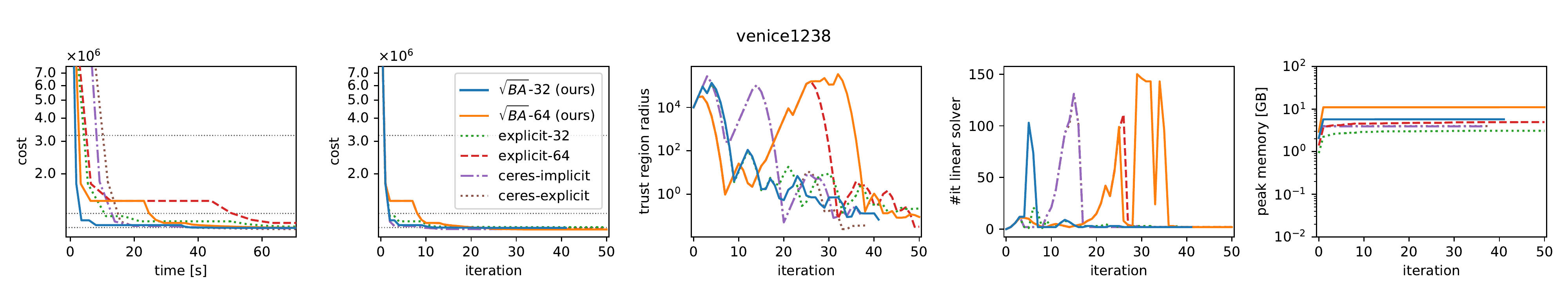}
\includegraphics[width=\textwidth]{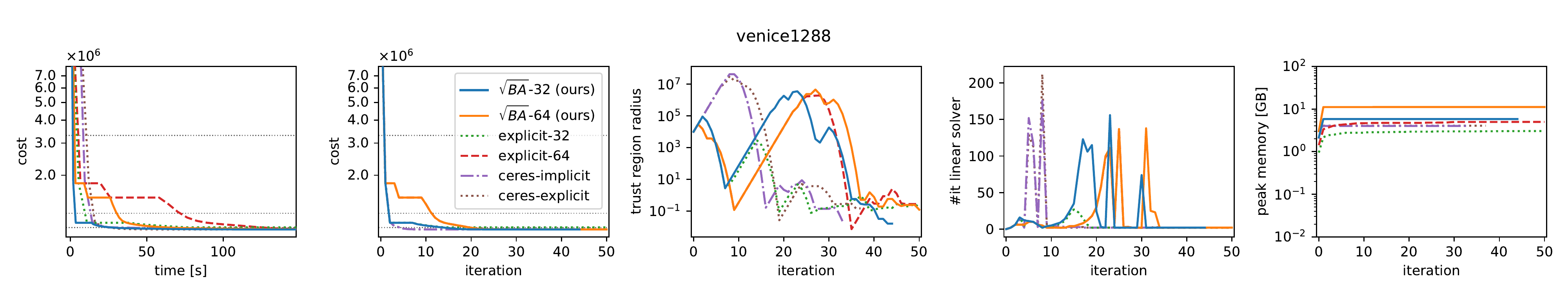}
\includegraphics[width=\textwidth]{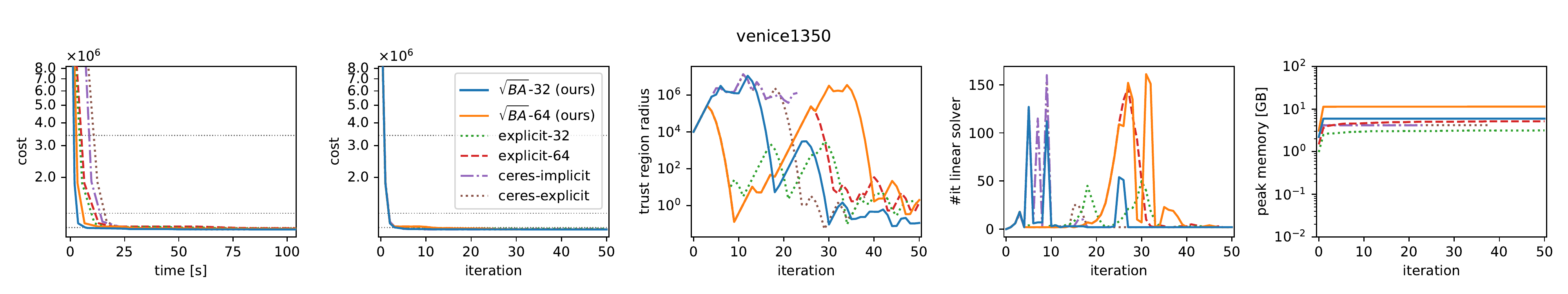}
\includegraphics[width=\textwidth]{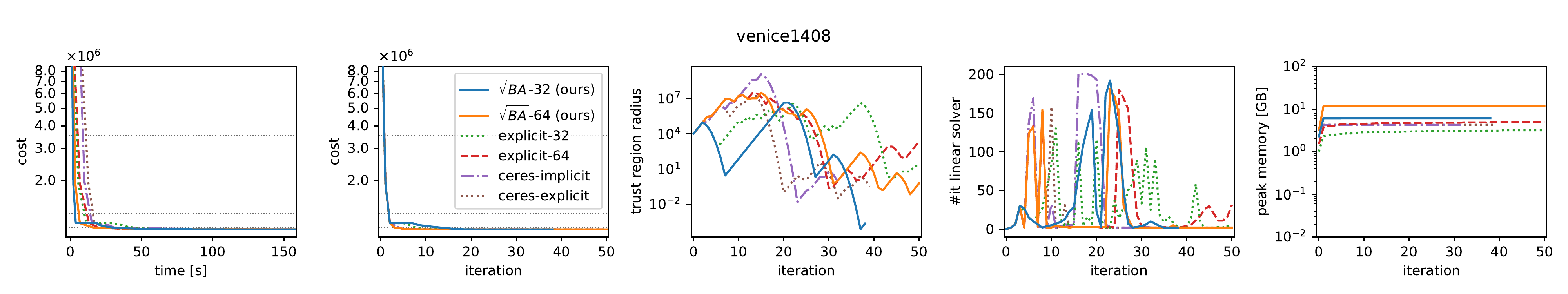}
\includegraphics[width=\textwidth]{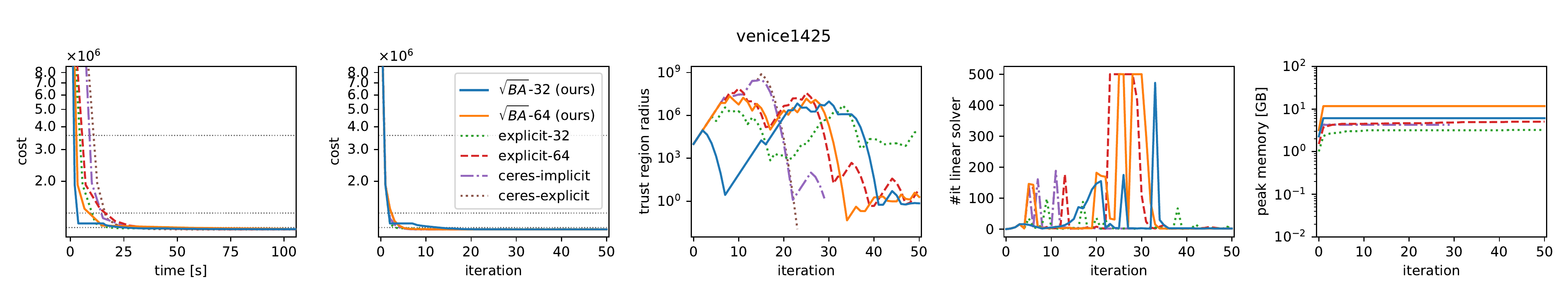}
\includegraphics[width=\textwidth]{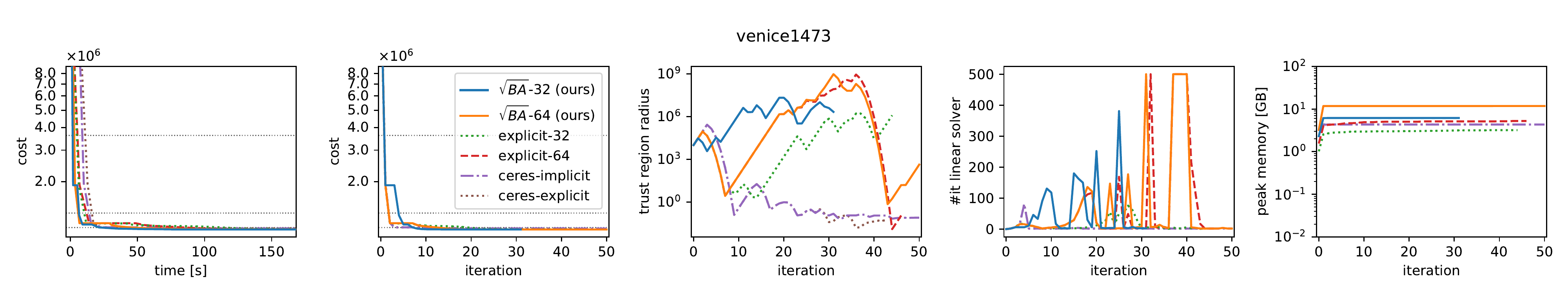}
\includegraphics[width=\textwidth]{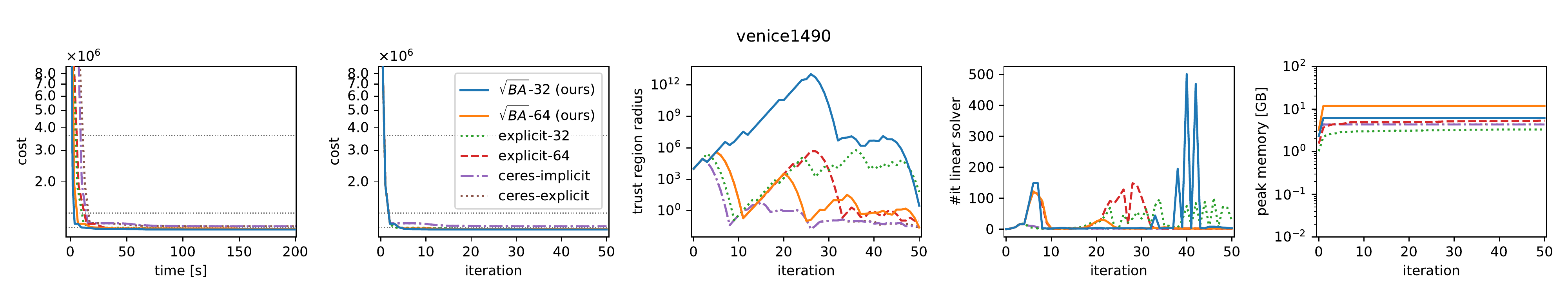}
\includegraphics[width=\textwidth]{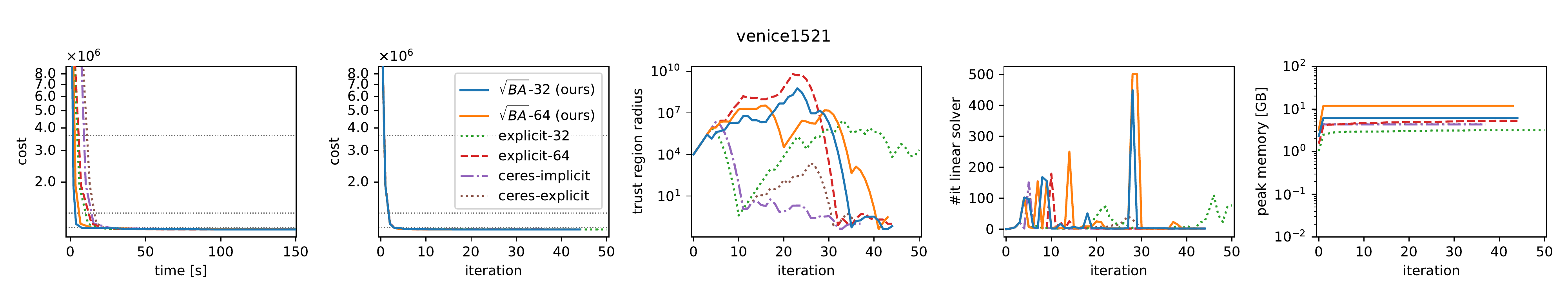}
\includegraphics[width=\textwidth]{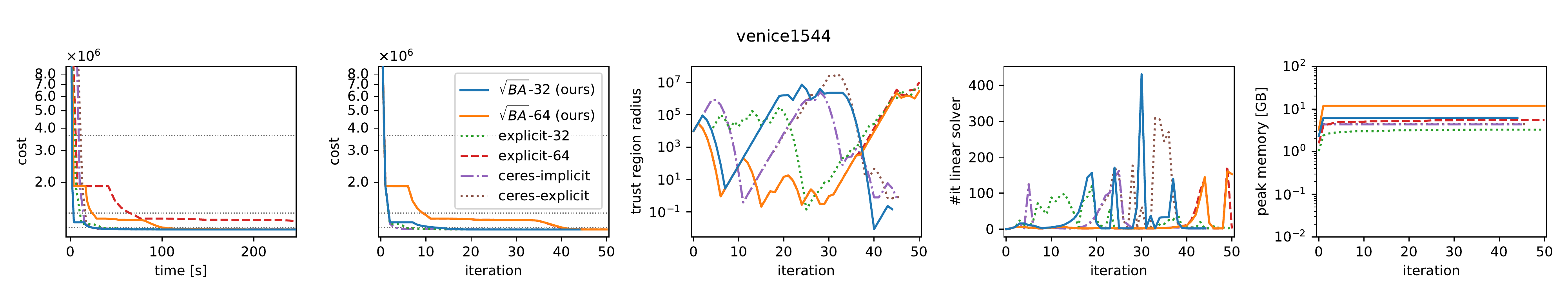}
\includegraphics[width=\textwidth]{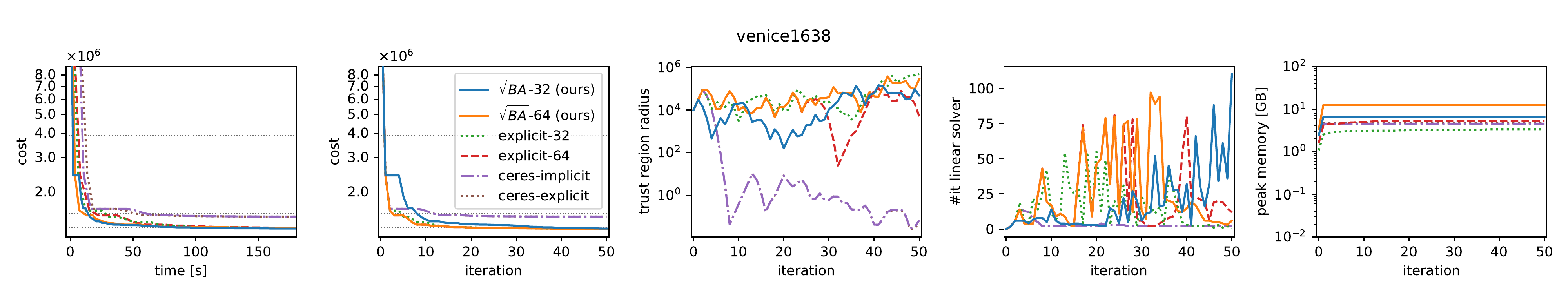}
\includegraphics[width=\textwidth]{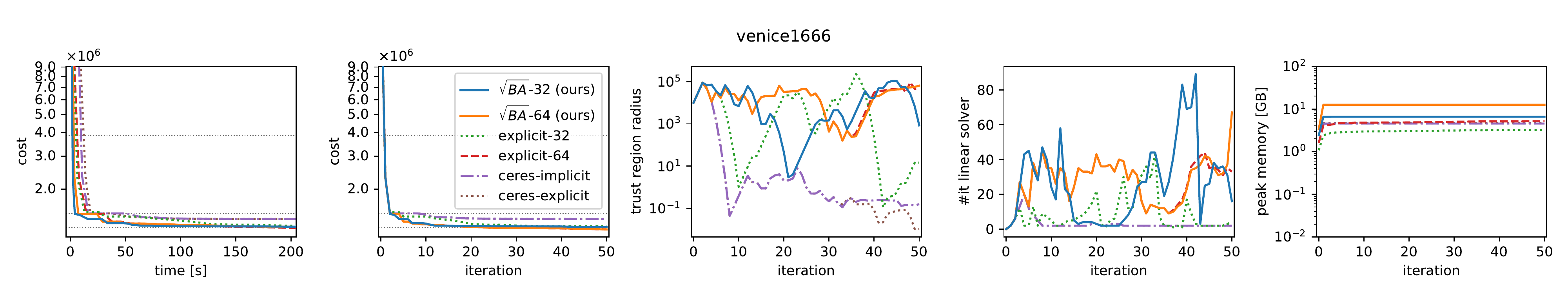}
\includegraphics[width=\textwidth]{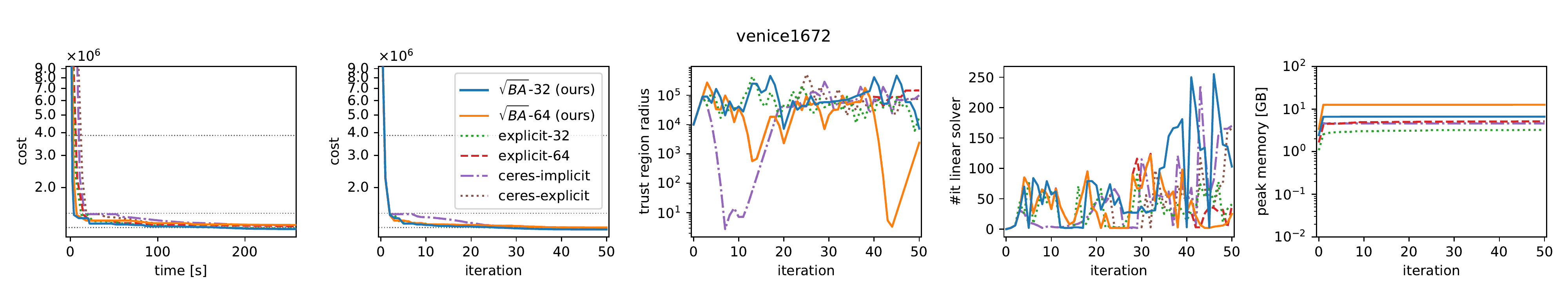}
\includegraphics[width=\textwidth]{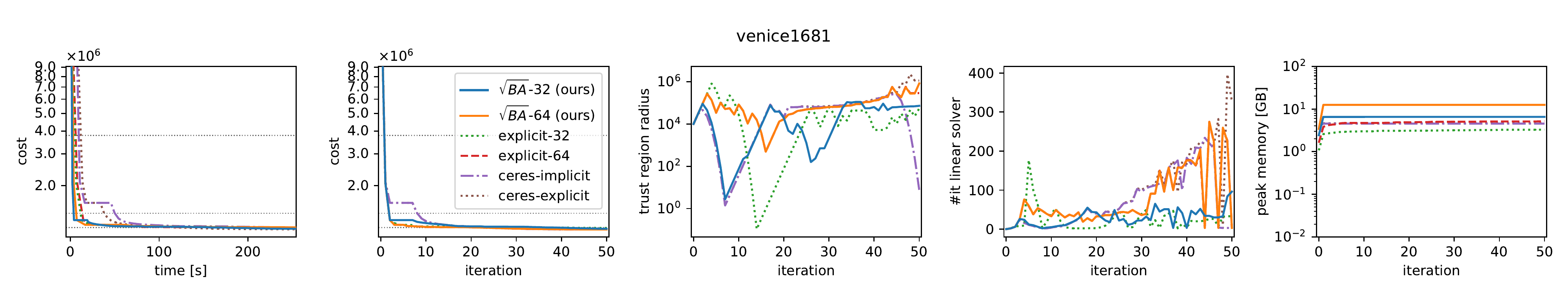}
\includegraphics[width=\textwidth]{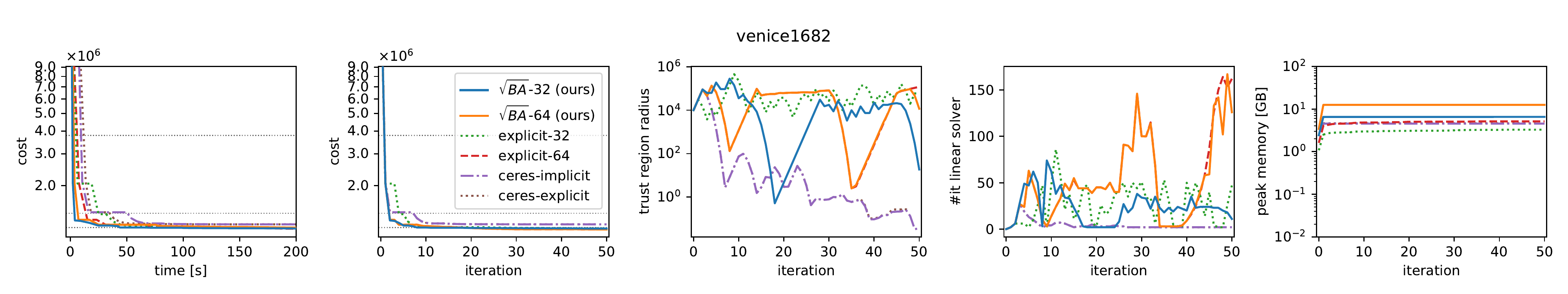}
\includegraphics[width=\textwidth]{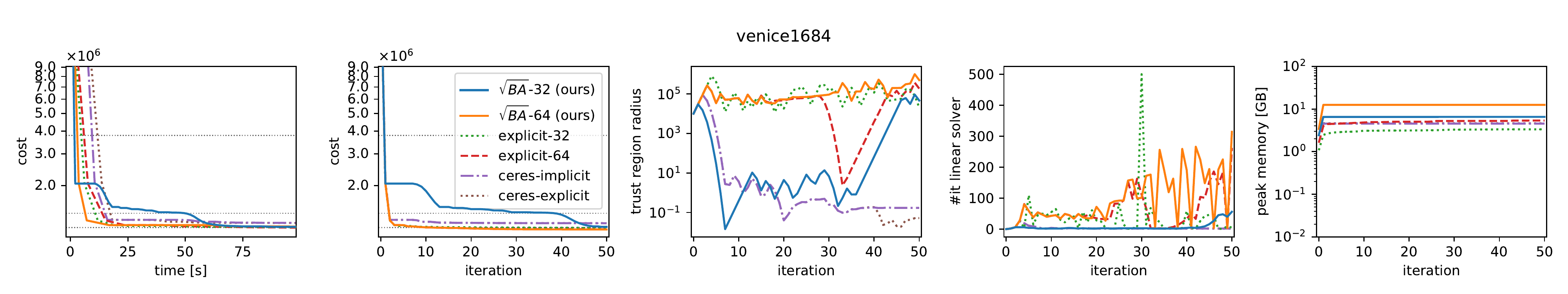}
\includegraphics[width=\textwidth]{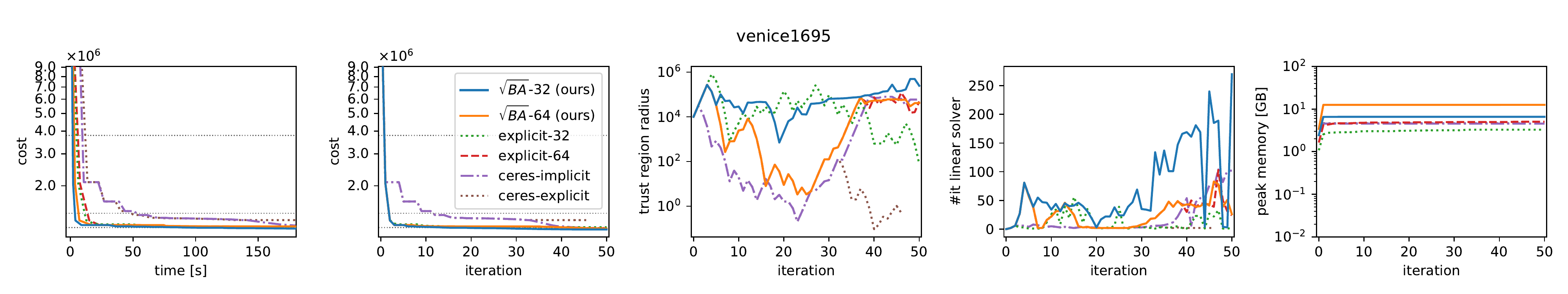}
\includegraphics[width=\textwidth]{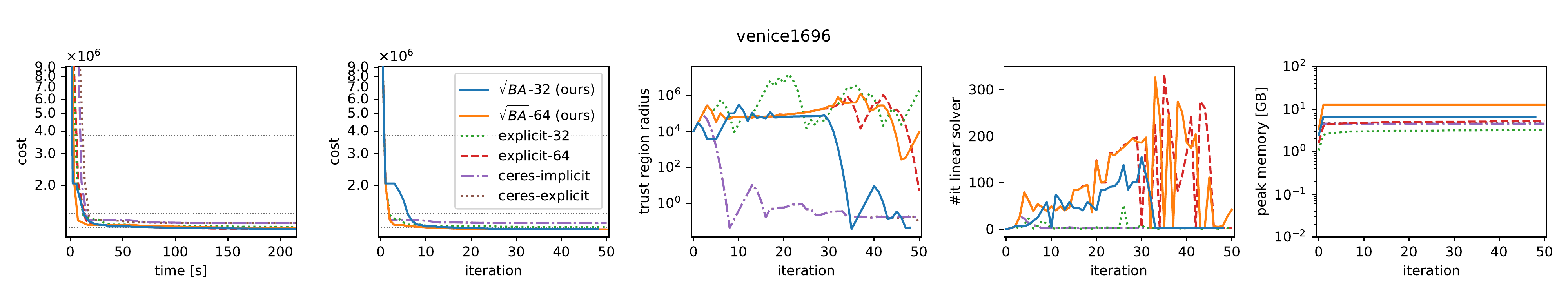}
\includegraphics[width=\textwidth]{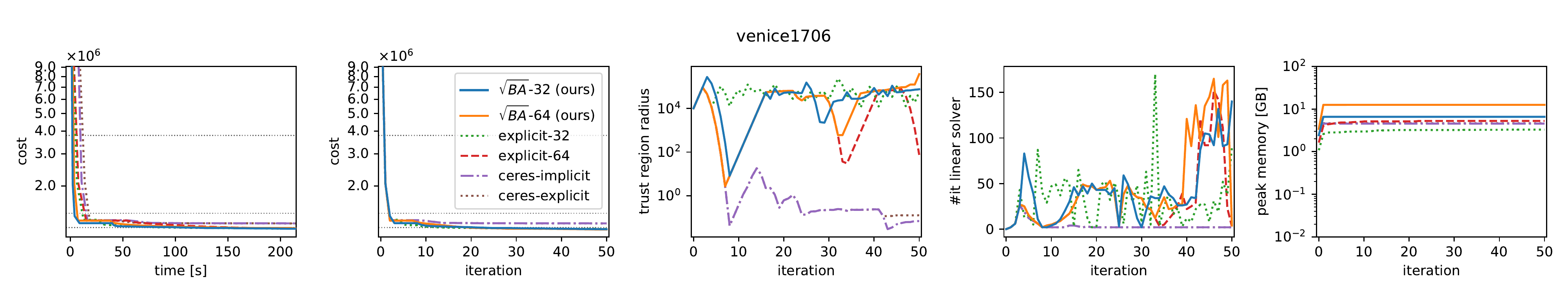}
\includegraphics[width=\textwidth]{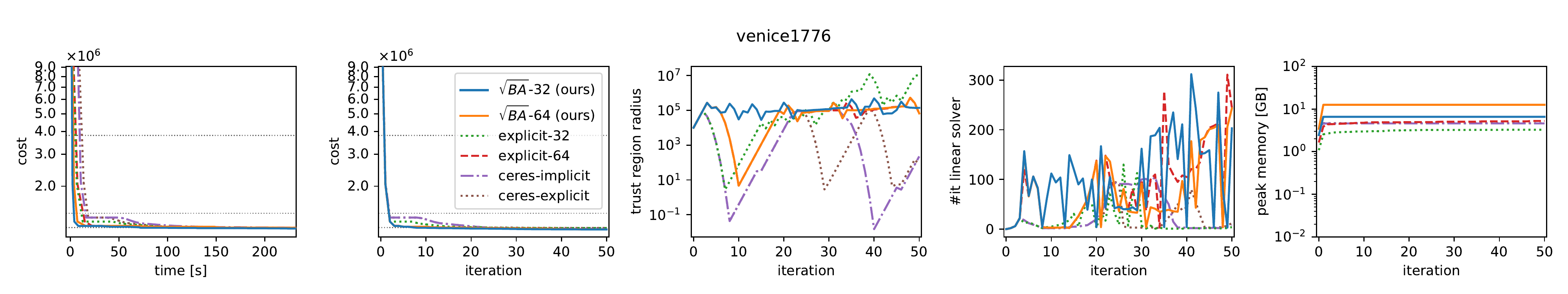}
\includegraphics[width=\textwidth]{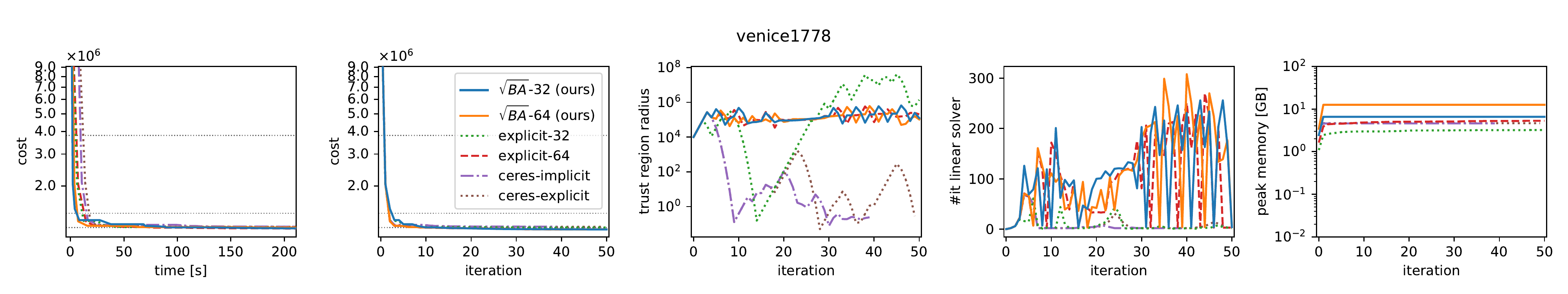}
\end{center}

\subsection{Final}
\label{sec:final}

\begin{center}
\includegraphics[width=\textwidth]{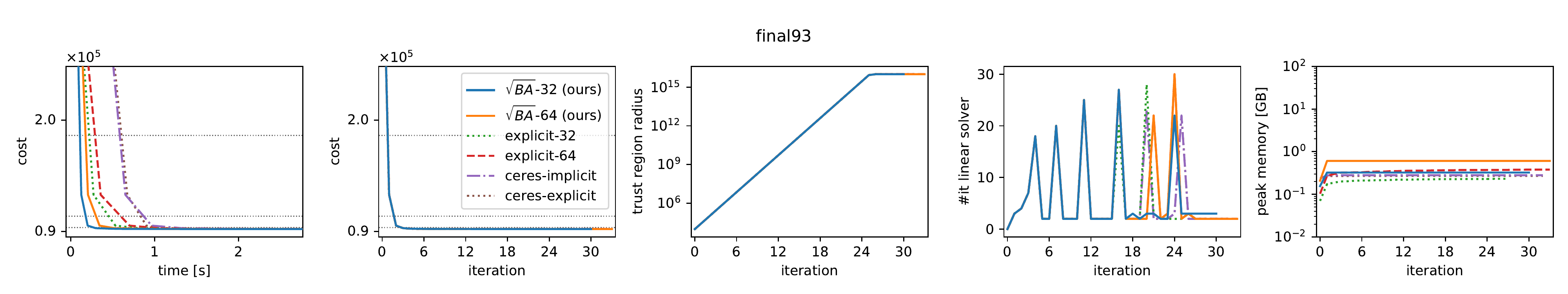}\\
\includegraphics[width=\textwidth]{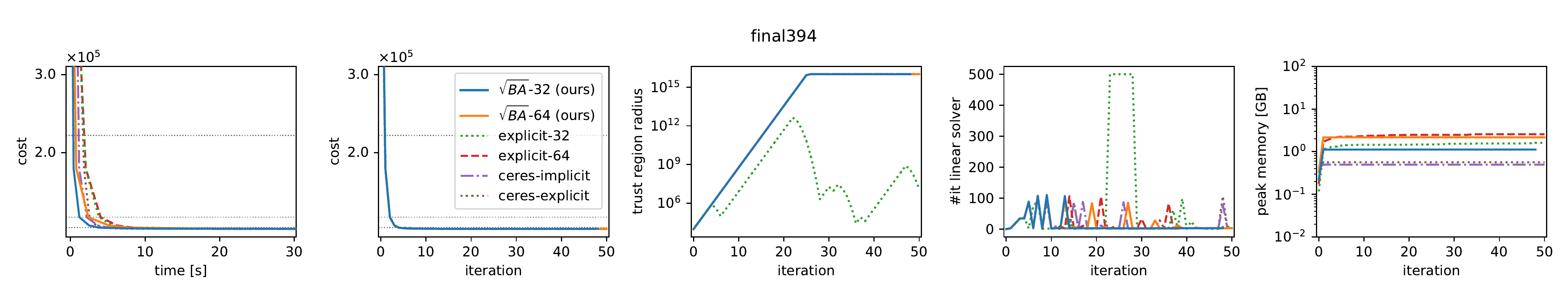}\\
\includegraphics[width=\textwidth]{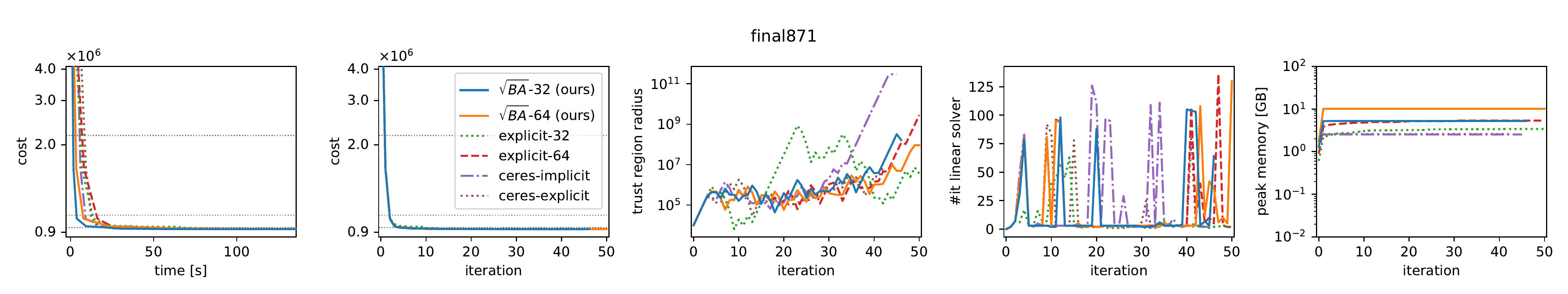}
\includegraphics[width=\textwidth]{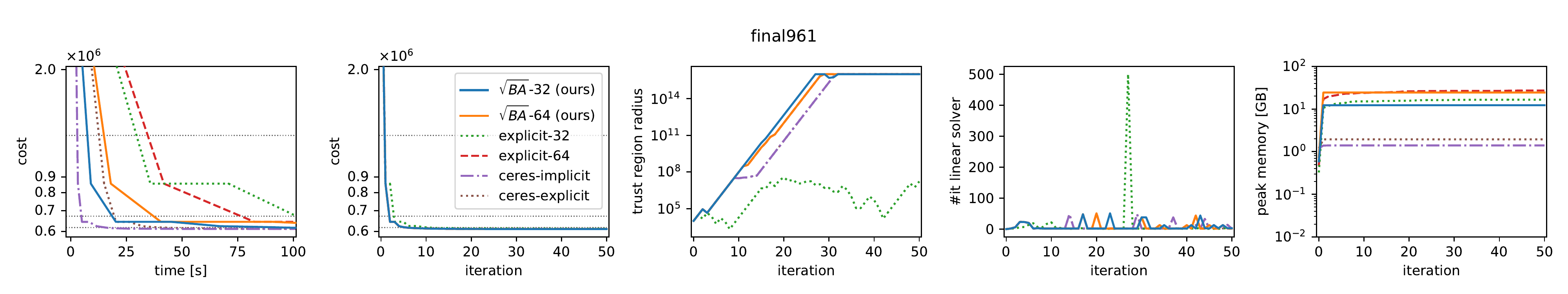}
\includegraphics[width=\textwidth]{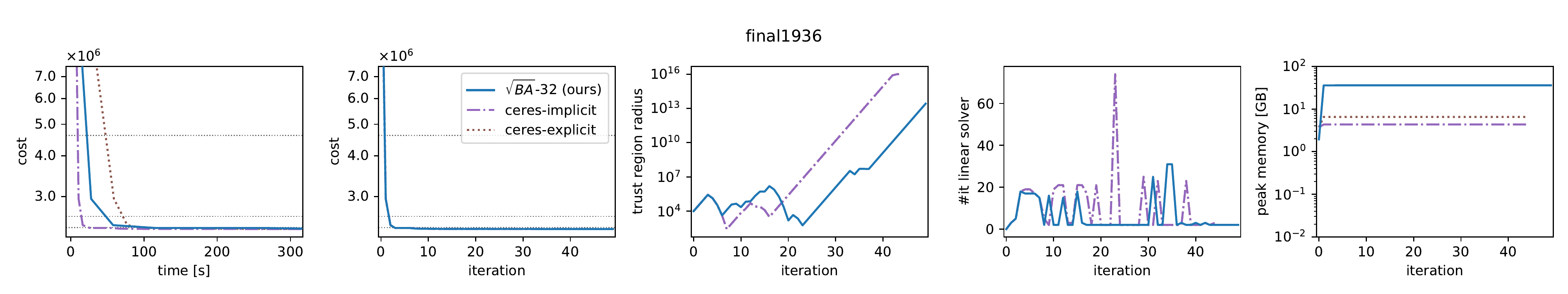}
\includegraphics[width=\textwidth]{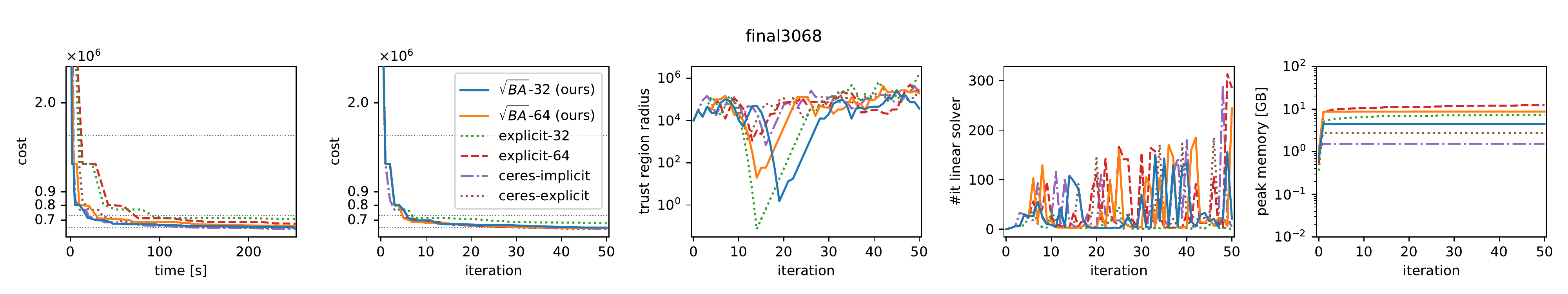}
\includegraphics[width=\textwidth]{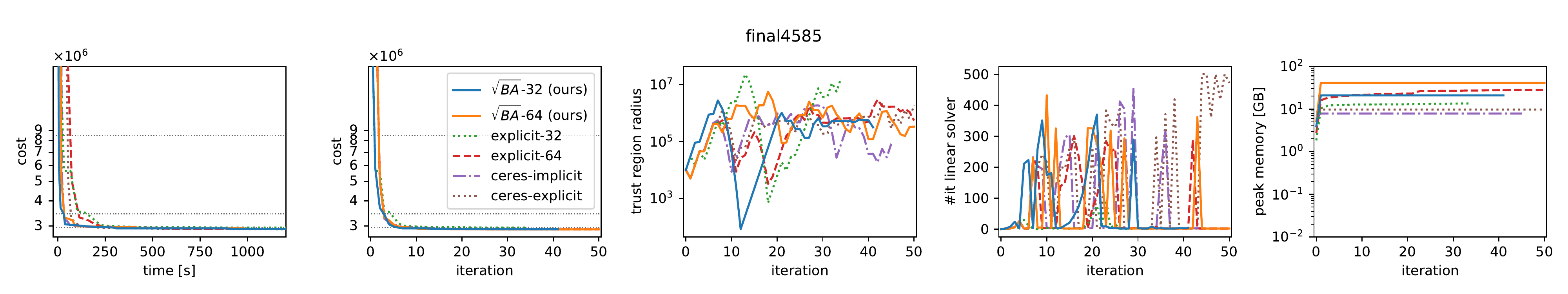}
\includegraphics[width=\textwidth]{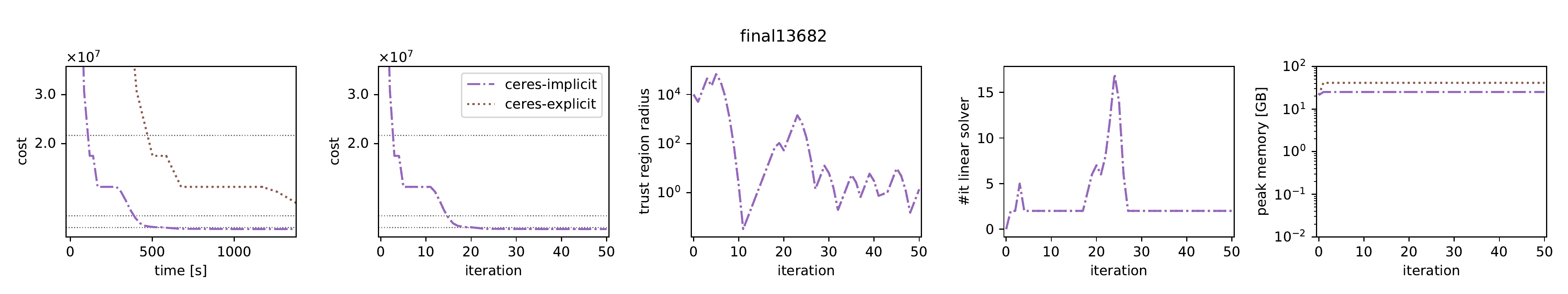}
\end{center}

\end{document}